\documentclass[a4paper,fleqn]{cas-sc}
\usepackage{framed} 
\usepackage{multicol} 
\usepackage{nomencl} 
\usepackage{booktabs}
\usepackage{lipsum}

\newcommand\blfootnote[1]{%
  \begingroup
  \renewcommand\thefootnote{}\footnote{#1}%
  \addtocounter{footnote}{-1}%
  \endgroup
}

\makenomenclature
\renewcommand*\nompreamble{\begin{multicols}{2}}
\renewcommand*\nompostamble{\end{multicols}}

\usepackage[numbers]{natbib}
\usepackage{longtable}
\usepackage{hyperref}

\def\tsc#1{\csdef{#1}{\textsc{\lowercase{#1}}\xspace}}
\tsc{WGM}
\tsc{QE}
\tsc{EP}
\tsc{PMS}
\tsc{BEC}
\tsc{DE}

\begin{document}
\let\WriteBookmarks\relax
\def\floatpagepagefraction{1}
\def\textpagefraction{.001}
\shorttitle{Scientometric Analysis of AI for Wind Energy}
\shortauthors{Chatterjee and Dethlefs}

\title [mode = title]{Scientometric Review of Artificial Intelligence for Operations \& Maintenance of Wind Turbines: The Past, Present and Future}                      

\author[1]{Joyjit Chatterjee}[orcid=0000-0003-2672-3832]
\cormark[1]
\fnmark[1]
\ead{j.chatterjee-2018@hull.ac.uk}

\credit{Conceptualization, Methodology, Software, Formal analysis, Investigation, Writing - Original Draft}

\address{Department of Computer Science \& Technology, Dependable Intelligent Systems Research Group, University of Hull, Cottingham Road, Hull, HU6 7RX, United Kingdom}

\author[2]{Nina Dethlefs}[orcid=0000-0002-6917-5066]

\fnmark[2]
\ead{n.dethlefs@hull.ac.uk}

\credit{Supervision, Methodology, Writing - Review \& Editing}


\maketitle

\begin{abstract}
Wind energy has emerged as a highly promising source of renewable energy in recent times. However, wind turbines regularly suffer from operational inconsistencies, leading to significant costs and challenges in operations and maintenance
 (O\&M). Condition-based monitoring (CBM) and performance assessment/analysis of turbines are vital aspects for ensuring efficient O\&M planning and cost minimisation. Data-driven decision making techniques have witnessed rapid evolution in the wind industry for such O\&M tasks during the last decade, from applying signal processing methods in early 2010 to artificial intelligence (AI) techniques, especially deep learning in 2020. In this article, we utilise statistical computing to present a scientometric review of the conceptual and thematic evolution of AI in the wind energy sector, providing evidence-based insights into present strengths and limitations of data-driven decision making in the wind industry. We provide a perspective into the future and on current key challenges in data availability and quality, lack of transparency in black box-natured AI models, and prevailing issues in deploying models for real-time decision support, along with possible strategies to overcome these problems. We hope that a systematic analysis of the past, present and future of CBM and performance assessment can encourage more organisations to adopt data-driven decision making techniques in O\&M towards making wind energy sources more reliable, contributing to the global efforts of tackling climate change. 
\end{abstract}




\section{Introduction}
\begin{table*}[!h]
   \begin{framed}
     \printnomenclature
   \end{framed}
\end{table*}

The worldwide capacity of wind power generation has continued to evolve rapidly, with increasing deployments of wind farms, especially offshore \cite{globalcap_review,STETCO2019620}. \blfootnote{This is a preprint version of the accepted manuscript in the Renewable and Sustainable Energy Reviews journal, shared under a CC-BY-NC-ND license. The final published version can be found at: \url{https://doi.org/10.1016/j.rser.2021.111051}} However, owing to the complexity of the deployed turbine environments and the very nature of their electrical and mechanical components \cite{windenergy_journal}, they experience irregular loads and operational inconsistencies. \nomenclature{$O\&M$}{Operations and maintenance} Operations and maintenance (O\&M) is crucial for preventing such incidents and/or providing corrective actions to avert/fix any occurring faults \cite{Rockmann2017}. 

A vital aspect of O\&M is \nomenclature{$CBM$}{Condition-based monitoring} Condition-based monitoring (CBM), which plays an integral role in identifying operational changes in various turbine components \cite{STETCO2019620}. CBM methods span the areas of fault detection, fault prediction/prognosis and fault diagnosis \cite{en12020201}, and facilitate condition-based maintenance for early detection of any degradation or incipient faults before they can lead to significantly costly failures. Besides this purpose, CBM can help in keeping healthy turbines in continued operation, reducing outages which can occur due to redundantly scheduled maintenance activities \cite{ibrahim_cbmchallenge}. Besides CBM for turbine control, failure diagnosis and prediction, there are some other areas which are integral to O\&M planning and performance assessment/analysis for energy cost minimisation \cite{perf_energycost}. These pertain to design optimisation of turbines and wind farms, forecasting and prediction of vital parameters (like wind speed, power, torque, power factor) etc. \cite{en12020225}. All such activities are essential for ensuring efficient O\&M, especially for offshore wind power systems owing to the multifaceted systems and the harsh environments in which they generally operate \cite{LIN2020117693}.

During the last decade, most existing studies have utilised signal processing or physics-based numerical models towards CBM pertaining to turbine health monitoring, particularly leveraging vibration data for this purpose. More recently, with the rising interest in adopting data-driven solutions for CBM \cite{STETCO2019620,recent_cbm} and performance assessment/analysis of turbines \cite{7947229,perf_energycost}, \nomenclature{$AI$}{Artificial intelligence} Artificial intelligence (AI) techniques have been applied for decision making to learn from \nomenclature{$SCADA$}{Supervisory Control \& Acquisition} Supervisory Control \& Acquisition (SCADA) data regularly generated by turbines through various sensors. \cite{yang_jiang_2011,scada_windturbine_review,scada_planning}. While AI techniques have been game changers for many domains such as healthcare and finance \cite{exp_review,wang2018blessings}, the wind industry has not benefited as much from recent advances in AI, especially in deep learning, likely due to lack of a clear perspective and limited trust in such models. The multitude of directions in applying AI techniques (e.g. for supervised learning, unsupervised learning, reinforcement learning etc.) make comprehensive analysis of AI integral for the wind industry.

Some previous studies have reviewed applications of \nomenclature{$ML$}{Machine learning} Machine learning (ML) techniques for CBM and performance assessment/analysis towards data-driven decision making from various perspectives. Stetco et al. \cite{STETCO2019620} reviewed 144 papers post-2011, presenting an overview of the challenges and potential of such techniques for classification and regression tasks in the wind industry. Wang et al. \cite{windcontrol_ai} have outlined the applications of AI towards optimising wind farm control systems for improved efficiency. Pliego-Marugán et al. \cite{ann_surveywind} reviewed 190 papers in the last five years, presenting the challenges and technological gaps in utilising artificial neural networks in time-series forecasting of certain parameters (e.g. wind speed and turbine power), with a perspective on fault diagnosis and prognosis. Maldonado-Correa et al. \cite{doi:10.1177/0309524X19891672} have reviewed 37 articles in applying AI techniques for short-term energy forecasting. In another recent study, Maldonado et al. \cite{recent_cbm} systematically reviewed 95 papers from the past three years, analysing present challenges in utilising CBM techniques (including AI) and the increasing growth in number of CBM-related publications in the wind energy sector across different journals. This included the analysis of some principal publication metrics such as impact factor, and segregating articles pertaining to different CBM techniques (such as signal processing methods, machine learning techniques etc.). This can be useful for a ready reference of historical literature in CBM for O\&M of wind turbines, such as the different techniques suitable for monitoring vital metrics e.g. wind speed, pitch angle etc., and predicting faults. The study does not however provide a more thorough analysis of AI in the wind industry (especially recent advances in deep learning, such as the application of recurrent neural networks and causal inference \cite{sun_sun_2018,torque_paper}) and its evolution over time. Most existing studies evidently focus on time-series forecasting of vital parameters in data-driven decision making, rather than predicting incipient faults and suggesting maintenance actions \cite{LEITE20181917,wcci_paper}. 

Despite playing an important role in summarising applications of AI towards data-driven decision making in wind turbines, these studies lack systematic and comprehensive analysis of the changing trends and insights from a broader perspective beyond the narrow focus on either fault prediction or power forecasting, which is especially important with the ongoing rapid growth of academic publications in this area. Additionally, there is little attention towards very recent developments, especially in applying e.g. natural language generation and reinforcement learning techniques for decision support in the wind industry and associated challenges and opportunities. Such developments demand systematic analysis to show the bigger picture, by analysing existing literature to identify changes in research trends with time, key themes in data-driven O\&M, and shifts in boundaries of applying AI towards CBM and performance assessment/analysis. 

\begin{table*}[t]
\small
\renewcommand{\arraystretch}{1.5}
\caption{Summary of logical queries and retrieved records for the scientometric analysis.\label{table_queries}}
\begin{tabular}{|p{4cm}|p{7cm}|p{2cm}|}
\hline
\textbf{Domain} & \textbf{Logical Query (Inclusion criteria)} & \textbf{No. of Retrieved Records after Filtering}\\ \hline
All papers relating to CBM &  ("wind turbine" AND "condition monitoring") OR ("wind energy" AND "condition monitoring") & 734  \\ \hline
Papers specifically relating to AI for O\&M (Includes publications pertaining to CBM and performance assessment/analysis) &  ("wind turbine" AND "machine learning") OR ("wind energy" AND "machine learning") OR ("wind turbine" AND "deep learning") OR ("wind energy" AND "deep learning") OR ("wind turbine" AND "artificial intelligence") OR ("wind energy" AND "artificial intelligence") OR ("wind turbine" AND "AI") OR ("wind energy" AND "AI") & 422 \\ \hline
\end{tabular}
\end{table*}

\begin{figure*}[!h]
\centerline{\includegraphics[width=0.8\textwidth]{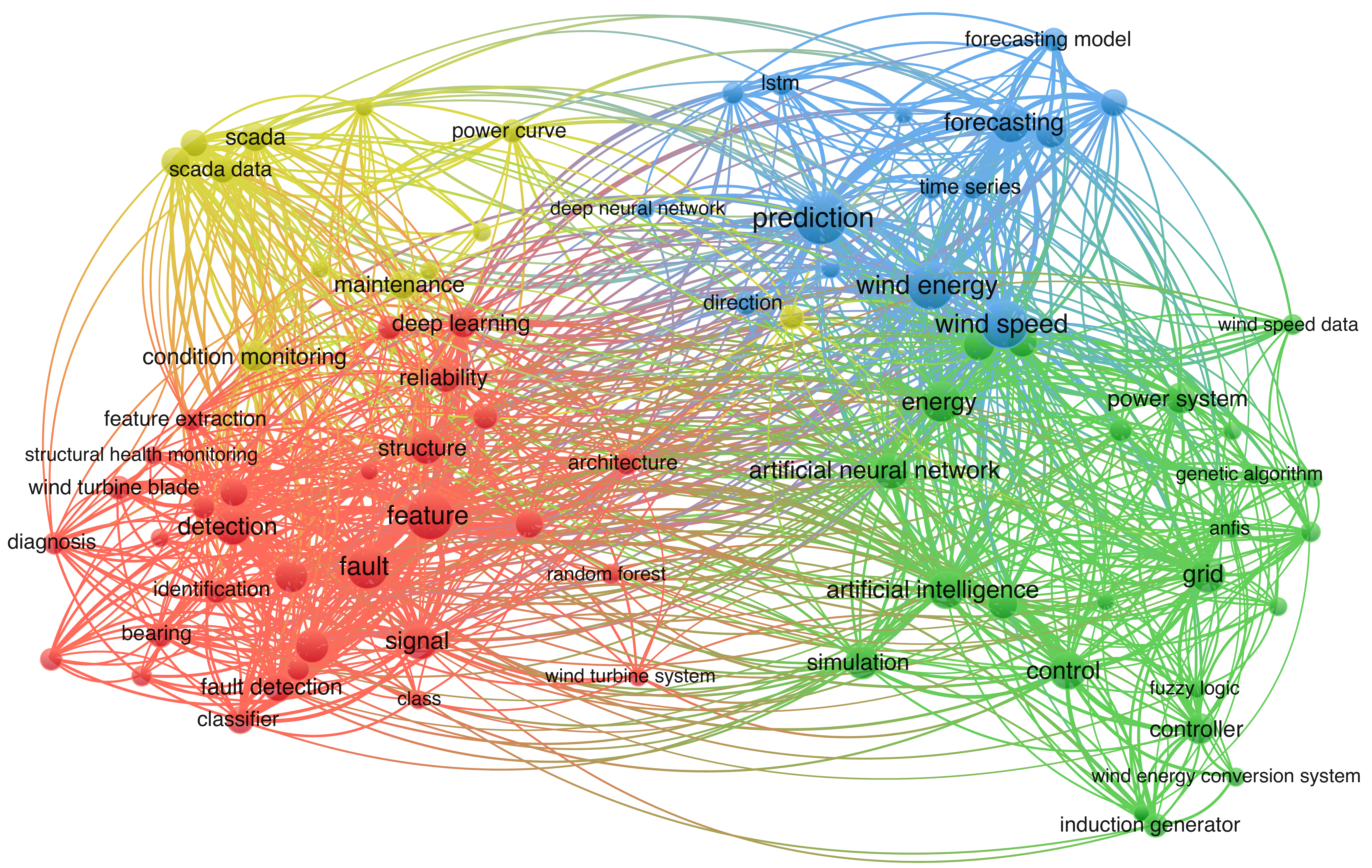}}
\caption{Network visualisation showcasing the complexity of data-driven decision making in wind industry, based on  publications from 2010-2020. The different colours indicate specific clusters to which each keyword belongs, with strong association between terms from the same cluster. \label{nvis_cbm}}
\end{figure*}

Bibliometrics is a family of statistical techniques commonly utilised in library and information sciences for analysing and discovering patterns in publications. For analysis specific to scientific literature, the sub-field of Bibliometrics called Scientometrics has gained prominence. While various domains such as medicine and finance \cite{bib_med,bib_finance} have reaped the benefits of Bibliometrics (and Scientometrics), the wind industry has seen limited application of such techniques. Few studies perform scientometric assessment of the wind energy domain as a whole, such as Kanagavel et al. \cite{ramasamy2012}, Ye et al. \cite{doi:10.1080/1331677X.2020.1734853} and Mohanathan et al. \cite{mohanathan}, who analyse the growth in research productivity based on the rise in number of publications to show the increased uptake of wind turbines in recent times. However, these studies do not focus on the assessment of data-driven decision making models for the wind energy sector, and cannot provide thematic descriptions and analysis of the research trends and patterns for AI in CBM and performance assessment/analysis. The domain of O\&M for the wind industry is highly complex, with widely varying methodologies, data utilised and tasks performed. This is enunciated in Figure ~\ref{nvis_cbm}, through network visualisation of data-driven decision making publications over the last decade.

In this paper, we aim to provide comprehensive evaluation of the applications of AI for data-driven decision making (mainly focusing on CBM but also including performance assessment/analysis wherever relevant to O\&M) in the wind industry by harnessing statistical computing for scientometrics. To this end, we utilise Bibliometrix \cite{ARIA2017959} (a statistical computing technique in R) along with CiteSpace  \cite{doi:10.1002/asi.20317} (a Java program for analysing science literature) and VOSviewer \cite{eck_waltman_2009} (a software application for visualising bibliometric networks) to derive insights into the conceptual and thematic structure of data-driven decision making for wind turbines. We also utilise Datawrapper \cite{lorenz_aisch_kokkelink_2012} (a web application for data analytics) to develop insightful visualisations. The proposed technique can provide novel insights on the evolution of AI for the wind industry, establishing a knowledge taxonomy for research themes in this area. 

This study, to the best of our knowledge, is the first in the wind energy domain to apply statistical computing towards systematic analysis of historical literature for data-driven decision making in O\&M of wind turbines. The key contributions of this article are:-

\begin{itemize}
    \item A data-mining approach is applied for bibliographic analysis, utilising state-of-art statistical computing for scientific mapping. This can help reduce bias and subjectivity in varying perspectives on AI in the wind industry, providing a comprehensive thematic analysis.
    \item The insights derived from this study can provide an understanding of the conceptual developments, emerging trends and thematic areas, challenges and opportunities in applying AI for data-driven decision making in the wind industry.
    \item We perform an extensive analysis of the past and present of AI in data-driven decision making for wind turbines by reviewing 422 research publications in this domain from the last decade, and additionally provide insights into the future based on identified successes and failures. This analysis is scalable and can be extended as future publications emerge. Our data used for this study is publicly available \footnote{Data utilised for scientometric review: https://github.com/joyjitchatterjee/ScientometricReview-AI}, and can help future researchers build upon the analysis in this study.
\end{itemize}

By tracing the evolution of data-driven decision making techniques for the wind industry through scientometrics, we show the role which AI plays at present, and the rapidly evolving growth in application of AI techniques for O\&M. We also provide a perspective into the future, including key issues such as lack of transparency and interpretability in AI models, deployment of models for real-time decision support, and data availability and quality, which presently hold the field back in adopting data-driven decision making, and suggest possible ways to overcome these challenges. The paper is organised as follows: Section ~\ref{data_collect} describes the data utilised for our analysis. Section ~\ref{science_mapping} discusses the evolution of data-driven decision making techniques in the wind industry, from signal processing in the past to rise of AI at present. In Section ~\ref{perspective}, we outline the challenges and opportunities presently faced by the wind industry in adopting AI techniques, along with possible strategies to overcome these issues. A discussion on the roadmap with likely major focus areas for the wind industry towards adopting AI is provided in Section ~\ref{discussion}. Finally, Section ~\ref{conc} concludes the paper and describes the path for future.

\section{Data Collection}\label{data_collect}
As the primary analysis step, the Web of Science database \footnote{Web of Science Database: https://apps.webofknowledge.com/ \label{wos_link}} was queried to retrieve all papers relating to CBM of wind turbines in the last decade (2010 to present), leading to an initial record of 818 publications. Of this, we eliminated the review articles to ensure robustness in identifying specific topics rather than broad perspectives. The search was further refined by only retaining papers from journals and conference proceedings in English language. There were some non-relevant papers which were manually removed from consideration. Finally, we arrived at a total of 734 records pertaining to CBM in wind turbines consisting of a variety of techniques (e.g. signal processing, vibration analysis etc.) besides using AI.

As our key motivation was in identifying papers applying AI to data from wind turbines, we separated the publications utilising AI techniques for CBM as well as performance assessment/analysis by specifying an additional logical criterion. This led us to a total of 422 records, which consist of mainly papers pertaining to CBM but also include some instances of performance assessment/analysis tasks in O\&M. Table ~\ref{table_queries} summarises the logical criteria utilised for retrieving the historical literature. All retrieved records were exported as plain text files, which can later be utilised for scientific mapping, as described in the following sections. 

\section{Science Mapping Analysis}\label{science_mapping}
\subsection{The Past: Prevalence of Signal Processing \& Vibration Analysis}
For our systematic review, the period from 2010 to 2015 is considered as the past \footnote{Note that while our scientometric analysis would mainly focus on utilising publications in this period, for the sake of thoroughness in analysing the past, we would also mention few notable studies outside this period in our reviews which are relevant to the scope of this paper}. This is based on careful consideration and the mostly prevalent consensus that any period beyond the last 5 years falls outside scope of current literature \cite{bloomberg_volpe_2016}. Analysis of past literature is important for deriving insights in predicting future trends based on identified strengths and weaknesses of existing techniques. In this period, interestingly, we observed that the majority of influential papers focused on CBM pertaining to health monitoring of turbines and their sub-components, with negligible focus on performance assessment/analysis (e.g. for wind power forecasting). Therefore, we will direct our discussion and analysis of the past mainly on CBM for a comprehensive analysis, but wherever relevant, we would also include the few studies which focused on performance assessment/analysis.
\begin{figure*}[!h]
\centerline{\includegraphics[width=0.7\textwidth]{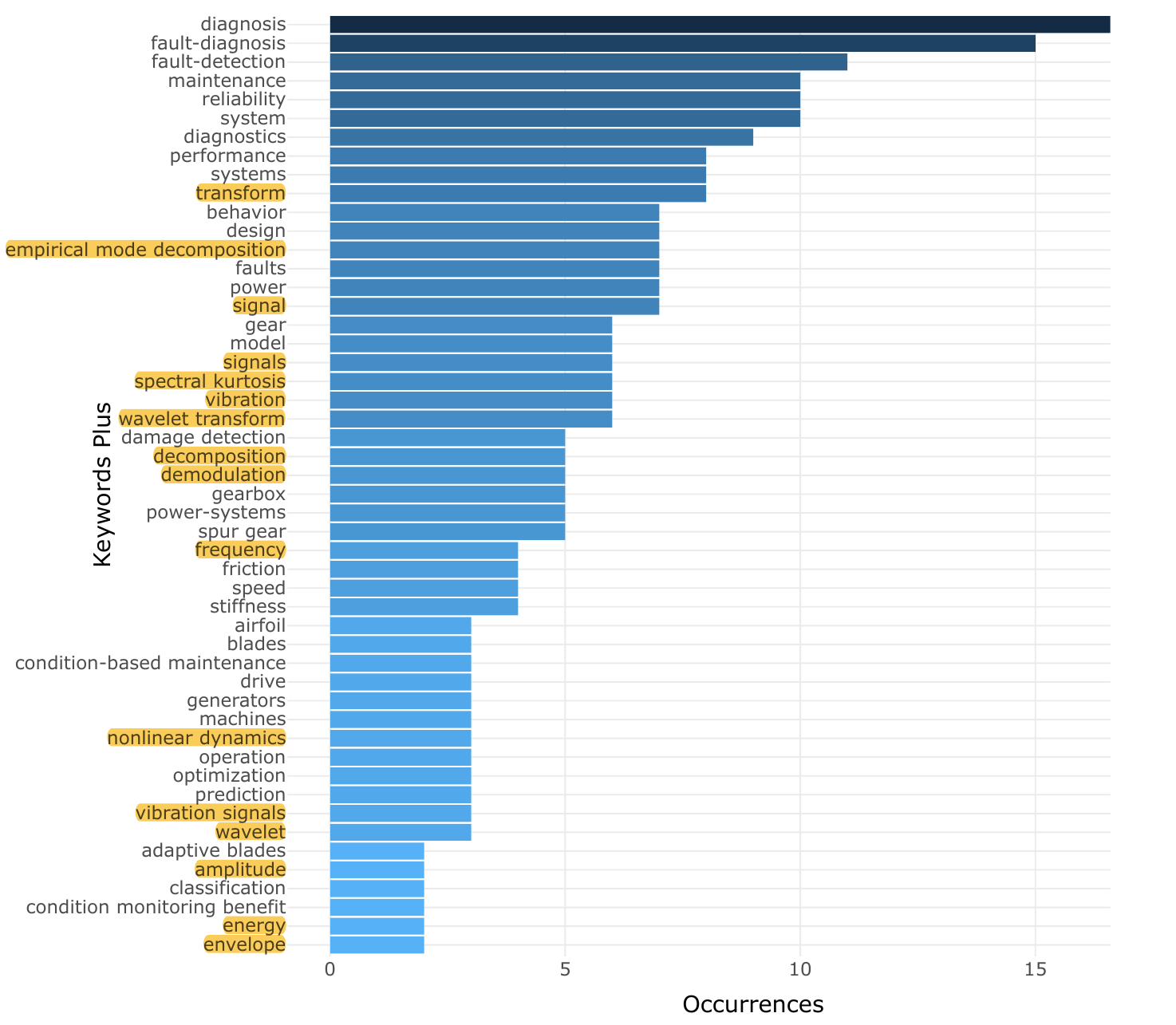}}
\caption{Frequency of top-50 words in CBM publications for wind turbines in the past. Words relevant to the signal processing domain are highlighted. \label{freq_cbm_past}}
\end{figure*}

From our 734 total records on CBM, 311 records belonged to this period of analysis. Figure ~\ref{freq_cbm_past} outlines the frequency of the most common words in CBM-based publications during this period. Note that these words were automatically determined based on the \textit{Keywords Plus} metric \footnote{Keywords Plus: http://interest.science.thomsonreuters.com/content/WOKUserTips-201010-IN \label{keyword_plus}} \cite{keywordplus_paper} which, besides using the author provided keywords in the papers also considers the titles and abstracts of the paper alongside the references and highlights relevant content which may have potentially been overlooked based on keywords listed by the authors, and can thereby help expand the search for analysing all relevant papers in this period of time. From all identified keywords, as not all keywords were relevant so signal processing/vibration analysis based on their semantic definition e.g. classification, we manually highlighted those which were relevant to signal processing/vibration analysis based on their occurrence in relevant domain-specific publications. For instance, we considered \textit{frequency} to be a part of the signal processing domain based on its occurrence in the abstract of \cite{RePEc:eee:renene:v:47:y:2012:i:c:p:112-126}, wherein, the authors performed vibration signal processing, and a similar approach was adopted for highlighting other keywords based on relevant publications. As can be seen, multiple keywords pertain to the domain of signal processing (e.g. wavelet transform, empirical mode decomposition etc.) and vibration analysis (e.g. vibration signals, amplitude etc.). To further discuss the prevalence of signal processing and vibration analysis for CBM, we will focus our discussion on some of the most relevant papers during this period.

Vibration analysis has been popular for fault diagnosis in turbine structures and sub-components, especially in the rotational parts \cite{LIU2013954}. Zimroz et al. \cite{10.1007/978-3-642-28768-8_52} utilised data with RMS of vibration acceleration signal and generator power obtained through a professional monitoring system to perform vibration analysis and identify abnormal behaviour of turbine bearings under non-stationary load/speed conditions, and decomposed the data into multiple sub-ranges of loads to facilitate CBM. Additionally, they utilised these parameters as features to identify any deviations in operational behaviour through statistical processing. In a similar vein of work, Liu \cite{LIU2013954} proposed the statistical estimation of total wind force prevalent in the turbine's blade-cabin-tower system by applying physics-based techniques for vibration analysis which was described as a mathematical framework, although they did not utilise any data to demonstrate the applicability of the method. Their paper focuses on deriving kinetic equations and natural frequency of the coupling system using Fourier transform and other probabilistic techniques. Their technique can help in identifying random wind vibrations and its effects based on the analysed spectrum, facilitating fault diagnosis.

Multiple studies \cite{5673197,2011JSV...330.3766Y,RePEc:eee:renene:v:47:y:2012:i:c:p:112-126} have utilised vibration signals and applied \nomenclature{$EMD$}{Empirical Mode Decomposition} Empirical Mode Decomposition (EMD) for detecting incipient faults in the turbine's mechanical and electrical sub-components, by decomposing these signals into \nomenclature{$IMF$}{Intrinsic mode functions} intrinsic mode functions (IMFs). Feng et al. \cite{RePEc:eee:renene:v:47:y:2012:i:c:p:112-126} for instance, proposed demodulation analysis of planetary gearbox vibration signals, by accounting for the IMFs produced using an ensemble EMD method. Comparing the amplitude and instantaneous frequency of the demodulated signal envelope's Fourier spectra with the ideal theoretical values can help detect abnormalities in the gearbox operation, including wear and chipping faults. 
Some variations in this technique have also been applied for short-term forecasting of wind speed, wind power etc. Zheng et al. \cite{kalman_pf}, for instance, utilised historical wind farm data with wind speed, wind direction and turbine power outputs. The authors utilised EMD for decomposition of wind power into multiple IMFs and one residue, along with \nomenclature{$RBFNN$}{Radial basis function neural networks} radial basis function neural networks (RBFNN) as a prediction model. The paper also utilised statistical control algorithms like Kalman filtering for elimination of noise . 

Some studies have performed signal processing of vibration signals in the frequency domain by estimating \nomenclature{$SK$}{Spectral kurtosis} spectral kurtosis (SK) for CBM. In this technique, the kurtogram can help determine non-stationarities within the signals, potentially contributing to any defect in the turbine's sub-components. The SK technique plays an integral role in extending the general concept of kurtosis (which is a global value) to a function of frequency which is capable of indicating impulsiveness in signals \cite{7505124}. In a notable study in this area, Saidi et al. \cite{7505124} proposed a squared envelope technique based on SK for diagnosing skidding in high-speed shaft bearings through degradation analysis for performing run-to-failure testing. The paper utilises real-world vibration data from high-speed shaft bearings, and demonstrates that the maximum value of the SK can serve as an indication of severity of the prevailing damage, while the square root of the SK can help extract transients in the signal. The authors performed experimental runs across different cases pertaining to normal zone, degradation zone and failure zone, and their study shows the immensely powerful role SK can play in diagnosing faults in critical parts of rotating sub-components in turbines.

As vibration signals can often be subject to high background noise, Jia et al. \cite{jia_lei_shan_lin_2015} have proposed an improvement in the conventional SK technique for fault diagnosis in the rolling-element bearings, applying \nomenclature{$MCKD$}{Maximum correlated kurtosis deconvolution} Maximum correlated kurtosis deconvolution (MCKD), which can help clarify periodic fault transients in noisy signals, making them more suitable for diagnosis of incipient failures. Despite their simplicity and no requirements for historical failure data to develop the fault-prediction model, these studies do not present any performance metrics (e.g. accuracy) in identifying faults, and their predictions cannot be validated, making them lack robustness. They also cannot be utilised for estimating vital O\&M parameters of the turbine and its sub-components, such as \nomenclature{$RUL$}{Remaining Useful Life} Remaining Useful Life (RUL), \nomenclature{$MTTF$}{Mean Time to Failure} Mean Time to Failure (MTTF) etc.


Besides the conventional Fourier transform, some studies have applied wavelet transform for CBM by analysing vibration signals in the frequency domain. Guo et al.  \cite{5554606} utilised \nomenclature{$DWT$}{Discrete Wavelet Transform} Discrete Wavelet Transform (DWT) to identify gear faults using the vibration acceleration signal. The DWT can better characterise time-varying components of the signal and its energy distribution over traditional stationary signal processing methods, providing easier identification of faults. In another similar study, Yang and An \cite{doi:10.1155/2013/212836} proposed a hybrid approach combining Empirical Mode Decomposition (EMD) with wavelet transform. In this methodology, the wavelet transform is used to analyse the vibration signals, while the EMD contributes to better decomposition of the signal into IMF components, facilitating a more thorough prediction of the signal's instantaneous frequency as it can address aliasing in the signal resulting from interference caused by high-frequency components of the transformed signal. Despite their simplicity in analysing signals for detecting faults, techniques such as wavelet transform can be highly computationally intensive for fine-grained analysis compared to present AI algorithms, and often require careful considerations in choosing shifting, scaling and other parameters for any potential success \cite{wavelet_dis}. 

\begin{figure*}[h]
\centerline{\includegraphics[height=0.5\textheight]{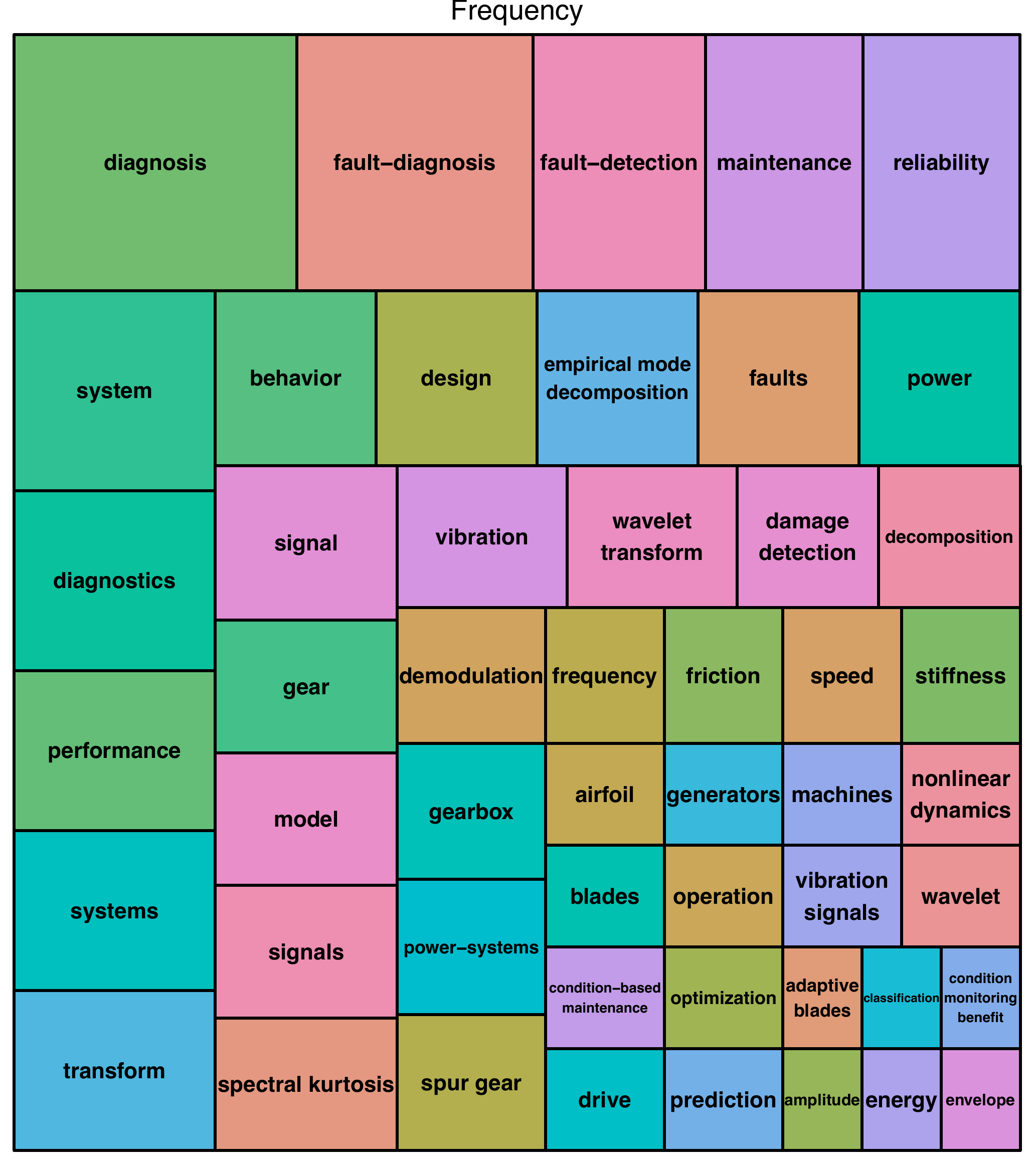}}
\caption{Treemap outlining the hierarchial composition of prevalent signal processing and vibration analysis techniques  \label{treemap_cbm_past}}
\end{figure*}

\begin{figure*}[h]
\centerline{\includegraphics[width=0.8\textwidth]{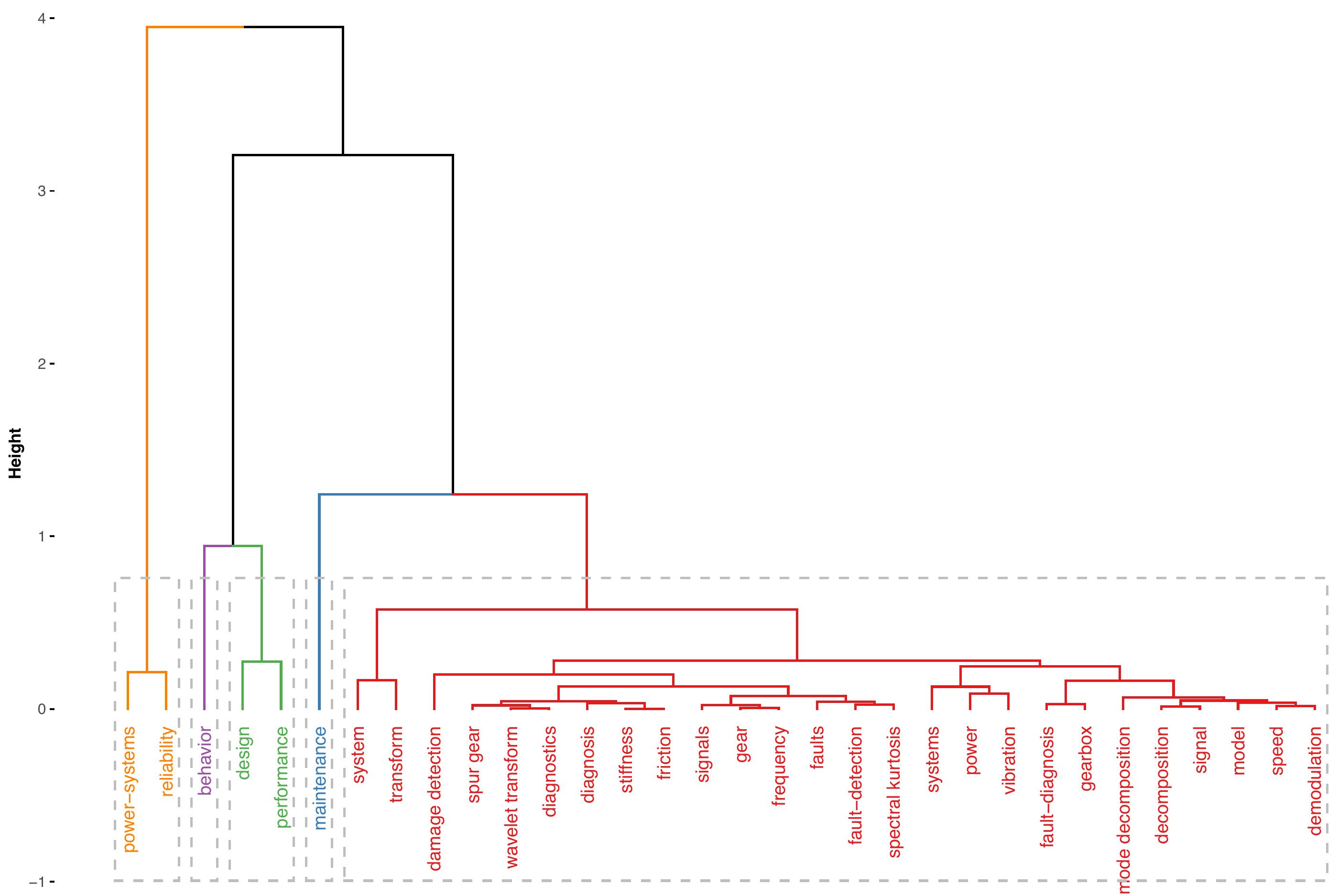}}
\caption{Dendogram depicting the clusters and context of applying CBM techniques in the past  \label{dendo_past}}
\end{figure*}

Figure ~\ref{treemap_cbm_past} depicts the treemap tracing the hierarchical composition of signal processing and vibration analysis methods used in the past. The treemap shows the combination of different possible keywords in this domain (e.g. \textit{empirical mode decomposition} co-occurred as a keyword alongside \textit{system}, \textit{design}, \textit{behaviour} etc., \textit{spectral kurtosis} co-occurred with keywords like \textit{transform}, \textit{spur gear}, \textit{drive} etc. in the majority of publications). This provides a low-level view in line with our reviews above. For instance, the treemap shows that empirical mode decomposition has been utilised in predicting faults, forecasting power output and designing turbine control systems. Spectral kurtosis, as another prevalent method has been associated with modelling vibration signals for turbine sub-components, especially the gearbox. Demodulation techniques have been commonly applied during signal processing of the vibration signals in O\&M. Similarly, other relationships outline key elements prevailing in the past for the application of signal processing and vibration analysis. For damage detection of turbine sub-components, as evident, the task has been performed with the angles of fault prediction, optimisation of operations and prevention of friction etc. Figure ~\ref{dendo_past} provides a more fine-grained view of the clusters of prevalent techniques and their common applications.

While the past has mostly seen applications of signal processing and vibration analysis techniques for CBM, a few important studies have demonstrated promising results in applying conventional AI techniques prevalent in the past for performance assessment and analysis of turbines. Clifton et al. \cite{Clifton_2013} utilised aerostructural simulations data for a turbine and applied regression trees to forecast turbine power output, accounting for wind speed, turbulence and shear, and their methodology has demonstrated success in forecasting turbine performance at new sites with simply the wind resource assessment data, which is generally available easily to turbine operators. Several other studies have also modelled wind turbine power outputs with conventional AI techniques such as time-series cluster analysis of power forecasts during periods of normal operations and anomaly \cite{Pravilovic_2014}, \nomenclature{$SVM$}{Support Vector Machine} Support Vector Machine (SVM) enhanced Markov models \cite{Yang_2015}, Gaussian processes and \nomenclature{$NWP$}{Numerical weather prediction} Numerical weather prediction (NWP) models \cite{Chen_2014} etc. There have been some attempts at applying hybrid models for power forecasting to achieve improved results. Soleymani et al. \cite{Soleymani_Mohammadi_Rezayi_Moghimai_2015}  for instance, utilised a real-world wind farm dataset and applied probabilistic approximation techniques in conjunction with conventional AI optimisation algorithms to develop a hybrid modified firefly algorithm which can provide forecasts of turbine power outputs, while also considering the prediction's confidence intervals. Moreover, such techniques are simple to apply and interpret, due to reliance on probabilistic and statistical inference in the prediction making process. However, as would be evident from the later discussions, these methods are generally significantly outperformed by more recent approaches, especially deep learning for time-series forecasting.

Interestingly, the past has seen very limited application of AI in predicting faults in turbines and its sub-components. 
This requires careful consideration of the trade-off in installing additional sensors for vibration analysis of signals, and utilising SCADA data, to ensure optimal results for data-driven decision making.

\begin{figure*}[!h]
\centerline{\includegraphics[width=0.7\textwidth]{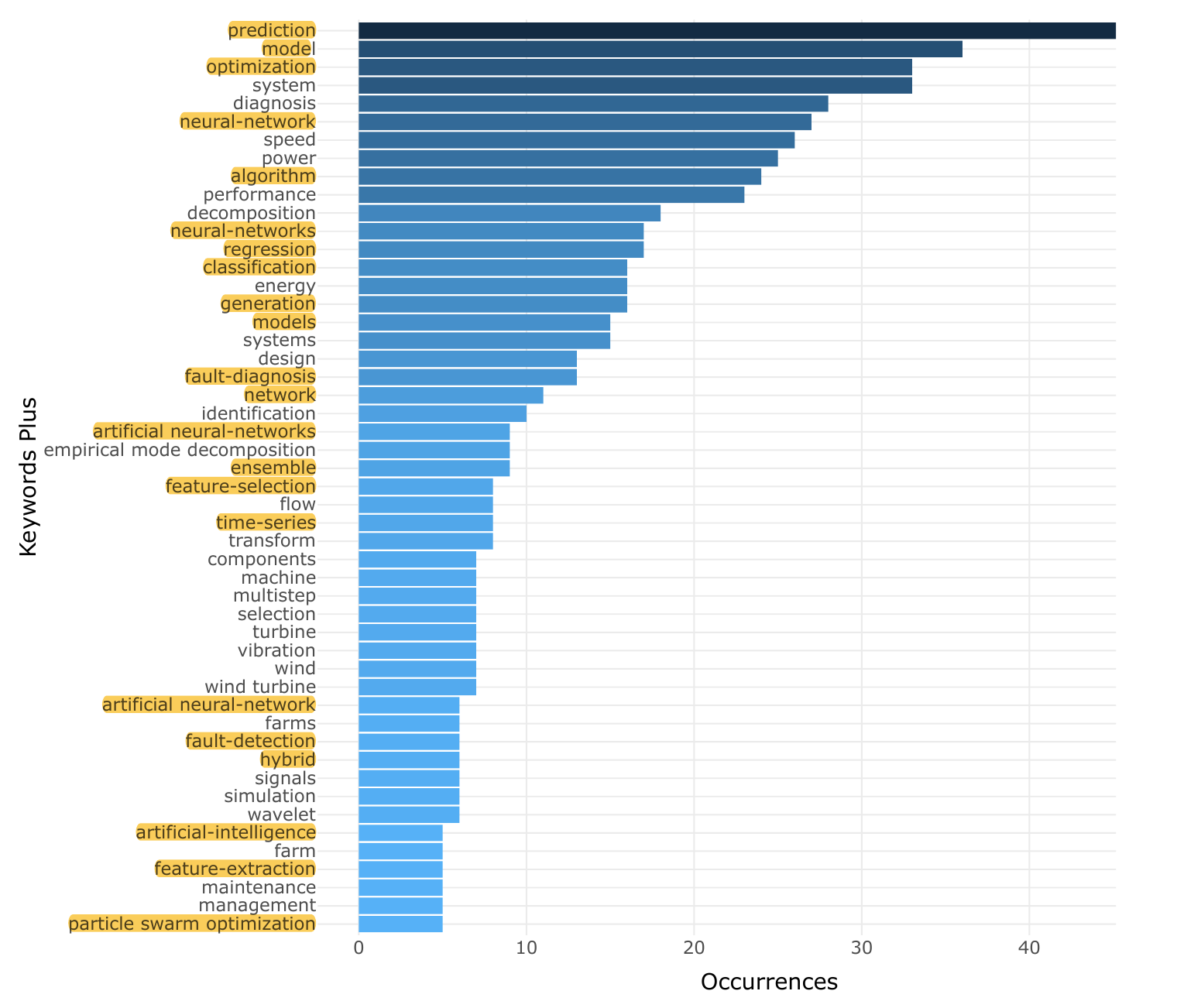}}
\caption{Frequency of top-50 words in CBM publications at present. Words relevant to AI techniques are highlighted. \label{freq_cbm_present}}
\end{figure*}

\subsection{The Present: Rise of AI in the Wind Industry}
In the past, numerical model-based and signal processing techniques have been the most popular for CBM, particularly leveraging vibration signals for health monitoring with promising results. There have been some instances of using AI with SCADA data for performance assessment/analysis, but these are rare in comparison to the focus on utilising vibration data for CBM. However, data-driven techniques, wherein, historical SCADA data is utilised to train AI algorithms are often much cheaper as well as simpler to use \cite{scada_windturbine_review}. Additionally, as present-day wind turbines are generally fitted with various sensors, taking periodic measurements and forming a part of the SCADA data as a standard, there are rarely any requirements to install measuring and instrumentation devices \cite{windtur_cbm}. 

\begin{figure*}[!h]
\centerline{\includegraphics[width=0.7\textwidth]{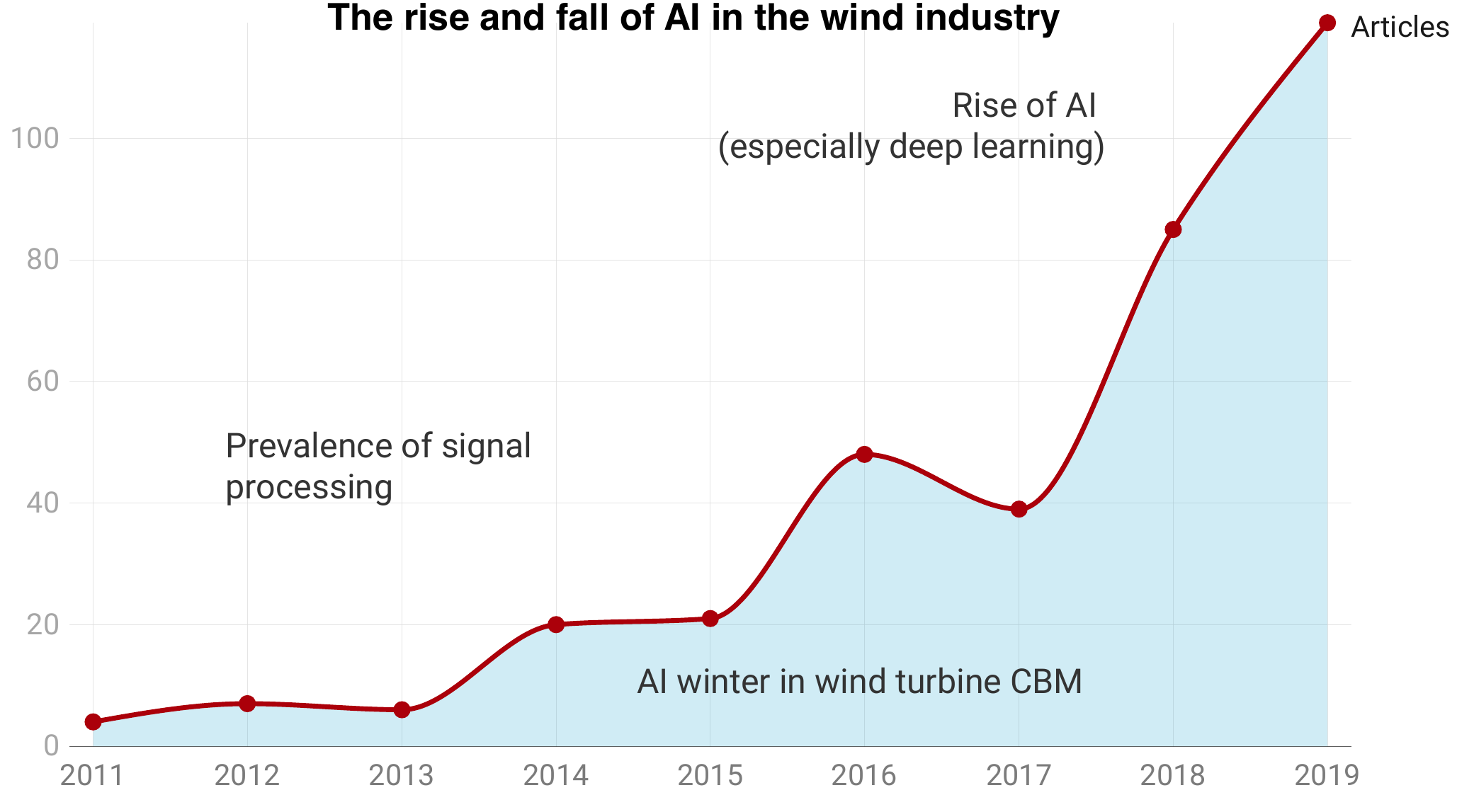}}
\caption{Evolution of AI in the wind industry. The significant interest in AI for CBM post-2017 can clearly be inferred. \label{rise_ai}}
\end{figure*}

\begin{figure*}[!h]
\centerline{\includegraphics[width=0.7\textwidth]{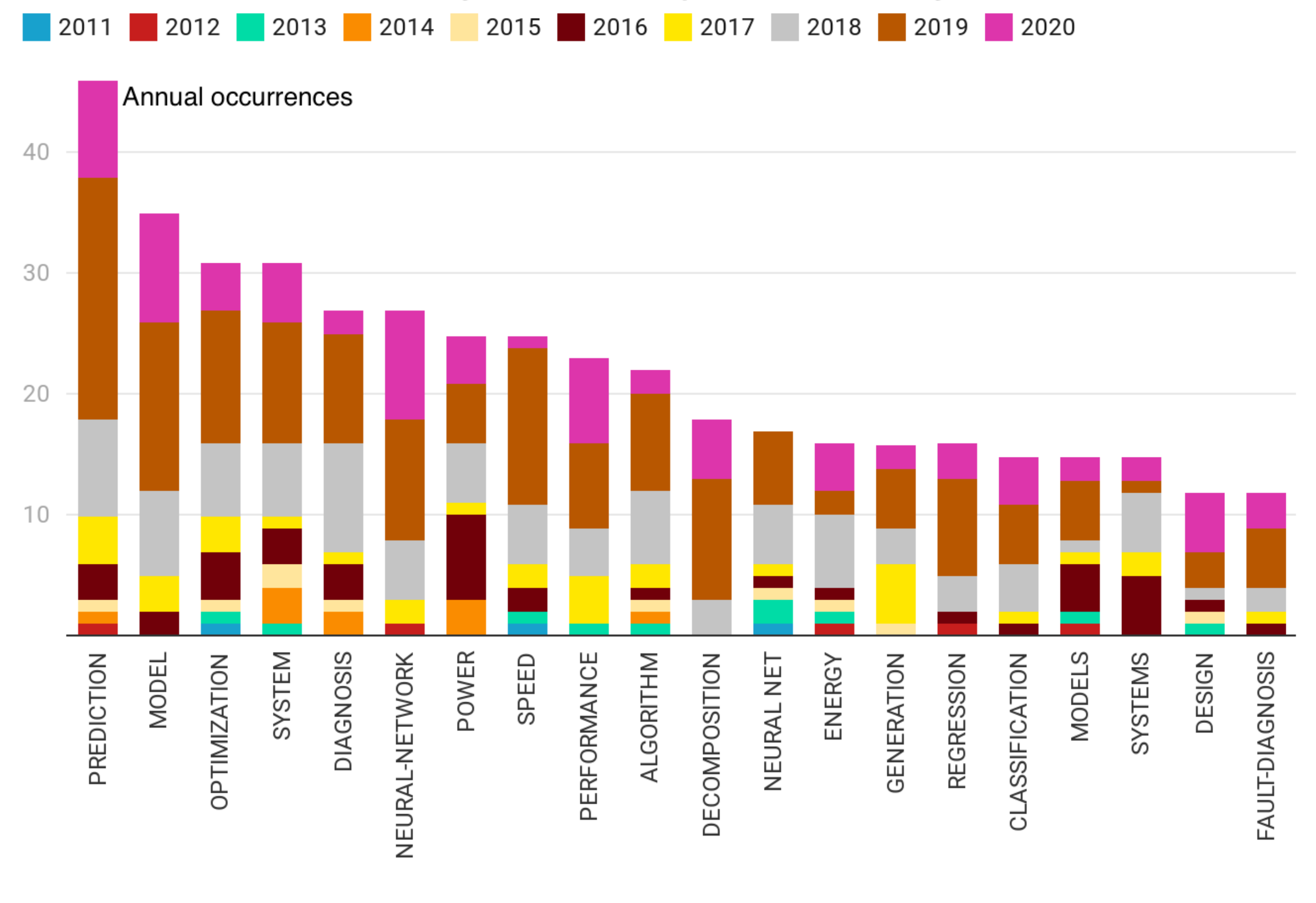}}
\caption{Word dynamics in CBM publications per year. The rise of AI algorithms (including neural networks) is clearly outlined. \label{worddynamics_cbm}}
\end{figure*}

\begin{figure*}[!h]
\centerline{\includegraphics[width=0.7\textwidth]{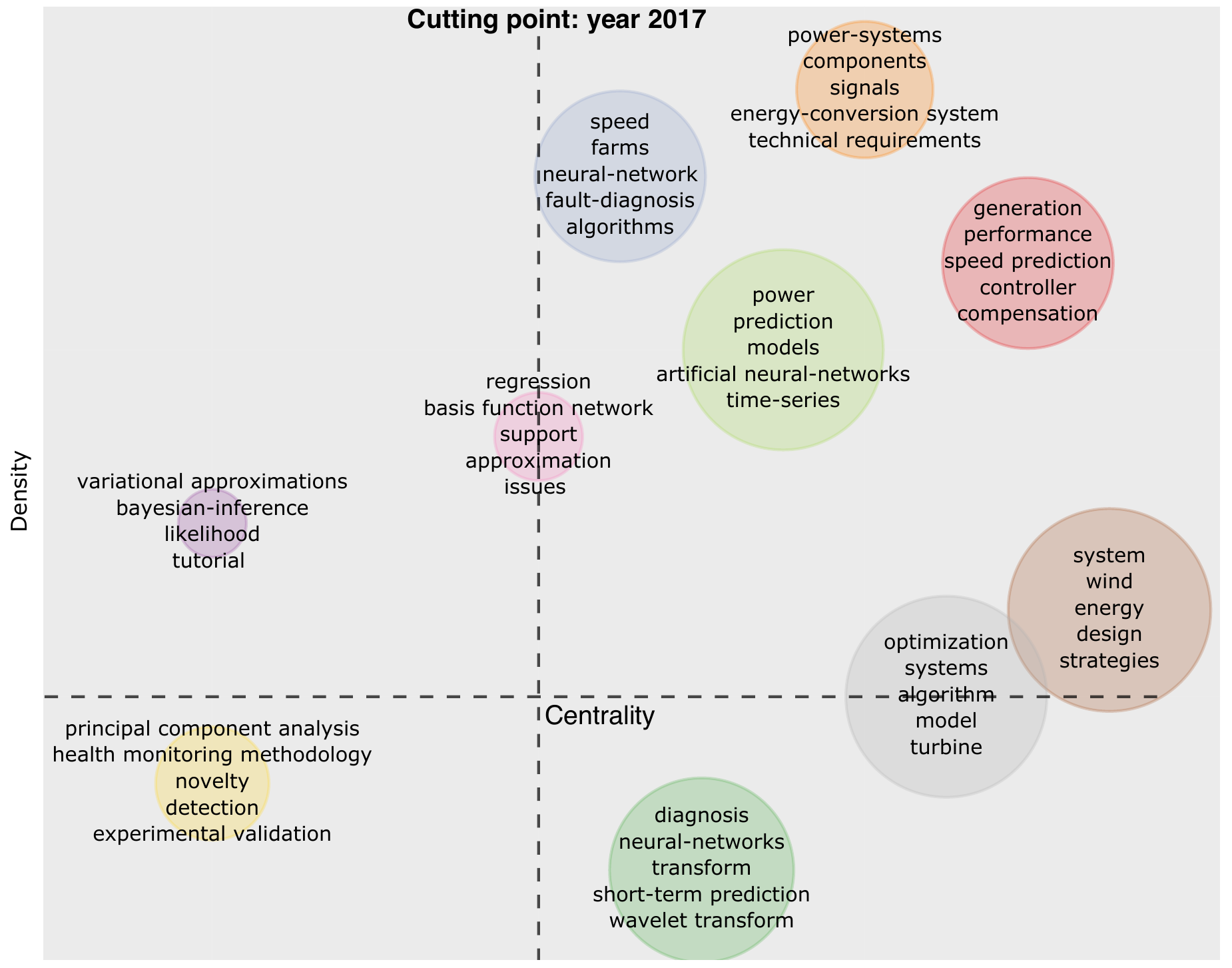}}
\caption{Thematic evolution of AI techniques over time. The cutting point here is the year 2017. \label{them_2017}}
\end{figure*}

Post the year 2015, the wind industry has seen a rapid growth in applications of Machine Learning (ML) models for data-driven decision support, particularly in utilising SCADA data for data-driven decision making. Interestingly, we observed that in this period, there was a significant rise in focus on performance assessment/analysis tasks for O\&M, while CBM techniques for health monitoring also continued to remain popular, although there was a significant shift from utilising vibration data to more attention on SCADA data. Figure ~\ref{worddynamics_cbm} shows the top 50 keywords for data-driven decision making publications at present, outlining the growing dominance of neural networks. This has directly contributed to an increase in the number of publications utilising AI for CBM and performance assessment of turbines, especially post-2017, as evident from the growing annual production of publications shown in Figure ~\ref{rise_ai}. Note that in 2017, the wind industry experienced an AI winter, with a significantly reduced interest in applying AI for O\&M, possibly due to reduced funding and/or resources, including quality data. In between 2015 to 2020, AI techniques have been utilised in a variety of aspects, for which the thematic evolution is depicted in Figure ~\ref{them_2017}. As evident, conventional ML techniques based on variational approximations, Bayesian inference and maximum likelihood estimation etc. have dominated the early evolution of AI in the wind industry. In particular, note that unsupervised linear transformation techniques such as principal component analysis (PCA) have been utilised for feature extraction and dimensionality reduction. There has also been an interest in novelty detection (e.g. abnormal events) in health monitoring of turbines. In the later half of this period (post 2017), we can observe the thematic rise of deep learning models, especially feedforward neural networks towards regression (e.g. predicting turbine power output time-series and short-term prediction of wind speed), fault diagnosis, optimisation of turbine operations and system design etc. Figure ~\ref{freq_cbm_present} outlines the most frequent words in AI publications applied to CBM at present. Such diverse applications of AI demand careful consideration and analysis of the present, which we discuss below specific to different categories of algorithms. 

\begin{figure*}[!h]
\centerline{\includegraphics[height=0.5\textheight]{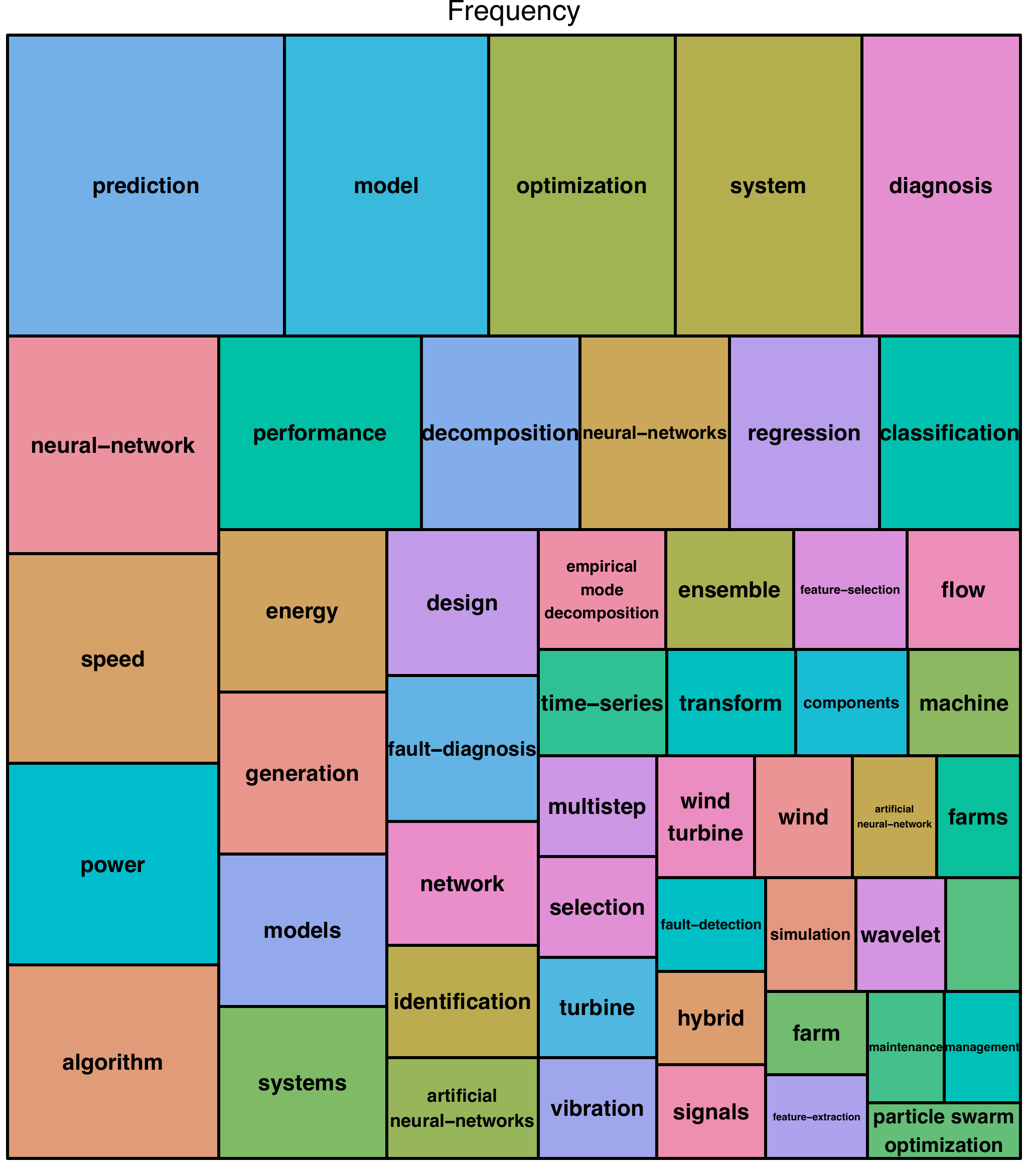}}
\caption{Treemap outlining the hierarchial composition of AI models utilised for data-driven decision making \label{treemap_ai}}
\end{figure*}

\begin{figure*}[!h]
\centerline{\includegraphics[width=0.8\textwidth]{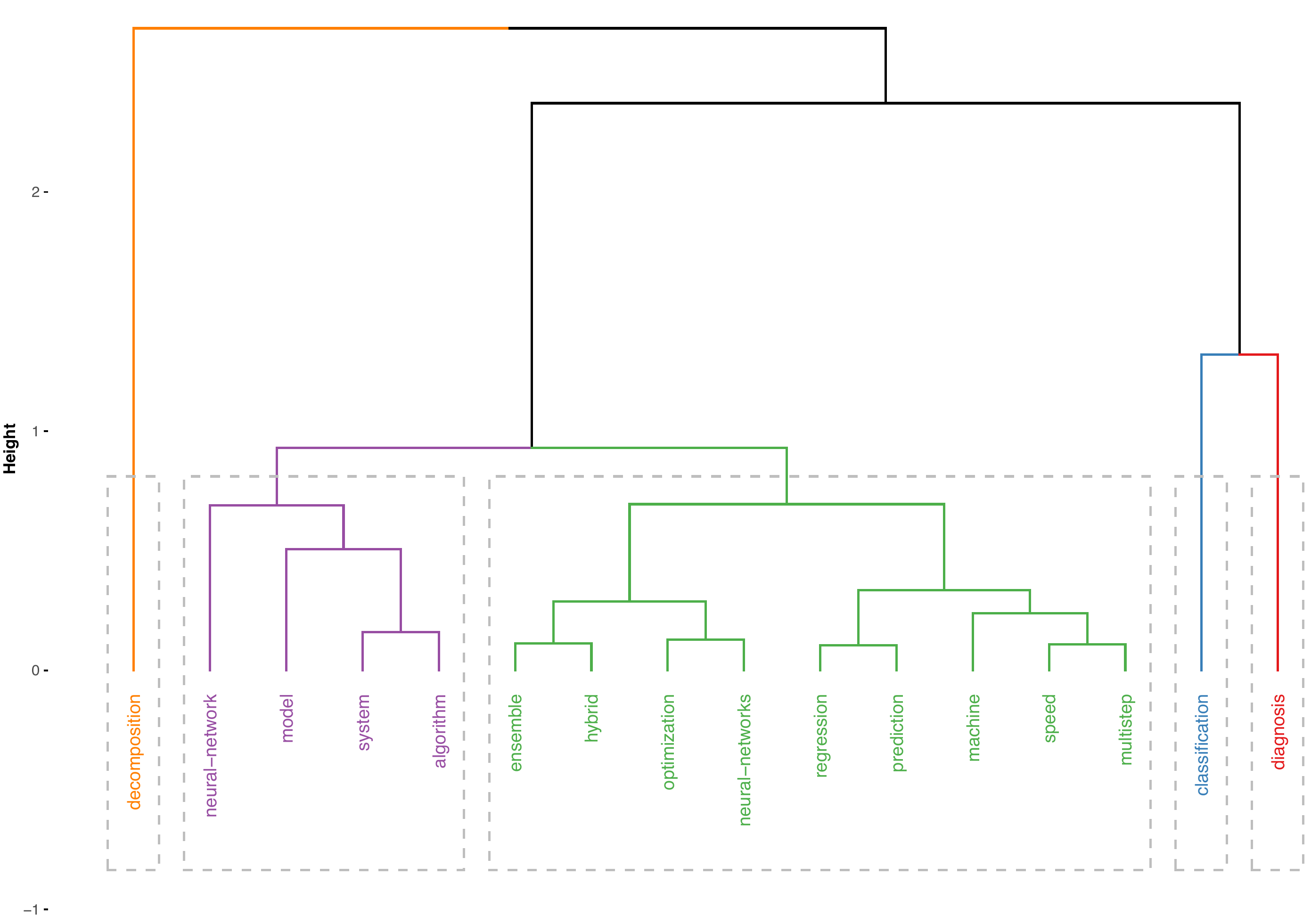}}
\caption{Dendogram depicting the clusters and context of applying AI techniques for O\&M \label{dendo_ai}}
\end{figure*}

Figure ~\ref{treemap_ai} depicts the treemap outlining hierarchical composition and the focus areas of various AI techniques for CBM and performance assessment of turbines in the wind industry, wherein, relevant keywords co-occur together (e.g. the keyword \textit{neural-network} co-occurred with \textit{performance}, \textit{decomposition}, \textit{regression}, \textit{classification} etc. in majority of the papers in this period of time). Different tasks e.g. classification and regression with applications in fault diagnosis, maintenance, power forecasting etc. are also clearly visible. Interestingly, there are still mentions of terms prevalent in signal processing such as wavelet transform and empirical mode decomposition, outlining that the wind industry has not foregone these techniques, but their use has continued to complement many modern AI techniques for data-driven decision making. Figure ~\ref{dendo_ai} shows the clusters and context of applying AI techniques for wind turbine O\&M, clearly outlining the dominating role of classification techniques for fault diagnosis and regression techniques for predicting vital turbine operational parameters in multi-step time series forecasting, utilised in conjunction with conventional decomposition techniques in signal processing. 

\paragraph{\textbf{Regression techniques in CBM and performance assessment}}
The simplest form of ML algorithms which have been utilised in the wind energy sector span the family of regression techniques, wherein, the continuous values of vital parameters are predicted over time. 
The time-varying nature of the SCADA data makes it extremely suitable for applying regression techniques in a supervised learning environment for predicting target values, such as power output, wind speed, wind direction etc. in new, unseen data.

Interestingly, while signal processing techniques have dominated CBM and performance assessment of turbines pre-2015, there have been some studies exhibiting early applications of conventional ML techniques which are worthy of mention. Clifton et al. \cite{paper4} utilised data from aero-structural simulations of a 1.5 MW turbine to apply regression trees in predicting turbine power output accounting for wind speed, turbulence intensity and shear expected. The technique was found to be significantly more accurate compared to conventional curve fitting (using power curves) in predicting the power output. Additionally, the paper demonstrated that regression tree models can further be applied to new turbine test data for predicting the performance of the turbine at a new site without requiring additional training data. In a similar work by Clifton et al. \cite{wind_mountainpass}, the authors utilised decision trees coupled with regression for evaluating and predicting turbine performance in a mountain pass region in response to the pass wind time series. Some studies have also applied support vector regression techniques \cite{paper2} for predicting the power output. Yang et al. \cite{realtime_reconstruction} have utilised support vector regression to develop a reconstruction-based machine learning model for real-time fault detection in turbine sub-components, identifying anomalies in SCADA features based on the residual error between these signals. While such approaches for CBM are simple to apply and interpret, they do not leverage historical turbine data for decision making. Also, mathematically modelling reconstruction models is computationally expensive, which defeats the purpose of scalable real time predictions, which should require minimal computational resources. 

In another area of CBM utilising regression techniques, Park et al. \cite{multivariate_windfield} have previously utilised \nomenclature{$GMM$}{Gaussian mixture models} Gaussian mixture models (GMM) and \nomenclature{$GDA$}{Gaussian discriminative analysis models} Gaussian discriminative analysis models (GDA) for structural health monitoring of turbines, with wind field characteristics data (such as wind speed, direction, turbulence, profile etc.). Their study shows that such approaches are  promising for load response prediction, and can be extended to new turbine sites, accounting for the openly available wind resource assessment data.

Post-2015, there has been a significant rise in the application of ML techniques for regression tasks. Especially, there has been a major move towards deep learning techniques. In one of the early works in this period, Du et al. \cite{paper5} proposed an anomaly detection technique, using the Pearson correlation coefficient for parameter selection in modelling the wind turbine's behaviour and self-organising map for dimensionality reduction of SCADA features. The paper utilises the end predictions to map the power outputs to the ideal power curve of the turbine, and thereby identify potential faults based on points that fall off the curve. Despite being promising and simple to apply, the proposed technique relies on power curves for the final anomaly prediction process, making it less competent compared to other AI algorithms which can directly generate predictions based on vital SCADA parameters \cite{outlook_windeurope}. In another study, Morshedizadeh \cite{paper3} focused on wind turbine power production based on the historical turbine performance data, demonstrating that a combination of a dynamic \nomenclature{$MLP$}{Multilayer perceptron} Multilayer perceptron (MLP) model and the \nomenclature{$ANFIS$}{Adaptive neuro fuzzy inference system} Adaptive neuro fuzzy inference system (ANFIS) can help predict turbine power outputs optimally, although such techniques have in the last few years been outperformed by more sophisticated algorithms, especially utilising deep learning.

In the move towards more sophisticated AI models, deep learning techniques have been utilised for predicting vital turbine operational parameters, which can help in performance assessment. Quereshi et al. \cite{QURESHI2017742} utilised operational data from a wind farm to develop an ensemble approach, combining deep auto-encoders (base-regressor) with Deep Belief Networks (meta-regressor) towards wind power prediction using meteorological features. The paper demonstrates that such hybrid ensemble approaches can significantly outperform conventional regression techniques. Additionally, the paper shows that the proposed model can facilitate transfer learning, providing power output predictions in the lack of additional training data. There has been a recent interest in utilising \nomenclature{$RNN$}{Recurrent neural networks} Recurrent neural networks (RNNs), especially \nomenclature{$LSTM$}{Long short-term memory networks} Long short-term memory networks (LSTMs) \cite{hochreiter_1998,hochreiter_schmidhuber_1997} for time-series SCADA features and meteorological parameters. Unlike conventional \nomenclature{$ANN$}{Artificial neural networks} Artificial neural networks (ANNs) \cite{ann_paper}, RNNs can account for past temporal information, making it competent for processing data with sequential nature, as evident in the wind industry. In an early development in this area, Kulkarni et al. \cite{doi:10.1177/0954406218797972} utilised LSTMs for long-term forecasting of wind speed at a farm site, wherein, the predictions were finally used to perform fatigue analysis of a 5MW wind turbine blade. The approach shows that RNNs are highly promising in facilitating dynamic wind load calculation. There have been similar applications of LSTMs for time-series forecasting of turbine power output such as by Zhu et al. \cite{st_windpower} and Liu et al. \cite{liu_guan_hou_han_liu_sun_zheng_2019}, demonstrating high forecasting accuracy of LSTMs for short-term wind power predictions, generally outperforming ANNs and SVMs. 

\paragraph{\textbf{Classification techniques in CBM and performance assessment}}
In machine learning, classification techniques serve an integral aspect of classifying/segregating two or more categorical variables e.g. fault types in different turbine sub-components, operations in different regions of the power curve etc. While some of the simplest classification techniques utilise e.g.  logistic regression \cite{log_reg}, which is simple to model and can make probabilistic predictions, thus making it possible to understand the most probable (or least probable) classes falling into a particular group. However, these  methods often perform poorly in modelling and classifying non-linear data, when there are multiple possible hyperplanes, which is generally the case with SCADA data being highly complex and non-linear. 

Classification techniques have widely been used over the last decade for analysing, diagnosing and predicting wind turbine faults. In an early application of classification techniques, Leahy et al. \cite{paper1} utilised SCADA data from the turbine and applied various classification algorithms towards filtering \& analysing faults and alarms, in conjunction with the turbine's power curve. The paper demonstrated that Support vector machines (SVM) were the best performing classifier model for predicting incipient faults in advance across multiple turbine sub-components. Despite showcasing the promise of AI for CBM of turbine sub-components, the paper lacks in using more sophisticated methods other than the binary classifiers for the multi-class classification problem in identifying specific faults. Also, no feature selection and dimensionality reduction was performed, thereby not accounting for that all SCADA features used may not be relevant when used with the SVM.

Some studies have utilised decision trees as an integral methodology for \nomenclature{$FDI$}{Fault detection and isolation} Fault detection and isolation (FDI). Si et al. \cite{data_randomforest} applied random forests with a combination of \nomenclature{$PCA$}{Principal component analysis} Principal component analysis (PCA) towards identifying faults in multiple sub-components of the turbine (such as pitch system, yaw drive, blades etc.) and determining dominant SCADA signals. The paper demonstrates that decision tree algorithms like random forest can measure and parameterise the importance of SCADA signals, which can be extremely useful in analysing predicted faults. Moreover, these algorithms can be fed with large datasets directly and are extremely efficient in terms of their training time, compared to more popular approaches like support vector machine (SVM), which despite their usual merits, suffer from the lack of capability to work on large datasets and preventing overfitting. In another notable study, Canizo et al. \cite{7998308} utilised big data frameworks such as Apache Kafka, Apache Spark, Apache Mesos and HDFS to develop a real-time predictive maintenance system consisting of an online fault tolerant monitoring agent, utilising the random forest algorithm as the learning model. They utilised SCADA data and historical logs of failure previously stored in the cloud server to provide predictions of turbine operational status every 10 minutes by performing SCADA data stream processing. This can be helpful in real-time decision support in the wind industry for assisting engineers \& technicians in O\&M activities. Also, this is likely the only paper in the area of utilising ML techniques for fault prediction to propose a complete solution right from model development and training to its deployment on a cloud server with a front-end dashboard. However, the paper mentions that the trained predictive models were not updated to adapt to the actual operational status of the turbines, and the authors propose to perform online updates in future work. 

In a similar vein of research on utilising classification techniques for CBM, Abdallah et al.  \cite{faultdiag_decisiontree} utilised \nomenclature{$CART$}{Classification and regression trees} Classification and regression trees (CART) to identify root causes of faults in the turbine. Specifically, the paper demonstrates that ensemble bagged trees are highly promising in identifying sequence of events that lead to a fault in particular sub-components of the turbine. Moreover, this approach also gives a brief description of the range of values for SCADA features leading to the fault (e.g. the gearbox oil temperature going beyond a particular range). The decision trees can be easily visualised by O\&M engineers, who can thereby take corrective actions. Despite the promise, the study does not provide any details regarding performance metrics (such as accuracy, prediction speed etc.) of the ensemble bagged tree classifier. Moreover, the paper fails to explain how incipient faults can actually be averted by using a decision tree classifier, given that SCADA data contains a series of measurements over time, and for temporal data, models like recurrent neural networks (RNN), \nomenclature{$ARIMA$}{Autoregressive integrated moving average} Autoregressive integrated moving average (ARIMA) etc. are generally better suited for reliable predictions. In another study in this area, Abdallah et al. \cite{abdallah_autodecision} proposed a conceptual framework with description of a hardware-software solution that utilises decision trees for real-time detection of faults. They also propose the interfacing of predictive decision tree model with a distributed data storage cloud server to perform analytics in real-time. The proposed framework can help provide autonomous decision support with simple and easy to interpret models like decision trees. However, utilising more sophisticated AI models (especially deep learners) for interfacing with such frameworks for real-time decision support would likely lead to significantly added complexity and challenges in deployment in the wind industry, which we believe is a critical issue that needs to be addressed in the near future.

\begin{figure*}[!h]
\centerline{\includegraphics[width=0.8\textwidth]{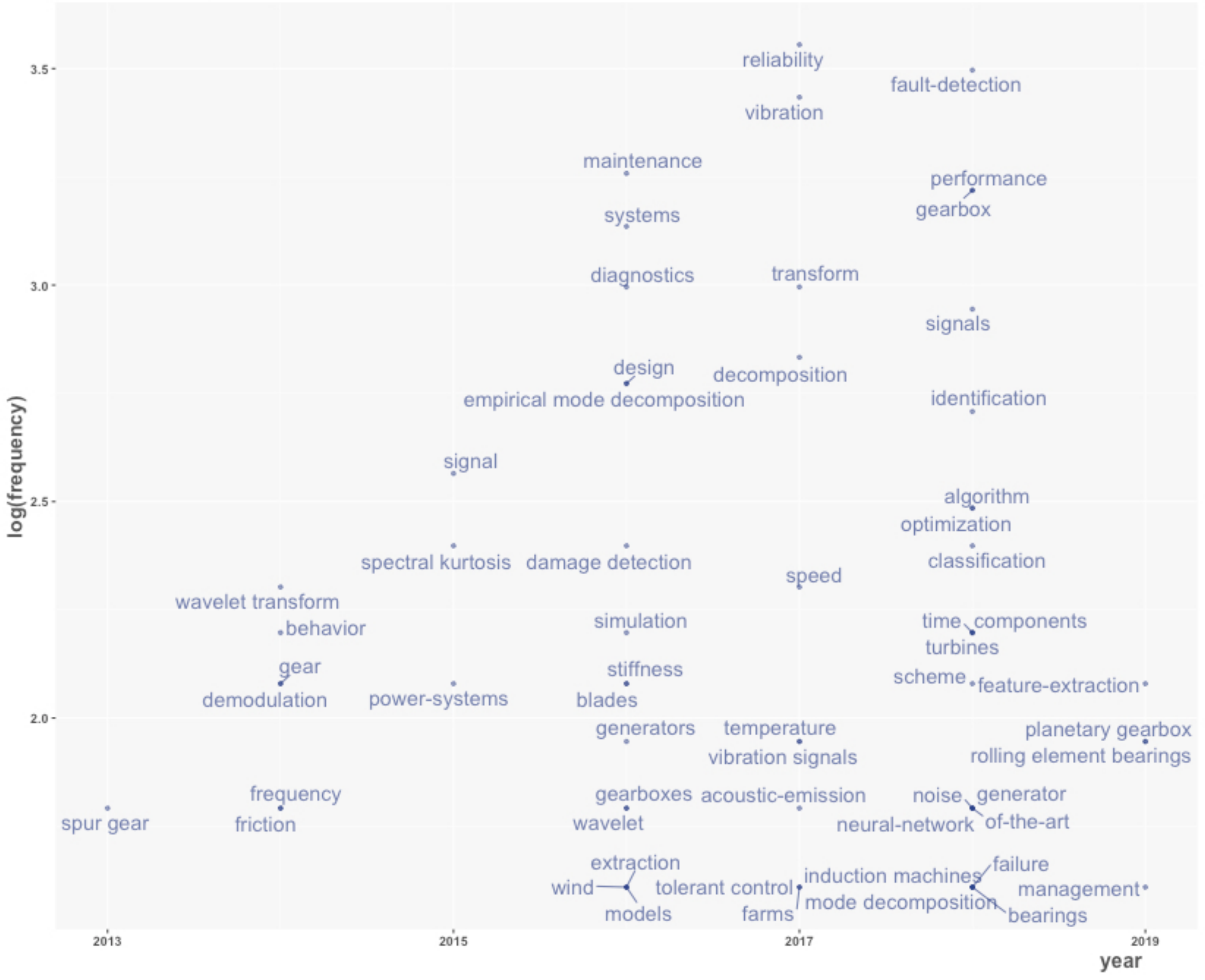}}
\caption{Trend topics for data-driven decision making in the wind industry. The rising interest towards more sophisticated algorithms e.g. neural networks is clearly outlined, while conventional techniques (e.g. based on signal processing) still continue to be in use. \label{trend_cbm}}
\end{figure*}

More recently, there has been rapidly growing interest in applying deep learning techniques for O\&M tasks, especially for classification of turbine faults. Figure ~\ref{trend_cbm} outlines the evolution of trend topics in data-driven decision making for the wind industry with time, clearly showing a move from more traditional methods based on signal processing in the past towards neural networks for time-series SCADA data.

The essence of deep learning is the use of neural networks with multiple layers, capable of learning from complex non-linear relationships in data \cite{neural_nonlinear}. Neural networks have the ability to find associations or patterns between inputs and outputs, making them extremely competent to learn and model complex intermediate representations within the data \cite{goodfellow_bengio_courville_2017}, which is generally the case with SCADA features. In a recent effort in utilising neural networks for CBM of specific turbine sub-components, Lu et al. \cite{opp_cbm} utilised SCADA data and applied ANNs for predicting life percentage of the turbine sub-components. Their approach can be utilised to identify faults based on conditional probability of failures, obtained through the ANN's predictions and the historical component failure time distribution. Such approaches can assist O\&M operators and technicians to better plan the inventory and maintain surplus storage for the sub-components most prone to failure. However, the paper does not enunciate on the training and testing performance of the ANNs, as well as the basis of choosing the specified network architecture. Moreover, the paper only focuses on predicting the lifetime percentage for four sub-components of the turbine, viz. pitch system, gearbox, generator and rotor, which, the authors mention,  are most prone to failures. However, there are many other integral sub-components in a turbine, including drive train, yaw system, hydraulic system, electrical system etc. which the paper does not address. 

Qian et al. \cite{cbm_elm}
have previously proposed an \nomenclature{$ELM$}{Extreme learning machine} Extreme learning machine (ELM) model for CBM, which can help identify faults based on deviation from ideal SCADA signals. The paper shows that the ELM model performs better than the conventional feedforward neural networks and takes considerably less time to train and make predictions, given that it can randomly update the weights and bias unlike ANNs, which utilise gradient-based learning algorithms for optimisation. Moreover, the model can predict incipient faults in advance for specific sub-components, directly contributing to reduced maintenance costs. However, the paper lacks in presenting the details of the ELM model used and a comparison between the performance of the ANNs and ELM in terms of fault prediction accuracy and training time. Also, as deviation from ideal signals might not always be indicative of a fault, this approach can often raise false alarms owing to the high sensitivity of the sensors, inadvertently causing forced outages and increased costs. 

Some studies have applied AI in computer vision techniques for fault diagnosis and indications of incipient failures in external turbine sub-components \cite{image_turblade,image_turblade1,image_turblade2,vision_turblade}. In this domain, \nomenclature{$CNN$}{Convolutional neural networks} (CNNs) have been utilised by Li et al. \cite{image_turblade1} to identify faulty instances of turbine blade images, achieving accuracy close to 100\% in some cases. In addition, hybrid models such as combinations of SVM with CNNs proposed by Yu et al. \cite{image_turblade} have also shown success in learning from small datasets of labelled turbine blade images. While such methods are well suited for CBM of external turbine sub-components like blades, they cannot be utilised for anomaly prediction in several other integral turbine sub-components such as gearbox, pitch system etc. Additionally, given that drones or other similar image capturing devices need to be employed for recording images, this methodology is not cost-effective, and is prone to failures and false alarms during rain, mist, snow etc.

Installing wind turbines at new sites generally requires critical analysis of the location and weather conditions to ensure maximum power production possible at the lowest cost. There has been some work in the area of optimising turbine performance by appropriate planning of layouts such as by \cite{windturlayout_1,windturlayout_2,windturlayout_3} etc. A significant study in this area by Dutta et al. \cite{windturlayout_3} makes use of AI techniques, such as genetic algorithms, for layout planning of turbines and optimisation algorithms like the ant colony algorithm for deciding optimal line connections in the topology. The study considers wake effect and utilises wind speed time series and cable parameters for turbine interconnections in the wind farm for optimising turbine layout. The paper demonstrates the immense promise which AI provides in optimising turbine topologies, as AI algorithms can take into account an exponential number of cases and distributions for finding optimal solutions, which is generally infeasible through manual optimisation techniques.

Clearly, while most existing studies focus on performance assessment based on power prediction metrics (e.g. power curve) and deviation from ideal SCADA signals, the research specifically focused on CBM pertaining to anomaly prediction and identification of incipient faults is still in an embryonic stage. In an early work in this area (which interestingly, is possibly the only paper pre-2015 to focus on deep learning for CBM), 
Zaher et al. \cite{online_faultdetection} have utilised SCADA data from turbines for temperature anomaly detection in sub-components such as the gearbox and generator. The paper applies multilayer neural networks towards identifying abnormalities in operational temperature of these sub-components, and the methodology was extended to an entire wind farm using a multi-agent system (MAS) architecture. This study did not utilise historical fault logs, which were not available to the authors and are generally difficult to obtain for research, owing to its commercial sensitivity to the wind farm operators. While this early work showed immense promise for deep learning towards CBM and anomaly prediction in specific sub-components, the purpose of complete automation is defeated, as it still requires professional technicians/maintenance engineers to identify and classify specific classes of faults (e.g. based on severity and specific alarm events).

Following a different methodology, Andersen et al. \cite{scalable_faultdetection} have utilised convolutional neural networks (CNN) for fault prediction using vibrational signals from turbines. The paper demonstrates highly promising results, with the CNN outperforming conventional ML techniques used as baselines significantly. Ibrahim et al. \cite{neuralnet_csa} have achieved similarly promising results with deep learning applied to SCADA data, specifically utilising current signature analysis and artificial neural networks for anomaly prediction. More recently, Pang et al. \cite{pang_he_jiang_xie_2020} utilised a hybrid spatio-temporal fusion neural network for multi-class fault prediction using SCADA data. Specifically, the paper proposes the application of a multi-kernel fusion convolution neural network to learn multiscale spatial features and correlation between these variables along with an LSTM to further learn temporal dependencies. The proposed technique outperformed several conventional ML techniques, outlining the promise of deep learners, especially hybrid models for fault prediction in turbines. In a closely related study, Kong et al. \cite{kong_tang_deng_liu_han_2020} developed a hybrid model consisting of CNNs along with \nomenclature{$GRU$}{Gated recurrent units} Gated recurrent units (GRUs) for fusing spatio-temporal SCADA features. Similar to LSTMs, GRUs are able to learn temporal dependencies in complex and non-linear SCADA data, while utilising fewer training parameters, which (in some cases) make it more computationally efficient and accurate as demonstrated in the paper. This CNN-GRU model was trained using historical data for normal behaviour of turbines, and any deviation from normal operation in terms of residuals was utilised to detect anomalies. The paper demonstrated the effectiveness of the method for anomaly prediction, especially as a monitoring indicator during CBM. Note that while these studies show such deep learners to predict faults with high accuracy, they cannot provide rationales and transparency in their decisions, regarding the features exactly leading to the predicted faults \cite{windenergy_journal}, which may make turbine operators reluctant to practically adopt such approaches.

Given the challenges (and time-consuming nature) of obtaining historically labelled SCADA data with fault records, some studies have applied unsupervised learning techniques for anomaly prediction. This includes application of  \nomenclature{$DAE$}{Denoising autoencoders} denoising autoencoders (DAE) by Jiang et al. \cite{8059861}, who utiliummed time-series SCADA data from multiple sensors and demonstrated the ability of DAEs to learn non-linear representations of SCADA features in situations of noise and input fluctuation. The authors trained DAE with normal data, and by using a multivariate reconstruction model, they analysed the reconstruction error for detecting faults. The study also utilised a sliding-window technique, which can help capture the prevailing non linear correlation between multiple SCADA features as well as the temporal dependencies in the features, which provides it highly promising performance for effectively detecting faults.  As neural networks have mostly been applied in supervised learning scenarios in the past, this study demonstrates the promise of unsupervised learning for real-world operational data from turbines. Note that one common challenge which most existing studies face is the lack of transparency in the black-box natured AI models, which despite generating highly accurate predictions fail to provide rationale behind their decisions. Also, unsupervised learning techniques when utilised without historical ground truth for failures cannot be validated, which makes them less robust in comparison to supervised learning methods.

\begin{figure*}[!h]
\centerline{\includegraphics[width=0.7\textwidth]{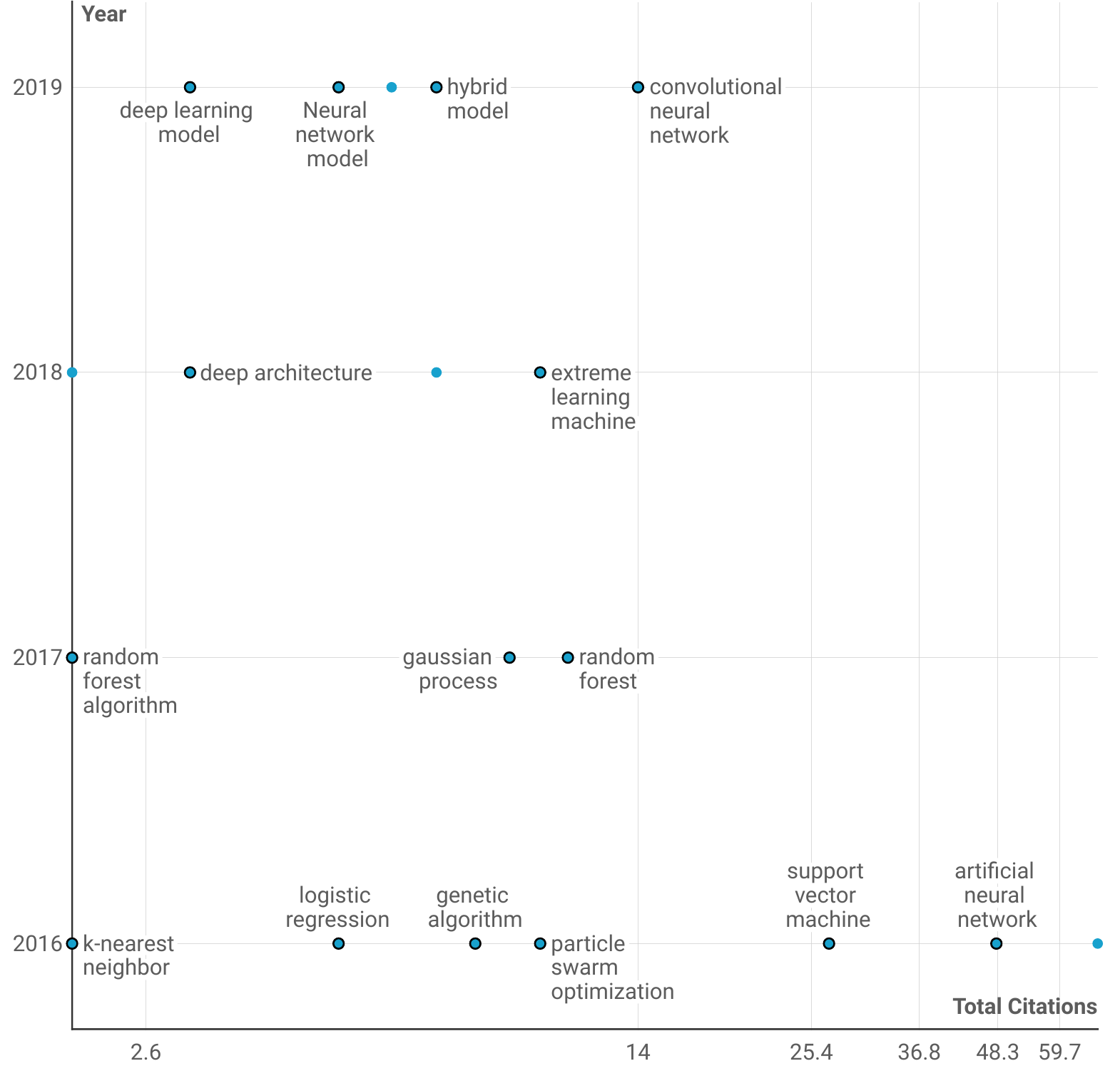}}
\caption{Citation burst for AI in the wind industry during 2016-2019. The top 15 terms (logarithmic scale) prevalent in cited papers utilising such models are used. \label{citeburst_ai}}
\end{figure*}

Some studies have employed Explainable AI models to tackle the issue of transparency. Chatterjee and Dethlefs \cite{windenergy_journal} utilised a hybrid model consisting of LSTMs along with a \nomenclature{$XGBoost$}{Gradient-boosted decision tree classifier} Gradient-boosted decision tree classifier (XGBoost) towards explainable anomaly prediction in wind turbines. This study also demonstrated the feasibility of transfer learning, facilitating prediction of faults in new domains (e.g. wind farms which have not been in operation for long) without access to historically labelled failure data. Wang et al. \cite{9041585} have utilised a specialised type of LSTM with attention mechanism to achieve transparent and interpretable wind power prediction. The attention mechanism \cite{luong_pham_manning_2015} in neural architectures facilitates learning models to dynamically focus on the vital and relevant predictive features in the sequential data (time-series) and also provides the list of features influencing the model's decisions, leading to more accurate and transparent predictions.  Similar efforts have been made to apply CNNs with attention mechanism in order to achieve highly accurate and explainable predictions, e.g. by Kumar et al. \cite{en13071772} for short-term prediction of wind speed, by Jianjun et al.\cite{en12142764} for imbalance fault detection in turbine blades and Chatterjee and Dethlefs \cite{torque_paper} to identify causal associations in SCADA data during fault predictions. All these studies provide novel insights on the feasibility and promise of AI models tailored for the wind industry, especially RNNs and CNNs. However, clearly, the applications of Explainable AI to the wind industry is very limited compared to other domains such as computer vision and natural language processing (NLP) \cite{renew_paper}, and we discuss more on this in Section ~\ref{perspective}.

\begin{figure*}[!h]
\centerline{\includegraphics[width=0.7\textwidth]{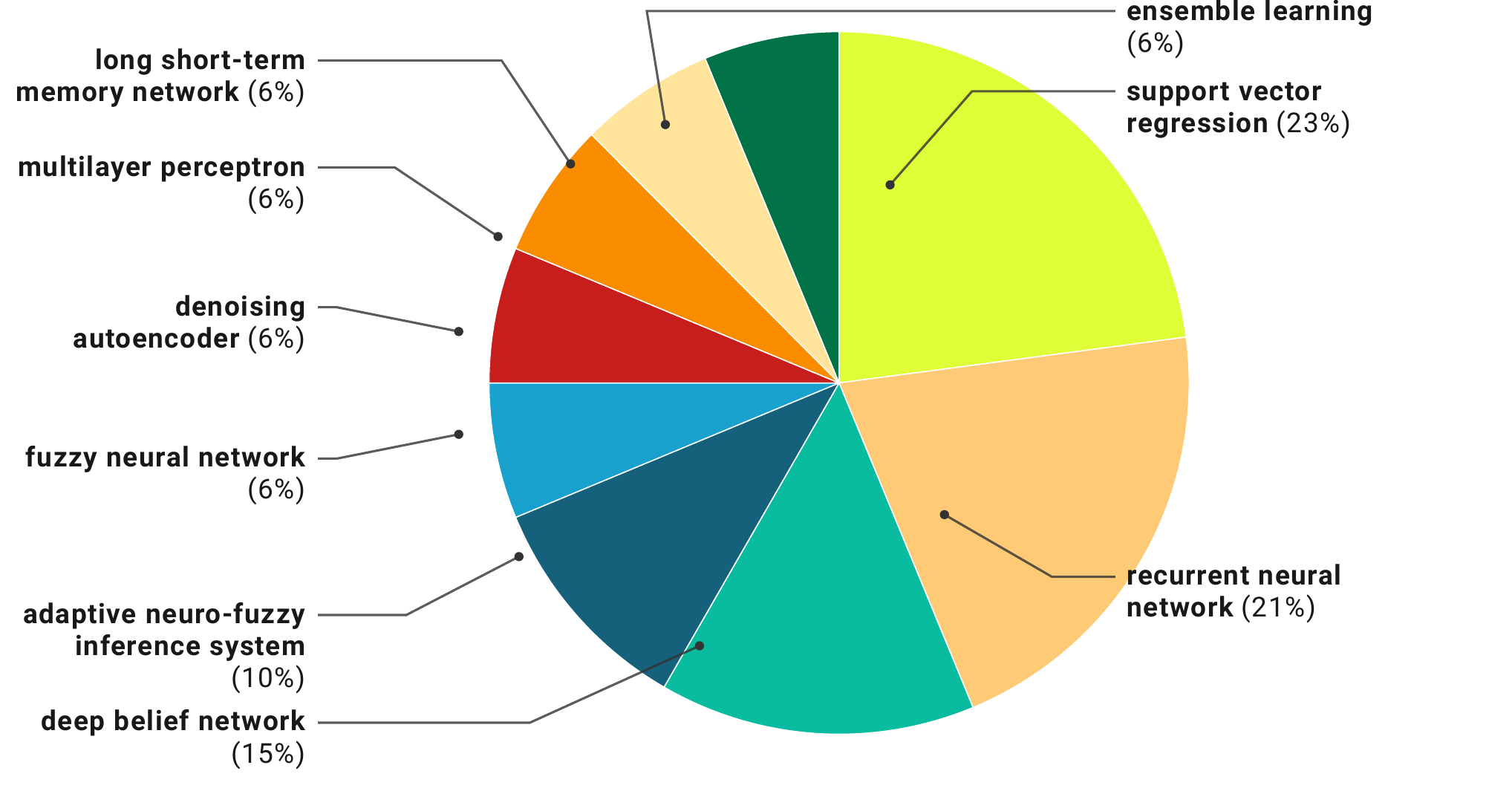}}
\caption{Pie chart outlining composition of less popular AI techniques for CBM based on keywords citation burst during 2016-2019. The least frequent keywords in cited papers across 291 CBM publications utilising such models are used.\label{pie_lessp}}
\end{figure*}

Figure ~\ref{citeburst_ai} shows the top 15 keywords in publications applying AI for CBM, which have received the strongest citation bursts (demonstrating significant research interest) in the wind industry. Interestingly, we note that this period was prevalent from 2016-2019, which clearly shows that AI for wind turbine CBM received a massive interest amongst researchers during this time. We also note that the publications utilising artificial neural networks in 2016 garnered maximum citations, with SVMs being the second most popular technique. An important inference is the dynamic shift of the interest from conventional ML techniques (such as k-nearest neighbour, logistic regression, genetic algorithms, random forests and particle swarm optimisation) in the early applications of AI in the wind industry (2016-17) to more sophisticated models, specifically utilising deep learning (extreme learning machine, deep neural networks etc.). From 2019, convolutional neural networks and hybrid models combining multiple neural net architectures have driven significant research interest in the wind industry. In comparison to this growth, there has been comparatively little interest in adopting many other AI models (including long short-term memory networks, autoencoders and fuzzy neural networks).  Figure ~\ref{pie_lessp} depicts the composition of these less popular techniques. Support vector regression and recurrent neural networks have interestingly dominated these less popular techniques despite their limited attention, which shows that they are likely highly promising for CBM. We believe it is integral for the wind industry to more widely adopt such models for optimal benefits from data-driven decision making.

\paragraph{\textbf{Natural language generation techniques for human-intelligible decision support}}
While most existing studies applying AI models for data-driven decision making focus on utilising SCADA data, they have significantly neglected additional vital information available, especially historical logs of alarms/failures. These records (generally referred to as \textit{event descriptions}) contain comprehensive information of the historical faults in turbines in the form of natural language phrases describing the alarms in turbine sub-components (e.g. pitch system, gearbox, yaw etc.) in addition to the time-stamps for the events in relation to the SCADA features. To generate \textit{informative} messages from SCADA data, which is a data-to-text generation problem, some studies have applied \nomenclature{$NLG$}{Natural language generation} Natural language generation (NLG) techniques, building upon the immense success which such methodologies have shown in domains like weather forecasting, spatial navigation, automated planning etc. \cite{gong-etal-2019-enhanced,webnlg-automatic2017,Garoufi:2010,JuraskaNAACL2018}.
NLG can often play a critical role towards shortening the analysis time frames in O\&M decision support as well as providing human-intelligible decisions, assisting engineers to better understand the context of occurring faults. Additionally, the purpose of data-driven decision making and automated planning is more or less defeated if AI models are not able to provide maintenance action suggestions besides accurate fault predictions. NLG techniques are a boon towards achieving transparent decisions, especially considering the sequential nature of data in the wind industry (alarm messages, maintenance report documents and SCADA features). Specialised NLG techniques, such as few-shot learning \cite{chen-etal-2020-shot} also provide the ability to generate informative messages even with limited training data, making NLG highly promising for adoption in the wind industry.

In one of the earliest works in this domain, Sowdaboina et al. \cite{10.1007/978-3-642-54906-9_42} utilised rule-based NLG techniques to summarise time-series information relevant to the wind industry, primarily wind speed, wind direction etc. Dubey et al. \cite{DBLP:conf/flairs/DubeyCK18} have applied \nomenclature{$CBR$}{Case-based reasoning} Case-based reasoning (CBR) techniques to develop an end-to-end system to generate textual summaries of such meteorological information, demonstrating highly promising results when CBR techniques are combined with rule-based NLG techniques. Despite showing success in presenting such information, a key drawback of current studies is that they can only present very limited information (i.e. 1-2 parameters), when there are multiple (often hundreds) of SCADA features and different failure types in turbines that could be utilised for transparency in decision support. Developing NLG systems for such tasks is not only challenging and time-consuming, but also creates specific constraints when sufficiently labelled information is not available (e.g. ground truth labels of SCADA features contributing to faults).

To tackle such challenges, AI models have seen very limited, but extremely promising results in the wind industry. In possibly the only work in this area, Chatterjee and Dethlefs \cite{wcci_paper} have demonstrated the ability to utilise AI-based NLG models, such as transformers for decision support. The transformer \cite{NIPS2017_7181,vig2019transformervis,devlin-etal-2019-bert} is a specialised neural architecture consisting of multi-head attention mechanism, giving it the ability to focus on relevant features in sequential datasets and eliminating recurrence used in vanilla RNNs completely. This generally helps the model better learn relationships between features and also reduces the computational complexity significantly, making transformers highly promising for training on modern ML hardware. Given the sequential nature of SCADA data and the desired outputs (alarm messages and maintenance actions), such techniques have shown success in providing detailed human-intelligible diagnoses for failures as well as suggesting maintenance actions appropriate to avert catastrophic failures. Additionally, such models are explainable and transparent, and can provide exact lists of features which lead to predictions of alarms and maintenance actions through mechanisms based on multi-head attention \cite{NIPS2017_7181}. Further brief details on such NLG models is provided later in Section ~\ref{perspective}. It is clear that while NLG techniques have seen promise in their early applications to CBM, they have not been widely applied and adopted for data-driven decision making. We believe it is vital to utilise NLG models, especially leveraging deep learning to generate human-intelligible maintenance reports for O\&M. 

\begin{figure*}[!h]
\centerline{\includegraphics[width=0.8\textwidth]{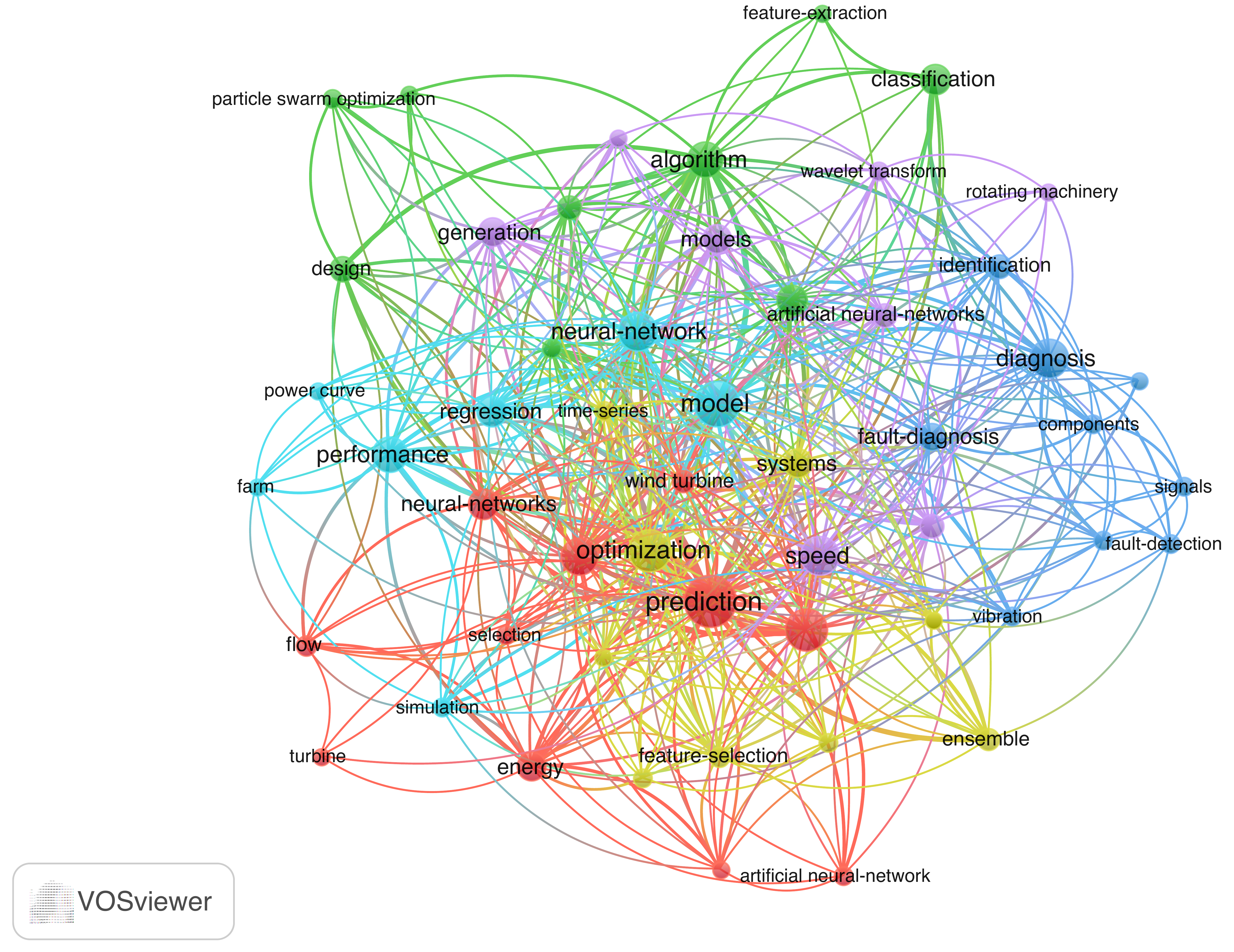}}
\caption{Network visualisation in VOSviewer for AI in the wind industry as evident from publications from 2010-2020. The graph edges indicate conceptual association between terms across clusters, with similar colours depicting strong association. \label{nvis_ai}}
\end{figure*}

\paragraph{\textbf{Reinforcement learning for planning and optimisation }}
Owing to the highly complex and uncertain environment in which turbines are deployed, optimisation and control of turbines as a system is often critical. To achieve this,  \nomenclature{$RL$}{Reinforcement learning} Reinforcement learning (RL) \cite{robbins1951,drl_paper}, a specialised branch of AI techniques, has seen some application in the wind industry for autonomous decision making and planning. In some early studies, e.g. Tomin et al. and Gauna et al. \cite{8771645,gauna_paper} trained RL algorithms for intelligent control of a \nomenclature{$MIMO$}{Multi-input-multi-output} multi-input-multi-output (MIMO)-based controller in the turbine system. These studies achieved promising results compared to traditional control methods, which often face challenges in multi-objective problems common to modern wind turbines. In another vein of work, Aguirre et al. \cite{doi:10.1002/we.2451} have applied deep reinforcement learning techniques for wind turbine yaw control and demonstrated that such techniques incorporated with the learning capabilities of ANNs significantly outperform traditional RL algorithms. Chatterjee and Dethlefs \cite{renew_paper} have demonstrated similar promise of deep reinforcement learning for maintenance planning of offshore vessel transfers based on operational SCADA data and other parameters e.g. weather conditions and predicted fault types and severity. This shows that reinforcement learning is highly promising and feasible for the wind industry, and we believe that more research should be pursued in this domain to provide better planning and optimisation in O\&M approaches.

Figure ~\ref{nvis_ai} shows network visualisation of all AI publications in the last decade, outlining the stagnant rise of AI for CBM in the wind industry. Based on the graph edges and multiple clusters (represented by different colours), it can be enunciated that there is prevalence of predictive techniques for classification, regression and optimisation tasks with neural networks and signal processing techniques (e.g. wavelet transform) and power curves being used for such purposes. Also, note that feature selection and feature extraction continue to play an important role in such models, outlining significant focus on feature engineering in SCADA data, as in conventional ML algorithms. There is clearly extremely limited focus on NLG and reinforcement learning techniques. 

\section{Perspectives into the future}\label{perspective}
As evident from scientometric analysis of the past and present, the wind industry is facing an interesting and challenging problem in autonomous prediction and scheduling of O\&M using data-driven techniques. While existing studies make advances in some specific areas, such as wind power forecasts \cite{doi:10.1002/we.2497} and anomaly prediction \cite{anomaly_dl}, there has been limited research in incorporating explainability and transparency into data-driven AI models. The lack of research, particularly in fault prediction during CBM, can most likely be attributed to the issue of obtaining SCADA data from wind turbines (especially with labelled history of alarms and failures), which is often commercially sensitive to the wind farm operators. 

Below, we discuss some of the major challenges the wind industry is presently facing (and will likely continue to face in the near future) in applying AI techniques for data-driven decision making, and provide a perspective on possible ways to tackle them.

\begin{table*}[t]
\small
\renewcommand{\arraystretch}{1.5}
\caption{Summary of openly available datasets in the wind energy sector which can be utilised for CBM \label{data_open_cbm}}
\begin{tabular}{|p{4cm}|p{7cm}|p{2cm}|}
\hline
\textbf{Dataset} & \textbf{Type} & \textbf{Year Released}\\ \hline
2011 PHM Society Conference- Anemometer Fault Detection Data Challenge \cite{phmsoc_datachallenge} &
Paired anemoemeter data based at same height and shear data for anemometers at different heights, comprising parameters like wind speed, wind direction and temperature aimed at identifying excessive error owing to damage or wear conditions    & 2011  \\ \hline
Wind Turbine High-Speed Bearing Prognosis Dataset \cite{hsb_prognosismatlab,hsb_prognosisgithub, conf_hsbprognosis} & Bearing health prognosis dataset consisting of vibration and tachometer signals from a real-world turbine high-speed shaft bearing \cite{conf_hsbprognosis}, which also faced actual inner race fault conditions  & 2013  \\ \hline
ENGIE La Haute Borne \cite{engie_data} & SCADA data from an operational onshore wind farm & 2013  \\ \hline
NREL Wind Turbine Gearbox CBM Vibration Analysis Benchmarking Dataset \cite{nrel_wgbbench} & Vibration data obtained through accelerometers and high-speed shaft RPM signals collected during dynamometer testing, alongside information on real damage conditions in turbine gearbox for performing benchmarking of vibration based CBM techniques & 2014  \\ \hline
Platform for Operational Data: Levenmouth Demonstration Turbine \cite{ldt_data} & Data from an operational offshore wind turbine, including SCADA, historical logs of alarms, substation data, and Met mast data & 2017  \\ \hline
Ørsted Offshore Operational Data \cite{orsted_data} & SCADA data from 2 operational wind farms, with on-site 10 minutes statistics from wave-buoy and ground based LiDAR & 2018  \\ \hline
EDPR Wind Farm Data \cite{edpr_data,edpr_data_comp} & Historical dataset from an operational offshore wind farm comprising of SCADA signals, Met mast data, turbine failure logs and relative positions of turbines and Met mast  & 2018  \\ \hline
\end{tabular}
\end{table*}

\begin{table*}[t]
\small
\renewcommand{\arraystretch}{1.5}
\caption{Summary of openly available datasets in the wind industry which can be utilised for performance assessment of turbines\label{data_open_performanceass}}
\begin{tabular}{|p{4cm}|p{7cm}|p{2cm}|}
\hline
\textbf{Dataset} & \textbf{Type} & \textbf{Year Released}\\ \hline
NREL Western Wind Dataset \cite{west_data} & Data with historical weather information (wind speed, air temperature, pressure etc.) and power output from multiple operational wind turbines & 2004  \\ \hline
NREL Eastern Wind Dataset \cite{east_data} & Simulated data of wind speed and turbine power output, with short-term forecasts & 2004  \\ \hline
Platform for Operational Data: Floating Turbine Design Cases \cite{floating_data} & Measurements from an operational floating turbine, with operational cases for multiple wind speeds and wave heights & 2019  \\ \hline
\end{tabular}
\end{table*}

\subsection{Data availability and quality ensurance}
AI techniques rely on huge amounts of data for optimal decision making in real-world applications \cite{FORESTI2020}. However, given the commercially sensitive nature of data from wind turbines \cite{windenergy_journal}, most wind farm operators are reluctant to share such information openly in the public domain, which is vital for researchers. Additionally, annotating rapidly changing events and alarms for complex engineering systems like wind turbines is challenging for engineers and wind farm operators, and may not always be on top of the agenda. In some cases, new turbines may not have been in operation for long \cite{windenergy_journal}, creating a challenge in acquiring even small datasets. To analyse the present situation in terms of data availability in the wind industry, we present a summary of some openly-available datasets which can be utilised for CBM of turbines in Table ~\ref{data_open_cbm}, which to the best of our knowledge, are the only sources of information available in the public domain. Notably, only two of the above sources of data contain historical logs of alarms and failures. It is also interesting to note that there are some other types of openly available datasets falling outside the scope of CBM as listed in Table ~\ref{data_open_performanceass}. While these cannot generally be utilised for informative decision making to train AI models pertaining to fault diagnostics or prognostics in O\&M, they can help in performance assessment of turbine operations (e.g. efficiency and power production). The interested reader is referred to \cite{en13184702} for comprehensive details on the applications of various datasets available in the wind industry for O\&M.

While the datasets with historical weather information and turbine power outputs can be beneficial for forecasting future trends and deriving useful insights about operational feasibility of turbines (e.g. through power curves) \cite{7011548}, they cannot be utilised for 
predicting faults in specific sub-components of turbines and providing possible causes behind a particular fault. The datasets consisting of vibration data, especially those which include vibration signals recorded during circumstances of faulty conditions are potentially more useful as they characterise the operational status of specific turbine sub-components (e.g. gearbox) and can play an integral role in supporting vibration analysis based CBM in O\&M of turbines. The SCADA datasets presently available openly provide the ability to identify operational parameters in the turbine and its sub-components (such as pitch angle, gearbox oil temperature, active and reactive power etc.). Additionally, these datasets usually contain meteorological information (wind speed, air pressure, temperature etc.) measured at the Met Mast \cite{Mittelmeier_2016}, and can be useful for wind resource assessment. In cases wherein historical records of alarms logs in the turbine are not available, unsupervised AI techniques for outlier detection \cite{goldstein_uchida_2016} can be applied to discover hidden patterns in the SCADA features and identify potential faults based on discriminatory features affecting data points in certain clusters. 
However, such techniques cannot be validated due to lack of ground truth for normal behaviour/anomaly, and more importantly, lack the ability to provide more detailed description of faults and their causes, e.g. through alarm messages.

SCADA datasets containing labelled history of alarms prove to be more useful in applying AI techniques, as supervised learning techniques for fault prediction \cite{RePEc:eee:rensus:v:98:y:2018:i:c:p:189-198} can be used to develop predictive models by training on a portion of the historical data, and facilitate making predictions on new, unseen test data. This is often more reliable, as performance metrics (e.g. accuracy) can be obtained given that original labels (normal operation/anomaly) are available. Also, as alarm logs contain detailed description of faults in terms of messages describing exact sub-component having the fault and its characteristics \cite{wcci_paper}, it can provide significantly detailed insights for O\&M. However, as already outlined earlier, such datasets are extremely difficult to obtain. In addition, this creates a major challenge in producing meaningful new results and comparisons with baselines, because researchers generally apply models to very specific SCADA datasets (which vary widely in terms of features and specifications), and the data used in the papers are mostly not shared with the published research. This trend severely limits comparability and replicability of published research.

Besides the significant difficulty in access to data, there are also major challenges posed by the quality of datasets available in this domain. With continuing developments in the wind industry, different types of big data with high resolution and complexity are becoming available from lidars and buoys, wind and wave metrics and operational data from hundreds of sensors etc., making it integral to perform proper filtering and quality control for clearly providing vital information on state of the turbines \cite{data_longtermchallenge}.  Note that data quality issues affect not only fault diagnostics and prognostics in CBM, but also additional O\&M tasks which may be beyond the scope of CBM but could be vital to turbine operators pertaining to performance assessment.

The wind industry has witnessed very limited attention in identifying key issues that persist with utilising turbine data, which is particularly vital for training AI models which rely on accurate and scalable data \cite{8862913} for informative decision making. In possibly the only study which specifically focuses on data quality, Leahy et al. \cite{en12020201} outlined the pressing issues pertaining to lack of unified standards for different datasets in the wind industry (such as SCADA, alarm codes, maintenance and work orders etc.), limited availability of alarm data with useful context and the significant requirements for manually processing datasets into usable formats for training data-driven CBM models. More recently, with growing research in utilising AI for O\&M in the wind industry, particularly deep learning techniques, other new issues are emerging in this domain for development and deployment of highly sophisticated models as described below:-

\begin{enumerate}
    \item \textbf{Imbalanced datasets:} SCADA datasets consisting of historical records of alarms generally suffer from a major imbalance prevailing between the data samples for normal operation and anomalies \cite{windenergy_journal}, with a significantly higher number of data samples categorised as normal operation owing to limited records for failure conditions, or with some types of faults (e.g. in gearbox, generator and blades) having much higher failure rates than others \cite{reliabilityanalysis_turbine}. Training on imbalanced datasets can make the AI models biased towards the majority class (labelled as normal operation), and thereby lead to missed detections with the model classifying anomalous situations as normal. These situations are likely to be overlooked by turbine operators during O\&M \cite{Leahy_2017}, and can result in unexpected failures and significant costs.
    
    \item \textbf{Inadequate quality of contextual information on faults:} The SCADA alarm systems record alarm patterns which can indicate failure occurrences in turbine sub-components, as well as the relationship of component failures amongst other sub-components and adverse environmental conditions \cite{Gonzalez_2016}. More recently, these alarms are often available in the form of brief natural language phrases, which provide contextual information of the faults (e.g. \textit{Wind direction transducer error 1 \& 3}) and can be utilised in data-to-text generation systems for producing event descriptions from SCADA data to fix/avert failures corresponding to expert judgements \cite{wcci_paper}. Data-to-text generation systems utilise natural language generation (NLG) techniques for generating human-intelligible unstructured textual descriptions of failures from structured SCADA data. NLG techniques, especially neural machine translation models heavily depend on appropriate quality of data samples for training and low-quality examples are quickly memorised by such models \cite{nlgquality_paper1}. In the wind industry, the alarm messages available are often of inadequate quality for training NLG models to achieve human-level intelligence, and suffer from a lack in diversity of available corpus as some types of alarm messages (e.g. \textit{Pitch System Fatal Error} owing to the fairly frequent occurrence of pitch angle disorientation in turbines) \cite{wcci_paper,doi:10.2514/6.2013-1695} are generally very common in O\&M routine tasks and are thereby given more attention in the wind industry. Engineers \& technicians may not prioritise manually annotating (or developing suitable automation techniques) to develop corpus for alarm messages summarising contextual information for low-priority faults (e.g. \textit{HPU 2 Pump Active For Too Long}) \cite{joyjit_nips}, which leads to a significant variation in the quality of available messages across different sub-components. Appropriate diversity in data samples is essential for generating coherent text and providing useful insights \cite{ji-etal-2020-amazing} for domain-specific tasks, which makes utilising NLG in decision support challenging for the wind industry. Moreover, unlike plain text (e.g. utilised for translating from one language to another) alarm messages in the wind energy domain often consist of important symbols \& numbers (e.g. \textit{(DEMOTED) Gearbox oil tank 2 level shutdown} \cite{joyjit_nips} detailing the exact tank in the gearbox which was shut down as a result of the fault), and NLG models generally miss out on learning these nuances with such symbols potentially contributing as noise within the natural language message phrases \cite{khayrallah-koehn-2018-impact, wcci_paper}. 
    
    Besides the aspect of contextual failure information being vital for NLG, there are also some other areas which may potentially benefit significantly from adequate availability of such information. This may, for instance, influence planning of offshore vessel transfers for O\&M, wherein, improved context on faults can generally help maintenance personnel better anticipate the required parts to carry. This can thereby help facilitate improved inventory management and planning in the wind industry. In other potential uses of contextual failure information, the context may include thorough details presented in the form of service logs for turbines, which can be instrumental in contextualising rarer faults and new errors/operational inconsistencies which were witnessed by maintenance personnel during O\&M of the turbine. All these aspects thereby directly contribute to human-intelligible and informative decision making, which can be integral for O\&M in the wind industry.
    
\end{enumerate}

Below, we outline some of the key areas wherein the wind energy sector may focus to tackle the challenges in data availability and quality:-

\begin{itemize}
    \item \textbf{Encouraging more wind farm operators to provide open data:} The simplest way to apply AI models is to acquire more \textit{useful} data, which can be used to train the decision making models over more diverse scenarios of turbine operation. While a few organisations have already taken the positive steps towards making SCADA data (and in some cases, historical logs of alarms) publicly available (as per Table ~\ref{data_open_cbm}), clearly, this is not sufficient for training present-day AI models to become more robust (and autonomous), especially deep learners which generally require high volumes of data \cite{dl_hugedata} to tune the model and its parameters. 
    
    If a few turbine operators can make their data public, what stops other operators from sharing their SCADA data (and failure logs) for the purposes of Research \& Development? According to \cite{kusiak_2016}, \textit{competition} is the prime reason behind this, as the sensitive data from turbines can often reveal performance metrics and expose poor design practices. While this is indeed a challenge, we believe that there are multiple options which the wind farm operators could explore, including developing non-disclosure agreements and anonymising certain sensitive information (e.g. detailed technical specifications). Also, as wind turbines suffer degradation and are  decommissioned after the end of their useful life\cite{TOPHAM2017470}, we believe that historical data from the decommissioned (thereby non-operational) turbines can help facilitate development and training of AI models for experimentation, tests etc., which can later be adapted to any new data sources when they become available, at the same time not falling into the constraints of commercial sensitivity.

    \begin{figure}[!h]
\centerline{\includegraphics[width=0.6\textwidth]{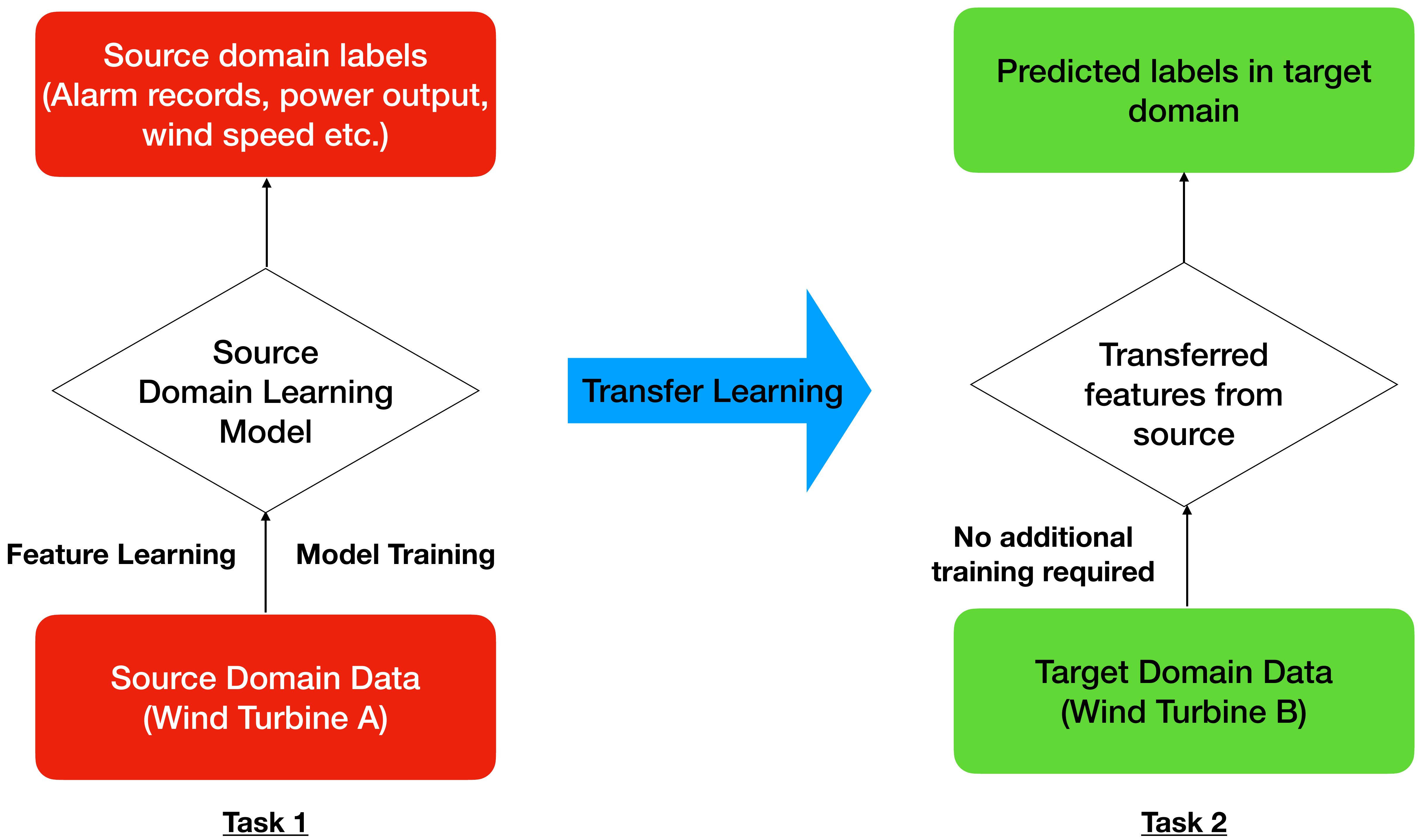}}
\caption{Depiction of the typical process for knowledge transfer. Source domain data can be SCADA features, meteorological parameters as well as unstructured data such as maintenance manuals etc. \label{tl_proc}}
\end{figure}
     \item \textbf{Wider adoption of transfer learning in leveraging insights from any available data:}
    In machine learning, it is often challenging to obtain training data for creating high-performance learning models matching the feature space distribution of test data \cite{tl_defination}. This makes it integral to create learning models for the target domain by training on a closely related source domain as depicted in Figure ~\ref{tl_proc}. The wind industry has seen very limited application of transfer learning techniques in comparison to applications in other domains such as natural language processing and computer vision \cite{10.1007/978-3-319-97982-3_16,ruder-etal-2019-transfer}. Only a few studies focus on applying transfer learning techniques to SCADA data from turbines, and are primarily aimed at wind power prediction \cite{QURESHI2017742} for performance assessment/analysis. Some other studies utilise transfer learning for short-term wind speed prediction \cite{RePEc:eee:renene:v:85:y:2016:i:c:p:83-95} and ice assessment on turbine blades \cite{8409794}. However, to the best of our knowledge, there has been scarce application of transfer learning towards predicting faults in turbines, which is an integral aspect of O\&M. The only works in this area either focus on prediction of faults in different turbine sub-components \cite{windenergy_journal} or monitoring vital parameters e.g. of the gearbox to identify deviation from normal behaviour towards fault prediction \cite{doi:10.1177/1475921720919073}. 
    
    It is vital for the wind industry to apply AI techniques towards more fine-grained analysis and prediction of failures in turbines by utilising transfer learning  from historical alarm message records, operator manuals, work orders etc. Moreover, given that such records (as described before) are the most difficult to obtain for researchers and challenging to annotate for engineers, applying transfer learning can be extremely beneficial in facilitating the  development of high-performance learners even in the absence of sufficient training data. We envisage that a wider adoption of such techniques in the wind industry can help in enhancing the uptake of AI for CBM and contribute towards making the O\&M process more dependable in situations with paucity of data.

    \item \textbf{Quality control of datasets:} Some studies have highlighted the necessity for quality control of datasets in the wind industry, especially through standardisation of information and development of unified standards and taxonomies by turbine operators and manufacturers \cite{en12020201, data_longtermchallenge}. This is indeed important to successfully develop and deploy highly sophisticated AI models in a long-term perspective. Based on our review in this paper, we believe that there some options which could be leveraged to encourage quality control of data. Firstly, wind farm operators could provide basic skills training to engineers \& technicians on following a standardised pathway towards annotating, analysing and interpretation of information on turbine operational conditions in line with a common framework or industry standards which could be developed for O\&M based on consensus of multiple turbine operators globally. Secondly, it would likely be beneficial to encourage the adoption of data science and analytics techniques in the wind industry, by providing specialised resources in this domain (e.g. software applications with interactive graphical user interfaces (GUIs) \nomenclature{$GUI$}{Graphical user interface} to simplify the storage, annotation and analysis of SCADA data, failure logs and alarm messages) to engineers \& technicians, and supporting them with guidance and insights from data scientists. While the wind industry has invested heavily on some critical areas (e.g. design and manufacturing), we believe that there is insufficient investment in monitoring, development and analysis of datasets. By adopting unified standards and investing in this area, we envisage that the wind industry can benefit greatly in terms of Return on Investment (ROI) \nomenclature{$ROI$}{Return on Investment} , which can help train AI models for decision support with high quality datasets, making such models potentially more informative, accurate and scalable. 
    
    \item \textbf{Utilising specialised statistical and AI techniques for overcoming issues in data quality:} Data from turbines often consists of noise and outliers (e.g. power production at zero wind speed) resulting from communication failures, abnormality in equipments etc. \cite{Wu_2020}, which poses significant challenges in utilising such information to train AI models. Some studies have shown that applying specialised techniques to remove such noise and outliers can help make datasets more efficient, versatile and suitable for information mining. In one of the earliest demonstration of statistical techniques for robust data filtering, Llombart et al. \cite{llombart_robust} proposed the utilisation of a \nomenclature{$LMedS$}{Least Median of Squares} Least Median of Squares (LMedS) approach to detect noise and outliers prevalent in turbine power curves. The paper mentions that the approach outperforms other classical statistical methods for filtering (e.g. based on mean and standard deviation for binned segments of the data) and can help eliminate the requirements for manual filtering to reject outliers. Another similar study by Sainz et al. \cite{sainz_robust} combined the LMedS method with a random search approach, providing the ability to filter modelled data based on parameters beyond the wind speed, considering metrics like wind direction. In more recent studies, Shen et al. \cite{8330024} have shown that specialised algorithms such as change point grouping and quartile algorithm can help improve the quality of data pertaining to turbine power curves based on the outlier distribution characteristics. Some other studies have utilised filtering techniques based on popular methods in the statistics and control domains, such as Kalman filters to localise noise and outliers for wind energy assessment \cite{melero_kalmanfilter}, but these techniques are generally complex to apply and require extensive mathematical modelling. While data filtering approaches in the existing literature are suited to traditional AI models, especially for regression, they cannot handle other challenges posed by imbalanced datasets and lack of contextual information on failures as discussed before in our study. 
    
    To handle imbalanced datasets, oversampling techniques such as \nomenclature{$SMOTE$}{Synthetic Minority Over-Sampling technique} Synthetic Minority Over-Sampling technique (SMOTE) \cite{chawla_bowyer_hall_kegelmeyer_2002} have been successfully utilised in the wind industry in some studies \cite{9302680,8245530,windenergy_journal}. SMOTE is a highly popular statistical algorithm which can generate synthetic data points for data samples which belong to the minority class (e.g labelled records of anomaly), thereby balancing the overall distribution of majority (e.g. normal operation) and minority classes in the dataset. There are some other oversampling techniques which are popular in the AI community and in some cases, can be more efficient, especially \nomenclature{$ADASYN$}{Adaptive synthetic sampling}  Adaptive synthetic sampling (ADASYN) and \nomenclature{$RAMOBoost$}{Ranked minority oversampling in boosting}  Ranked minority oversampling in boosting (RAMOBoost) \cite{SAEZ2016164} but these are yet to be utilised in the wind industry for CBM to the best of our knowledge. We believe that the wind industry needs to focus on more widely adopting oversampling techniques, to facilitate informative decision making even in situations with limited and imbalanced data.
    
    For better uptake of NLG models, the optimal solution is to focus more on increasing the diversity of contextual information on faults, which would mean that the wind industry needs to focus not only on the critical types of failures discussed in our review, but also on low-priority faults. There are some options which could be possibly explored to facilitate the generation of human-intelligible O\&M policies with inadequate quality datasets. Firstly, the wind turbine alarm logs and maintenance records should be better organised, and regularly updated with time. Secondly, specialised NLG techniques such as few-shot learning \cite{chen-etal-2020-shot} can be utilised to facilitate NLG even in situations with limited availability of high quality training datasets. Thirdly, the wind industry can likely benefit from utilising generalized language models which have been pre-trained with billions of parameters, such as Bidirectional Encoder Representations from Transformers (BERT) and OpenAI's Generative Pre-trained Transformer (GPT-2/GPT-3), and have achieved state-of-art results in several downstream natural language processing and generation tasks \cite{shoeybi2020megatronlm,devlin-etal-2019-bert}. Such models can be fine-tuned with custom data from small corpuses and help overcome the challenges posed by inadequate availability (and quality) of alarm messages in the wind industry, as well as facilitate more fine-grained and comprehensive descriptions of faults and O\&M strategies to fix/avert faults in comparison to only brief alarm messages presently available in this domain. While some safety-critical domains such as for clinical decision support \cite{DEMNERFUSHMAN2009760} have significantly benefited from the better explainability and context which natural language messages and reports can provide, the wind industry needs to focus on optimally utilising every type of relevant and useful datasets to facilitate decision making under constraints of complex, unorganised and low-quality O\&M records in the short-term. This is necessary until better quality datasets are available in the wind industry, which we believe can likely only be achieved with a long-term perspective, given the challenges in transitioning from traditional data acquisition methods to high-quality storage and information retrieval, e.g. through cloud data centres. Besides these problems, the wind industry needs to deal with high resolution SCADA datasets (e.g. at 1 second intervals), missing values, and rapidly changing events and alarms, which is especially important for real-time decision making. Table ~\ref{dataquality_overcome} summarises the emerging data quality issues the wind industry is starting to witness in recent times, along with possible strategies to facilitate informative decision making under such circumstances.

\begin{longtable}{|p{3cm}|p{4cm}|p{6cm}|}
\caption{Emerging data quality issues in the wind industry affecting the development of AI models, and possible strategies to overcome these challenges \label{dataquality_overcome}} 
\\ \hline
\textbf{Data quality challenge} & \textbf{Affected AI techniques} & \textbf{Possible solutions}\\ \hline
Imbalanced datasets &
Supervised learning techniques for classification of faults; Reinforcement learning techniques for O\&M planning; Natural language processing techniques for classifying alarm messages  & Utilising oversampling techniques for balancing imbalanced class distributions \cite{SAEZ2016164}  e.g. Synthetic Minority Over-Sampling technique (SMOTE) \cite{chawla_bowyer_hall_kegelmeyer_2002}, Adaptive synthetic sampling (ADASYN) \cite{4633969}, Ranked minority oversampling in boosting (RAMOBoost) \cite{5559472}, \nomenclature{$DC GAN$}{Deep convolutional generative adversarial networks} Deep convolutional generative adversarial networks (DC GAN) \cite{8483334} etc.    \\ \hline
Lack in diversity of alarm messages &
Natural language generation techniques for generating contextual information on faults   & Few-shot learning techniques \cite{chen-etal-2020-shot} to learn from low-diversity and limited training data; Generalized language models pre-trained on large corpuses of information such as \nomenclature{$BERT$}{Bidirectional Encoder Representations from Transformers} Bidirectional Encoder Representations from Transformers (BERT) \cite{devlin-etal-2019-bert},  \nomenclature{$GPT$}{Generative Pre-trained Transformer} Generative Pre-trained Transformer (GPT-2/GPT-3) \cite{radford2019language,DBLP:conf/nips/BrownMRSKDNSSAA20} etc.\\ \hline
Low-quality datasets with noise, corrupted values and outliers &
Supervised and unsupervised classification (for fault prediction) and regression (for forecasting vital operational parameters) techniques; Reinforcement learning techniques for O\&M planning;  & Change-point grouping and quartile algorithms \cite{8330024}, Least Median of Squares (LMedS) method \cite{llombart_robust}, LMedS with random search \cite{sainz_robust}, statistical and control filtering techniques like Kalman filters \cite{melero_kalmanfilter}, specialised loss functions in deep learning, data re-weighting and training procedures \cite{KARIMI2020101759}, class noise and attribute noise identification techniques (especially ensemble-based noise elimination) \cite{GUPTA2019466}, specialised ML-based noise reduction techniques such as Multi-step finite differences, Splines, Mixture of sub-optimal curves etc. \cite{Minutti_2018}  \\ \hline
Missing values in datasets &
Supervised and unsupervised classification (for fault prediction) and regression (for forecasting vital operational parameters) techniques; Reinforcement learning techniques for O\&M planning; Natural language processing techniques for classifying alarm messages; Natural language generation techniques for generating contextual information on faults & Statistical imputation techniques \cite{JEREZ2010105} to replace missing values with substituted values based on other available data e.g. using measures of central tendency like mean/median, \nomenclature{$kNN$}{K nearest neighbours} K nearest neighbours (KNN) imputation  \cite{10.1007/s10489-015-0666-x}, Self-organizing maps imputation for incomplete data matrices \cite{FOLGUERA2015146} etc. \\ \hline
\end{longtable}
    
\end{itemize}

 \begin{figure}[!h]
\centerline{\includegraphics[width=0.8\textwidth]{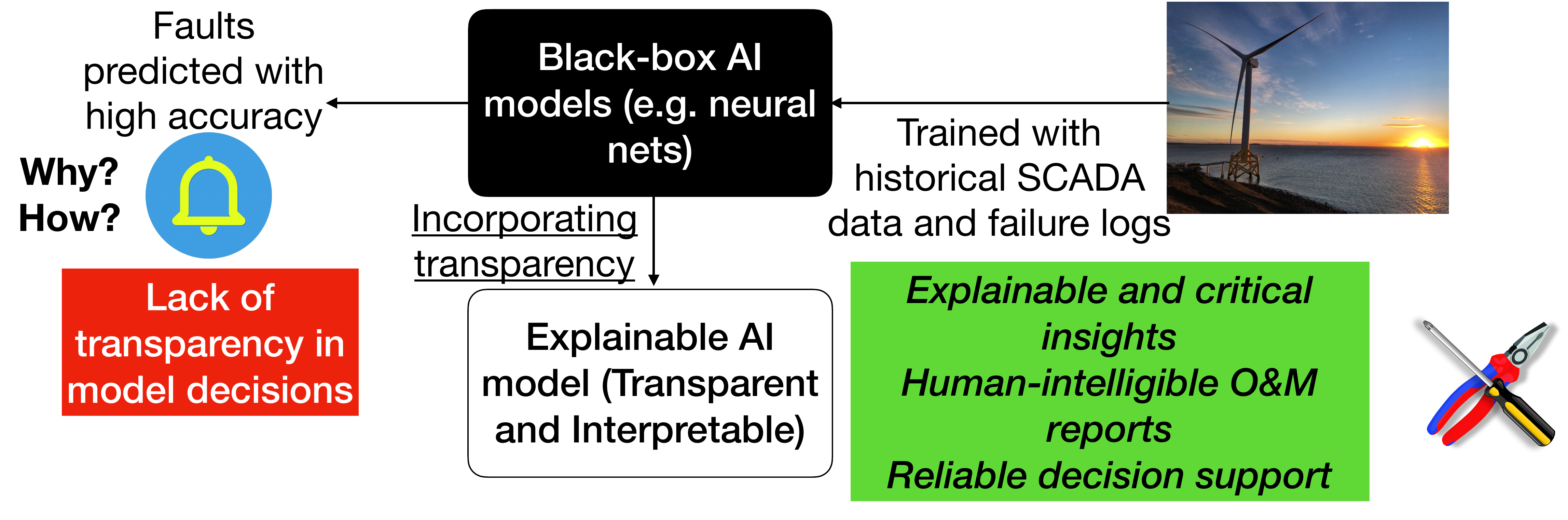}}
\caption{Explainability challenge for AI in the wind industry: black-box natured AI models can generate predictions with high accuracy, but fail to provide rationale behind their decisions. Explainable AI techniques can help tackle this challenge. \label{challenge_xai}}
\end{figure}

\subsection{\textbf{Challenges in deploying highly sophisticated O\&M models for real-time decision support}}
While our study shows that some promising efforts have been made in the wind industry for deploying decision support models in real-time environments \cite{realtime_reconstruction,abdallah_autodecision,7998308}, these are clearly rare in comparison to the significant research which has been pursued in the development of AI models. More importantly, the existing studies which deploy decision making models in real-time mostly utilise more traditional and simpler models such as decision trees, support vector regression and random forest, which despite being promising are clearly insufficient for present demands of the wind industry, which is experiencing an enormous rise in complexity of big data.  
With the global advent of \nomenclature{$IoT$}{Internet of Things} Internet of Things (IoT), some studies have outlined techniques which can be utilised for interfacing turbine sensors and actuators with the internet and cloud. An early work in this area by Kalyanraj et al. \cite{8052714} proposed the utilisation of IoT technology to facilitate turbine control, as well as data logging of vital parameters such as power generation and vibration levels. In another notable study, Alhmoud and Al-Zoubi \cite{IoTApplicationsinWindEnergyConversionSystems} proposed a framework for utilisation of IoT platforms for each turbine in a wind farm, which can be connected using a microcontroller to a cellular network. Thereby, any new data which is available can be saved and processing performed on the cloud servers, which also facilitates the access of such information from anywhere in the world through e.g. computers and mobile devices, wherein, suitable commands can also be provided to turbines in the wind farm. While this is a promising framework for gathering real-time data from sensors for performance optimisation and can also help in identifying O\&M activities, the paper mentions that the key factors limiting the practical realisation and wider deployment of IoT in the wind industry are lack of budget and necessary skills. Other barriers which the paper mentions include security concerns, challenges with communication protocols etc. While few such studies have clearly shown the immense potential of IoT for real-time decision making, they do not have a specific focus on AI and the challenges in deploying such models for O\&M tasks.

We believe that presently, the availability of continuous flow of information e.g. in the form of SCADA features is not a major challenge for real-time machine learning, as most wind farm operators have placed immense emphasis on developing effective and efficient data logging and processing systems on cloud servers, as is reflected by our reviews in this paper. However, the major challenge which the wind industry faces is the growing complexity of such datasets, which, due to lack of unified standards and simplified formats are difficult to be utilised for inference with trained AI models. Also, the increasing availability of high resolution data (e.g. at 1 second intervals) instead of the conventionally popular 10 minutes intervals creates additional constraints on feeding SCADA features into the AI models, especially for facilitating continual updates and re-training the model as new observations become available. We believe that the pressing issues which are holding the wind industry back from utilising AI models (especially deep learners) for real-time decision making are primarily inadequate computing power, growing memory cost and security/privacy concerns. While the scale of such challenges can be reduced by utilising e.g. deep learning models with fewer hidden layers, this would also generally limit their accuracy on complex tasks, as the ability of such networks to go deeper or wider is the key essence of their immense potential, especially in extreme-scale DNN models \cite{10.1145/3243904}. Thereby, only utilising simpler/shallower models is clearly not a viable solution to tackle these challenges, as it would generally lead to a trade-off with the model performance, which is often critical for O\&M tasks in the wind industry.

Table ~\ref{realtime_overcome} outlines some possible strategies which can be utilised by the wind industry to overcome the rising challenges in deploying sophisticated AI models for real-time decision support.

\begin{longtable}{|p{3cm}|p{4cm}|p{6cm}|}
\caption{Emerging challenges for real-time deployment of AI models in the wind industry, and possible strategies to overcome these challenges \label{realtime_overcome}} \\
\hline
\textbf{Real-time deployment challenge} & \textbf{Description} & \textbf{Possible solutions}\\ \hline
Memory cost constraints & AI models, especially deep learners face high bandwidth memory requirements; Memory usage during training is especially dominated by the need for intermediate activation tensors for storing temporary information during backpropagation \cite{MLSYS2020_084b6fbb}
  & In-memory computing to perform forward and backward pass in neural networks in place without the requirement to move around weights \cite{inmemory_computing} etc. , memory efficient deep learning frameworks for large-scale data mining \cite{largescalemem_frame} e.g. MXNet \cite{chen2015mxnet}, memory efficient adaptive optimisation method \cite{NIPS2019_9168}, specialised frameworks for developing memory efficient invertible neural networks e.g. MemCNN in PyTorch \cite{vandeLeemput2019MemCNN}, automatic efficient management of GPU memory by using specialised techniques e.g. computational graphs for models with swap-out and swap-in operations for holding temporary results in CPU memory \cite{10.1145/3315573.3329984} etc. \\ \hline
Inadequate computing power & Recent advances in AI, especially deep learning have led to models which often utilise tens of millions of parameters for tasks pertaining to real-time processing of data streams \cite{37631}; deep learning models require substantial computational resources during training and inference phases to run in a quick manner \cite{8763885}
  & Using dedicated hardware for ML with \nomenclature{$HPC$}{High Performance Computing} High Performance Computing (HPC) platforms such as \nomenclature{$GPU$}{Graphics Processing Unit} Graphics Processing Units (GPUs) and \nomenclature{$TPU$}{Tensor Processing Unit} Tensor Processing Units (TPUs) \cite{1575717,wang2019benchmarking,CHEN2020264}, cost-efficient training mechanisms for fast model training such as PruneTrain \cite{10.1145/3295500.3356156}, resource constrained structure learning for deep networks \cite{8578269}, utilising edge computing techniques during deep learning to accomplish  low-latency and high computational efficiency \cite{8763885}; using cloud computing platforms \cite{8279419} such as Google Cloud AI, Amazon Web Services, Azure Machine Learning, IBM Watson Machine Learning etc. \\ \hline 
  
 Concerns on communication security and privacy & Real-time decision support systems face risks of security concerns during data streaming; ML model policies can be interfered with malicious attacks when performing real-time control in dynamic environments \cite{8455947}; AI models can be subject to adversarial attacks during training/testing phases \cite{adv_attack}; Wind farm SCADA systems can be subject to cyber attacks and intrusion \cite{5772593,8591200}
  & Secure learning approaches for defense against training and inference time attacks \cite{8406613}, encryption and secure coding of data streams e.g. through \nomenclature{$LDPC$}{Low-density parity check } Low-density parity check (LDPC) \cite{ldpc_handling} etc. \\ \hline 
\end{longtable}

\subsection{\textbf{Lack of transparency in the black-box natured AI models:}} 
Evidently from our discussion, while most AI models are able to provide highly accurate predictions (e.g. of turbine power output and different faults), they continue to face a significant challenge of transparency due to their inherent black-box nature. This phenomenon is depicted in Figure~\ref{challenge_xai}. Additionally, while conventional ML techniques (such as decision trees) provide added transparency and are much simpler to interpret \cite{sezer2019financial}, they are generally significantly outperformed by deep learners. The lack of rationales behind decisions made by the AI models, in turn makes wind farm operators reluctant to adopt data-driven decision making techniques and focus on more traditional methods based on signal processing and numerical physics models. We believe that it is essential to incorporate trust in the decisions made by these black-box learners, and switching from black-box AI to Explainable AI as discussed below.
\begin{longtable}[ht]{|p{3cm}|p{7cm}|p{2cm}|}
\caption{Summary of Explainable AI models relevant to wind turbine CBM and performance assessment/analysis. For models which have not been applied till date, prospective applications are outlined.\label{xai_models}} \\
\hline
\textbf{Explainable AI Model} & \textbf{Description} & \textbf{Applicability to wind turbine CBM and performance assessment/analysis}\\ \hline
\nomenclature{$xDNN$}{Explainable Deep Neural Networks} Explainable Deep Neural Networks (xDNN)  \cite{ANGELOV2020185} & A non-iterative and non-parametric deep learning architecture, combining reasoning and learning in a synergy. Provides explanations based on probability density function automatically learnt from training data distribution & Can prospectively be applied towards fault prediction in turbine sub-components, anomaly detection in blade images, predicting vital SCADA parameters (e.g. wind speed and power output)\\ \hline
Long short-term memory networks (LSTMs) with attention \cite{luong_pham_manning_2015} & The attention mechanism allows LSTMs to focus on vital parts of input sequences, providing easier and higher quality learning; Attention weights can provide transparency in key features which cause LSTM to generate its predictions & Wind power prediction \cite{9041585}   \\ \hline
Convolutional neural networks (CNNs) with attention \cite{zheng_fu_mei_luo_2017}  & The attention mechanism provides ability to focus on vital segments of input data in the CNN layers and convolutional filters, with attention weights providing explainablity in predictions. & Short term wind speed prediction \cite{en13071772}, imbalance fault detection in turbine blades \cite{en12142764}, causal inference for discovering novel insights and hidden confounders \cite{torque_paper}   \\ \hline
SHapley Additive exPlanation (SHAP)  \cite{NIPS2017_7062} + Any Black-box AI model & Provides explainations for outputs generated by any ML model based on local explainations through game theory approach ; provides force plots and interpretable explanations of decision trees/ensembles of trees & Fault prediction in multiple turbine sub-components \cite{renew_paper}  \\ \hline
\nomenclature{$LIME$}{Local Interpretable Model-Agnostic Explanations} Local Interpretable Model-Agnostic Explanations (LIME) \cite{10.1145/2939672.2939778} + Any Black-box AI model & Provides local linear explainations for the ML model's behavior; can be utilised for explainable classification tasks with 2 or more classes  &  Prospective applications include explainable binary/multi-class anomaly prediction in turbine sub-components, classification of blade images, segmentation of alarm messages \\ \hline
Sequence-to-sequence \nomenclature{$Seq2Seq$}{Sequence-to-sequence} (Seq2Seq) model with attention \cite{luong-etal-2015-effective} & Specialised recurrent neural network architecture for sequential data; incorporates attention mechanism to focus on vital parts of input sequential data; can provide transparency in identifying relevant features used during the prediction process & Wind power forecasting \cite{8955569}, prediction of alarm messages \cite{wcci_paper} \\ \hline
Transformers \cite{NIPS2017_7181}  & Utilises multi-head attention mechanism and removes recurrence in the conventional encoder-decoder Seq2Seq architecture; can provide transparent decisions in terms of key features which lead to generated predictions through self-attention scores ;more computationally efficient than Seq2Seq models & Short-term load forecasting \cite{en12244612}, prediction of alarm messages and maintenance actions \cite{wcci_paper}   \\ \hline
eXtreme Gradient Boosting (XGBoost) \cite{10.1145/2939672.2939785}  & A novel, sparsity-aware tree boosting algorithm which can provide feature importances in datasets utilised for predictions; highly computationally efficient and scalable  & Fault detection in multiple sub-components \cite{8329419,windenergy_journal,renew_paper,app10093258}, wind power forecasting \cite{7981134}, gearbox fault prediction \cite{RePEc:gam:jeners:v:12:y:2019:i:22:p:4224-:d:283963}   \\ \hline

\end{longtable}
\paragraph{\textbf{Utilising Explainable AI models:}}
For tackling the lack of transparency in black-box AI models, \nomenclature{$XAI$}{Explainable AI} Explainable AI (XAI) \cite{BARREDOARRIETA202082} models provide immense promise in responsible, trustworthy and dependable decision-making. XAI can contribute to improved performance of AI models as explanations help trace issues and pitfalls in datasets and the behaviour of features, while also assisting engineers \& technicians to better trust predictions made by such models. The wind industry has seen very limited applications of XAI, with only a few studies applying such techniques for CBM and performance assessment/analysis. The current applications mostly focus on turbine power prediction, while limited attention has been received in the area of predicting faults and maintenance actions in O\&M. Table ~\ref{xai_models} summarises some of the major XAI models which have been applied in the wind industry, along with other prospective models for potential future applications.

\begin{figure*}[h]
\centerline{\includegraphics[width=0.6\textwidth]{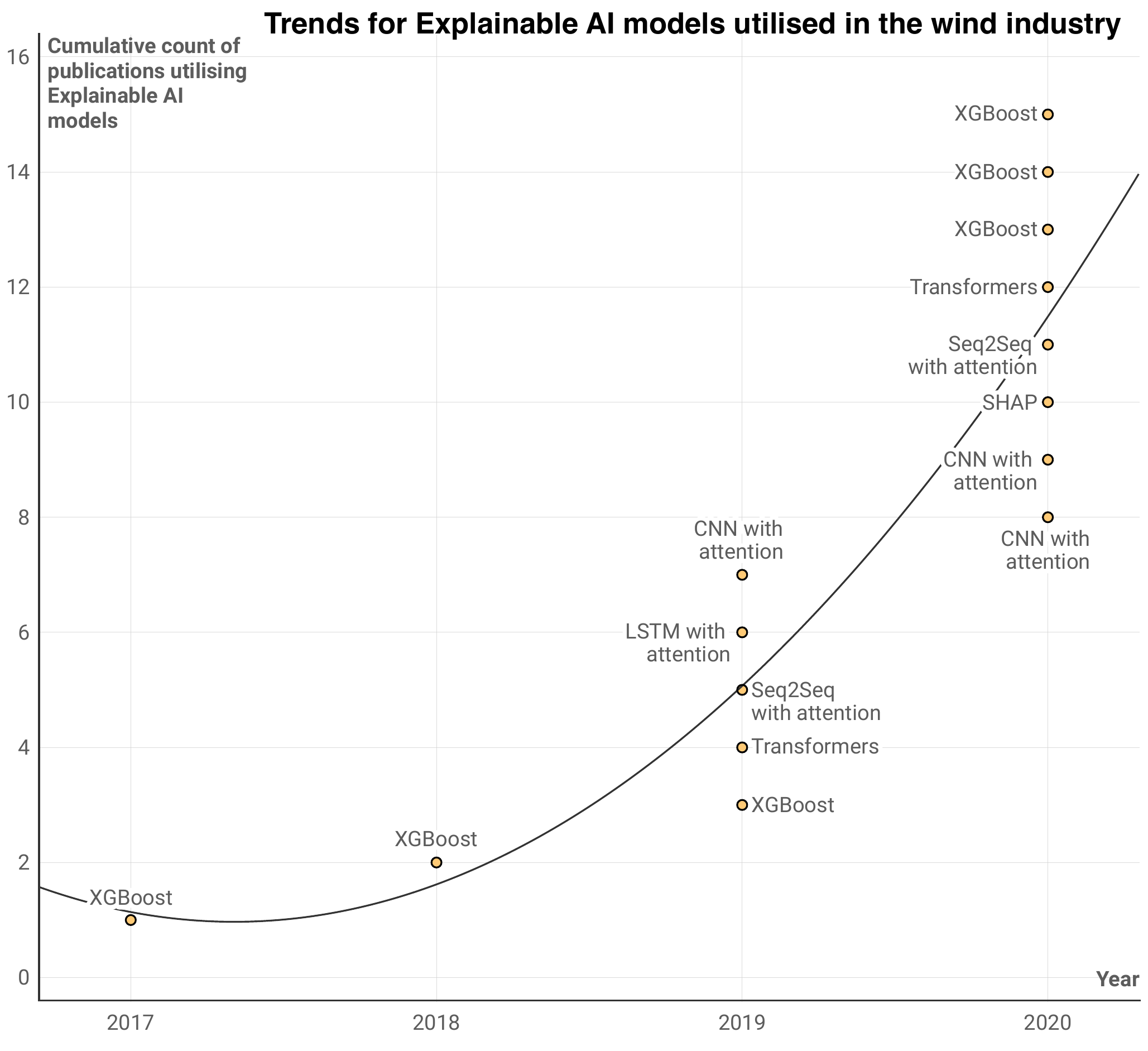}}
\caption{Trends for applying Explainable AI models for data-driven decision making. The slow and static growth is clearly visible. \label{xai_trends}}
\end{figure*}

Figure ~\ref{xai_trends} shows the trends for utilising various Explainable AI models for data-driven decision making in the wind industry, outlining the slow growth in this direction. Evidently, while explainable decision tree algorithms (such as XGBoost), specialised libraries and packages for incorporating transparency (such as SHAP) and CNNs with attention mechanism have received comparatively greater attention in the wind energy domain, several other techniques, especially those utilising LSTMs with attention, Sequence-to-sequence (Seq2Seq) models and Transformers are facing paucity for application to data-driven decision support. This is likely due to the lack of available insights, as well as the reluctance to apply AI in the wind industry, which we hope our paper can tackle. We believe that there is great potential for such models to be applied, in particular natural language generation techniques, given that besides making accurate predictions e.g. of turbine alarm messages, they can also generate feature importances for the causes contributing to such failures, alongside human-intelligible descriptions of most appropriate maintenance actions to avert/fix failures in such scenarios \cite{wcci_paper}. There are possibly a plethora of applications of XAI for data-driven decision making, which we believe need to be explored in the near future to support human-intelligible diagnosis and prognosis of operational inconsistencies in wind turbines. Appendix ~\ref{table_papersummary} provides a summary of all papers which we have reviewed in our study, including the techniques they use, key applications and findings, along with their limitations wherever applicable. This can help serve as a ready reference for 
researchers in the wind industry to utilise AI pertaining to CBM and performance assessment of turbines.

\begin{figure*}[!h]
\centerline{\includegraphics[width=0.8\textwidth]{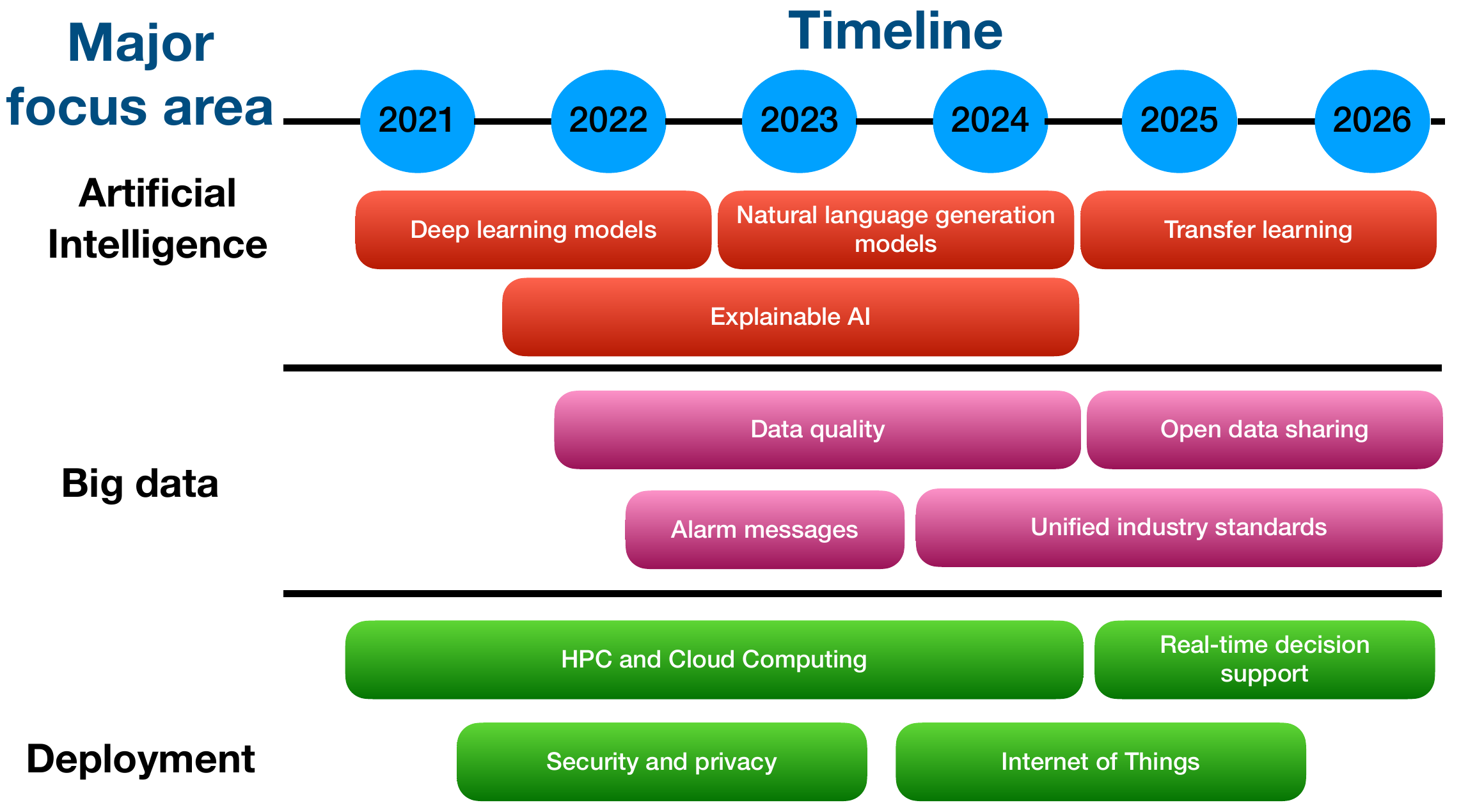}}
\caption{Graphical roadmap outlining the likely \textit{major} focus areas in the wind industry based on current trends, successes and challenges. First appearance of specific topics shows that they would receive majority focus at that time. \label{roadmap_figure}}
\end{figure*}

\section{Discussion}\label{discussion}
Figure ~\ref{roadmap_figure} shows a graphical roadmap summarising the likely future of utilising AI for decision support in O\&M in the wind industry over the upcoming 5 years. Note that while we cannot be certain of the definite future, the current successes and challenges outlined in this paper alongside the growing focus on AI in the wind industry shows a very likely promising future in this avenue. More details on these focus areas are outlined in Table ~\ref{roadmap_table}. 

\begin{table}
\caption{Summary of future roadmap likely viable for utilising AI in the wind industry \label{roadmap_table}} 
\begin{tabular}{|p{3cm}|p{10cm}|}
\hline
\textbf{Major focus area} & \textbf{Description} \\ \hline
Deep learning models &	Wider adoption of deep learning models for CBM (especially RNNs and CNNs)  \\ \hline
Explainable AI &	Growing focus on XAI, with transition from XGBoost to other transparent learners (e.g. transformers with attention mechanism) \\ \hline
Alarm messages	& Better availability of alarm messages with contextual information of faults \\ \hline
Natural language generation models	& Uptake of NLG models in "few" wind farms for real-time human intelligible decisions \\ \hline
Data quality	& Better quality data available with improved sensors and automated processing of alarms on cloud \\ \hline
HPC and Cloud Computing	& Wider adoption of HPC in the wind industry, especially GPUs and TPUs through cloud computing \\ \hline
Transfer learning & Transfer learning sees immense utilisation with rise in development of new wind farms which lack sufficient operational data to train AI models from scratch  \\ \hline
Open data sharing	& Increased open-sharing of datasets for CBM, especially with historical alarm and failure information \\ \hline
Unified industry standards	& Unified industry standards developed in the wind industry pertaining to standard taxonomies and data formats \\ \hline
Internet of Things	& IoT becomes widely popular in the wind industry, with uptake for real-time control and O\&M of wind farms from anywhere in the world \\ \hline
Security and privacy	& Security and privacy challenges are significantly overcome through adoption of AI models after critical testing for adversial attacks; SCADA systems become more secure to intrusions \\ \hline
Real-time decision support	& Highly sophisticated AI models are deployed widely by wind farm operators across the globe for real-time decision making in O\&M activities \\ \hline
\end{tabular}
\end{table}

We have segmented the roadmap into 3 major subgroups i.e. Artificial Intelligence (for general adoption of AI in the wind industry), Big Data (for tackling data availability and quality issues) and Deployment (for final deployment of AI models in real-world industrial use cases), wherein, each of these subgroups have relevant specific topics (e.g. deep learning in the AI subgroup) associated with them. Note that the roadmap shows the major focus areas for the wind industry over the next few years, and the first time new topics arise in the roadmap (e.g. data quality in 2022) indicates the likely high priority they would receive at that point of time. Once these highlighted topics disappear (e.g. data quality is given a timeframe till 2024), they would continue to be important but would likely receive lesser priority in the wind industry. Our insights are based on the general delay between the time an associated method is first published in the AI community to the time it takes to be utilised in the wind industry. For instance, the transformer model for NLG  \cite{NIPS2017_7181} was first published in 2017 in the AI community, but it was only utilised in the wind industry for short-term load forecasting \cite{en12244612} in 2019 and generation of human-intelligible alarm messages \cite{wcci_paper} in 2020. This clearly indicates that such methods are generally adopted 2-3 years after their first appearance in the AI community, with this time delay being an important factor to consider. From the roadmap, it can be seen that while we envisage that there would be an early transition to deep learning and NLG models in the short-term, along with a growing focus on adoption of XAI models to achieve transparency in O\&M decision making, the wind industry would take more time to improve the data quality and adopt HPC \& Cloud Computing techniques widely. Also, transfer learning techniques will likely see an immense growth in the time to come, and the competitive nature of the wind industry would possibly lead to open data sharing in the next few years, given that some wind farm operators have already taken the positive steps to share datasets openly. However, we believe that adoption of IoT for real-time decision support will likely not be realised soon, as there are ongoing challenges and concerns on data security and privacy which need careful consensus and thorough analysis in the wind industry. 

If the current trends in growth of employing AI models for data-driven decision making continue at the same pace in the wind industry, we believe that it would not be an impossible feat to achieve real-time decision support across most wind farms globally by the end of next 5 years (by 2026). This, surely is based on cautious optimism and the chances for an AI winter to again prevail in the wind industry. However, we would enunciate that following such a roadmap would likely lead to immense savings in O\&M costs to wind farm operators, and a significantly wider adoption of wind energy globally in the route to combat climate change.

\section{Conclusion}\label{conc}
We have provided a systematic review of the past, present and future of data-driven decision making techniques in the wind industry through scientometric analysis using statistical computing techniques. By tracing the thematic and conceptual structure of CBM and performance assessment/analysis through evidence-based insights, we demonstrate there is a significant interest in applying AI techniques for decision support, especially deep learning. An important insight from our study is that despite the growth of AI in the wind industry, more traditional techniques such as those based on signal processing will continue to complement AI models in this rapid transition. Our study shows that the AI applied in the wind energy domain is still in its embryonic stages compared to advances that other disciplines, such as computer vision and NLP, have made in this area. We outline the key challenges faced by the wind industry in widely adopting data-driven decision making techniques, particularly lack of access to quality data, problems in deploying AI models for real-time decision support, and the issue of transparency in black-box natured AI models. To overcome these challenges, we show that is vital to focus on more sophisticated and tailored AI algorithms, especially utilising deep learning and natural language generation techniques for explainable AI in achieving human-intelligible, transparent and trustworthy decision making. We envisage that this paper can encourage wind energy researchers to specifically focus on critical areas mostly neglected in the past, helping in a smoother transition to AI from academic labs to the wind industry. 

\section*{Declaration of competing interest}
The authors declare that they have no known competing financial interests or personal relationships that could have appeared to influence the work reported in this paper.

\section*{Funding source}
This research did not receive any specific grant from funding agencies in the public, commercial, or not-for-profit sectors.

\section*{Data Availability}
The datasets used in this article are publicly available, and can be found on Github at\\ https://github.com/joyjitchatterjee/ScientometricReview-AI.

\section*{Acknowledgement}.
We would like to acknowledge the Offshore Renewable Energy Catapult (ORE Catapult) for providing operational data from the Levenmouth Demonstration Turbine through Platform for Operational data (POD) \footnote{Platform for Operational Data: https://pod.ore.catapult.org.uk \label{orec_reference}}, which helped us understand the challenges and opportunities in data-driven decision making for the wind industry. We are also grateful to the Aura Innovation Centre, UK and the University of Hull for their support.

\appendix
\section{Summary of reviewed publications}\label{table_papersummary}

\begin{small}

\begin{longtable}{p{2cm}|p{3cm}|p{4cm}|p{4cm}}
\caption{Summary of reviewed techniques utilised in the wind industry for CBM and performance assessment of turbines pertaining to O\&M. The relevant papers are arranged mainly based on the order of their occurrence in our reviews, with similar methods placed closer to facilitate coherence.} \\
\toprule
\textbf{Relevant Paper(s)} & \textbf{Utilised Technique} & \textbf{Applications and Findings} & \textbf{Limitations (if any)} \\ \midrule
Zimroz et al. (2012) \cite{10.1007/978-3-642-28768-8_52}                           & Vibration analysis through statistical processing  & Abnormal behaviour of turbine bearings identified during non-stationary load/speed situations; Vibration signals used to identify deviations from ideal parameter values &                      \\
Liu (2013) \cite{LIU2013954}                & Vibration analysis through Fourier transform and other probabilistic techniques & Physics-based techniques used to estimate wind force in turbine's blade-cabin-tower system; Can help identify random wind vibrations and its effects & Only mathematical framework presented with no demonstration of the method on real-world data \\
Feng et al. (2012) \cite{RePEc:eee:renene:v:47:y:2012:i:c:p:112-126} & Demodulation analysis of vibration signals & Abnormalities in gearbox operation identified using planetary gearbox vibration signals; Accounts for IMFs produced by using an ensemble EMD method; Proposes comparison of demodulated signal envelope with ideal theoretical values for fault detection & \\ 
Li (2010) \cite{5673197} and Yang et al. (2011) \cite{2011JSV...330.3766Y} & EMD for vibration signal analysis & Can help identify incipient faults in mechanical and electrical turbine sub-components; Key idea is to decompose signals into multiple IMFs & \\
Zheng et al. (2013) \cite{kalman_pf}  & EMD for signal decomposition with RBFNN as prediction model & Short-term forecasting of wind power performed; Statistical control algorithms like Kalman filtering used to eliminate noise & \\
Saidi et al. (2015) \cite{7505124}   & Signal processing of vibration signals in frequency domain utilising SK & Squared envelope technique based on SK utilised to diagnose skidding in high-speed shaft bearings; SK utilised to identify severity of damage as well as transients in signal; Experiments demonstrated with real-world vibration data from high-speed shaft bearings in normal zone, degradation zone and failure zone
& \\
Jia et al. (2015) \cite{jia_lei_shan_lin_2015} & Signal processing of vibration signals in frequency domain utilising MCKD & Fault diagnosis performed in rolling-element bearings; Can help clarify periodic fault transients prevailing in noisy signals & Identified faults cannot be validated based on performance metrics (e.g. accuracy); The method cannot estimate vital O\&M parameters like RUL, MTTF etc.\\
Guo et al. (2010) \cite{5554606} & DWT for analysing vibration signals in frequency domain &  Gear faults identified by utilising vibration acceleration signal; DWT helps better characterise time-varying components in signal and its energy distribution, for easier fault diagnosis & \\
 
Yang and An (2013) \cite{doi:10.1155/2013/212836}  & EMD with wavelet transform for vibration analysis & Wavelet transform used to analyse vibration signals, and EMD utilised for better decomposition of signals into IMF components; Can address aliasing in signals to provide more thorough prediction of instantaneous signal frequency & Wavelet transform can be highly computationally intensive for fine-grained analysis; Requires careful consideration of parameters for shifting, scaling etc. \\
Clifton et al. (2013) \cite{Clifton_2013} & Regression trees for performance assessment & Aerostructural simulations data used to forecast turbine power output; Accounts for wind speed, turbulence and shear parameters & \\
Pravilovic et al. (2014) \cite{Pravilovic_2014} & Time-series cluster analysis & Power forecasts analysed under circumstances of normal operation and anomaly & \\
Yang et al. (2015) \cite{Yang_2015}  & SVM enhanced Markov models & Provides short-term wind power forecasts; Improved forecasts in comparison to models only utilising SVM & \\
Chen et al. (2014) & Gaussian processes and NWP modelling & Provides up to one-day ahead wind power forecasts; GP provides correction on wind-speed data; Significant improvement in forecasting accuracy compared to ANNs  & \\
Soleymani et al. (2015) \cite {Soleymani_Mohammadi_Rezayi_Moghimai_2015}   & Hybrid modified firefly algorithm & Turbine power outputs forecasted with consideration of confidence interval for predictions; Probabilistic and statistical inference used for prediction making, which is simple to apply and interpret & Generally outperformed by more sophisticated approaches (especially deep learning) \\
Clifton et al. (2014) \cite{wind_mountainpass} & Decision trees coupled with regression for performance assessment & Turbine performance predicted in a mountain pass region; Forecasts made on effects of pass wind time-series on turbine performance & \\
Yang et al. (2019) \cite{realtime_reconstruction}  & Support vector regression for reconstruction-based ML & Real-time fault detection performed in turbine sub-components based on residual errors in SCADA signals; Simple to apply and interpret for CBM & Such methods do not leverage historical turbine data for decisions; Mathematical modelling of reconstruction models is computationally expensive \\
Park et al. (2013) \cite{multivariate_windfield}  & GMM and GDA models for structural health monitoring & Wind field characteristics data utilised for load response predictions; Can be extended to new turbine sites by utilising wind resource assessment data & \\
 Du et al. (2016) \cite{paper5} & Pearson correlation coefficient and self-organising map for CBM utilising power curve & Parameter selection performed for modelling turbine behaviour with correlation coefficient; Self-organising map for dimensionality reduction of SCADA features; Potential faults identified based on SCADA data points falling of ideal power curve; Promising and simple to apply & Reliance on power curve for anomaly prediction makes the method less competent and robust \\
 
Morshedizadeh (2017) \cite{paper3}  & MLP with ANFIS model & Wind turbine power predicted by utilising historical turbine performance data; Hybrid model with MLP and ANFIS can generate optimal power production predictions & Generally outperformed by more sophisticated models (especially deep learning) \\
Quereshi et al. (2017) \cite{QURESHI2017742} & Ensemble model of deep auto-encoders and Deep Belief Networks for predicting vital operational parameters & Significantly outperforms conventional regression methods for turbine power prediction; Can facilitate transfer learning with power predictions in absence of additional training data \\
 Kulkarni et al. (2019) \cite{doi:10.1177/0954406218797972} & LSTMs for long-term wind speed forecasting & Can help in fatigue analysis of turbine blades; RNNs are shown to be promising for dynamic wind load estimation & \\
Zhu et al. (2017) \cite{st_windpower} and Liu et al. (2019) \cite{liu_guan_hou_han_liu_sun_zheng_2019} & LSTMs for short-term wind power forecasting & Outperforms ANNs and SVMs utilised in existing studies significantly & \\
Leahy et al. (2016) \cite{paper1} & SVM for classification & SCADA data from turbine used to predict incipient faults across multiple turbine sub-components with promising results; Focus on filtering and analysis of faults and alarms, in conjunction with power curve & Only binary classification is performed; No feature selection and dimensionality reduction performed \\
Si et al. (2017) \cite{data_randomforest} & Random forest classifier with PCA & Faults identified in multiple turbine sub-components with identification of dominant SCADA signals; Provides feature importance of SCADA signals assisting in fault analysis; Algorithm used can directly be fed with large datasets and having very efficient training time & \\
Canizo et al. (2017) \cite{7998308}  & Big data frameworks (Apache Kafka, Apache Spark, Apache Mesos and HDFS) with random forest learning model & SCADA data stream processing performed on cloud; Facilitates real-time decision making through an online fault tolerant monitoring agent; Complete solution right from model development to cloud server deployment with a front-end dashboard & No provision for online updates in the trained model \\
Abdallah et al. (2018) \cite{faultdiag_decisiontree} & CART, specifically ensemble bagged trees & Root-cause analysis of predicted faults in multiple turbine sub-components performed; Provides brief description of SCADA feature values leading to faults & No validation of performance metrics (e.g. accuracy and prediction speed); For temporal SCADA data, other models like RNN and ARIMA can generally give more reliable predictions \\
Abdallah et al. (2018) \cite{abdallah_autodecision} & Decision tree interfaced with distributed storage cloud server & Can provide real-time analytics and autonomous decision support Provides a complete hardware-software solution for detecting faults in real-time, with simple and easy to interpret decision tree model & Only a conceptual framework is presented without experimental demonstration on real-world data; More sophisticated models (e.g. deep learners) would likely face more challenges in interfacing with cloud servers \\
Lu et al. (2018) \cite{opp_cbm}  & ANNs for CBM & Can help predict life percentage of turbine sub-components; Faults can be identified based on conditional probability of failures; Utilises the ANN model predictions and historical component failure time distribution for fault detection; Can help in better inventory planning & No details on training and test performance of ANNs, and the rationale for choice of network architecture; Only focus on 4 sub-components (pitch system, gearbox, generator and rotor)  \\
Qian et al. (2015) \cite{cbm_elm} & ELM for CBM & Provides better performance for fault identification in comparison to feedforward neural networks;  Faults identified based on deviation from ideal SCADA signals; Model takes significantly less time for training and inference; Can predict faults in specific sub-components & No details presented on the ELM model architecture; Lacks in comparing performance of ANNs and ELM based on accuracy and training time; Deviation from ideal signals may not always indicate fault, so prone to false alarms \\
Li et al. (2014) \cite{image_turblade1}, Li et al. (2015) \cite{image_turblade2} and Moreno et al. (2018) \cite{vision_turblade} & CNNs for computer vision based CBM & Can provide high accuracy in identifying faults with turbine blade images & Cannot be used for predicting anomalies in internal turbine sub-components\\
Yu et al. (2017) \cite{image_turblade} & SVM-CNN hybrid model for computer vision based CBM & Have shown success in learning to predict faults in turbine blades with small labelled datasets & Drones/other image capturing devices are not cost-effective; Method prone to failures and false alarms e.g. during rain, mist and snow \\
Wu et al. (2014) \cite{windturlayout_1} and Menghua et al. \cite{windturlayout_2} (2007) & Turbine layout planning methods & Can help decide optimal positions for turbines to maximise performance; Helps in suitable performance assessment for deployment of turbines in new sites & \\
Dutta et al. \cite{windturlayout_3} & Genetic algorithms and ant colony optimisation algorithm & Wake effect considered and wind speed time series and cable parameters towards turbine interconnections utilised to optimise turbine layout; Can help cover exponential number of cases and distributions to find optimal solutions & \\
Zaher et al. (2009) \cite{online_faultdetection}  & Multilayer neural networks along with a MAS architecture & SCADA data utilised for temperature anomaly detection in sub-components like gearbox and generator; Shows immense promise of deep learning for CBM and specific-component based anomaly prediction & No historical fault logs utilised for validation of the method; Still requires manual intervention of professional engineers/technicians to identify and classify different faults in sub-components  \\
Andersen et al. (2015) \cite{scalable_faultdetection} & CNNs for CBM with vibrational signals & Vibration data pertaining to main bearing failures from multiple turbines used to provide scalable fault detection technique; CNN significantly outperforms conventional ML baseline models & \\
Ibrahim et al. (2016) \cite{neuralnet_csa} & Current signature analysis along with ANN & SCADA data utilised with promising results for anomaly prediction; Current signature analysis is able to identify faults based on frequency spectrum of electrical signals; ANNs help to identify faults during transient conditions &  \\
Pang et al. (2020) \cite{pang_he_jiang_xie_2020} & Hybrid spatio-temporal fusion neural network with CNN and LSTM & Facilitates learning of multiscale spatial features and correlation between SCADA features by utilising a multi-kernel fusion CNN; LSTM is used which further learns temporal dependencies in the data; Method outperforms multiple conventional ML baselines & \\
Kong et al. (2020) \cite{kong_tang_deng_liu_han_2020}  & CNN-GRU hybrid model & Performs fusion of spatio-temporal SCADA features; Model was trained with historical data on normal operation of turbines, and deviations based on residual values used to detect anomalies; Shows immense promise of the approach for serving as a monitoring indicator in CBM & Cannot provide transparent decisions with reasoning and rationale for predictions \\
Jiang et al. (2018) \cite{8059861} & DAE for unsupervised learning with sliding-window approach & Can help detect faults based on multiple SCADA features from sensors without historical ground truth for faults; Promising performance in situations of noise and fluctuating input; Helps capture non linear correlations between SCADA features and temporal dependencies for improved predictions & Generally cannot be validated without historical fault records; Cannot provide rationale for decisions \\
Kalyanraj et al. (2016) \cite{8052714} and Alhmoud and Al-Zoubi (2019) \cite{IoTApplicationsinWindEnergyConversionSystems} & IoT & Applications span turbine control, data logging of vital O\&M parameters such as wind speed, power etc. on cloud servers ; Can help facilitate remote access and control of turbines during decision support & Challenging to integrate with AI algorithms and deploy for real-time ; Security and privacy concerns are particularly a significant constraint\\
Chatterjee and Dethlefs (2020) \cite{windenergy_journal} & LSTM-XGBoost model for Explainable AI & Can provide transparent anomaly prediction in multiple turbine sub-components with SCADA data; Transfer learning is demonstrated for facilitating predictions in new domains lacking labelled training data & Large amounts of training data required during initial model training in source domain to achieve optimal results \\
Wang et al. (2019) \cite{9041585} & LSTM with attention for Explainable AI & Can provide interpretable wind power predictions; Generally outperforms conventional ML techniques significantly & More complex to model in comparison to simpler neural architectures like ANN \\
Kumar et al. (2020) \cite{en13071772}, Jianjun et al. (2019) \cite{en12142764} and Chatterjee and Dethlefs (2020) \cite{torque_paper} & CNN with attention for Explainable AI & Multiple applications spanning short-term prediction of wind speed, imbalance fault detection in blades, causal inference in SCADA features; Provides novel insights for transparent decisions &  More complex to model in comparison to simpler neural architectures like ANN \\
Sowdaboina et al. (2014) \cite{10.1007/978-3-642-54906-9_42} & Rule-based NLG & Can summarise vital time-series parameters like wind speed and direction etc. & Development is time-consuming and challenging \\
Dubey et al. (2018) \cite{DBLP:conf/flairs/DubeyCK18} & CBR for NLG & Provides textual summary of meteorological information which can be of relevance to wind industry; Demonstrate highly promising results on combining CBR with rule-based NLG methods & Can only provide very limited information \\
Chatterjee and Dethlefs (2020) \cite{wcci_paper} & Transformers for NLG & Facilitates generation of detailed human-intelligible messages for faults as well as maintenance strategies; Provides explainability with feature importance leading to predicted decisions & Significantly complex to model and develop in comparison to conventional rule-based and CBR techniques\\
Tomin et al. (2019) \cite{8771645} and Gauna et al. (2016) \cite{gauna_paper} & RL for intelligent control & Can help in intelligently controlling MIMO based turbine system controllers; Shows promising results compared to traditional control techniques in modelling multi-objective problems relevant to wind industry & \\
Aguirre et al. (2020) \cite{doi:10.1002/we.2451} & DRL for intelligent control & Demonstrates promising results in wind turbine yaw control; The learning capability of ANNs in DRLs helps significantly outperform conventional RL algorithms & DRL techniques can be more complex to model; Requires significant training time in comparison to vanilla RL algorithms \\
Chatterjee and Dethlefs (2020) \cite{renew_paper} & DRL for planning & Demonstrates immense promise for maintenance action planning in offshore vessel transfers; Considers SCADA data along with other vital parameters like weather conditions and types/severity of faults & More complex to model in comparison to vanilla RL techniques; Requires historically labelled failure data for O\&M planning \\
Qureshi et al. (2017) \cite{QURESHI2017742},  Hu et al. (2016) \cite{RePEc:eee:renene:v:85:y:2016:i:c:p:83-95}, Zhang et al. (2018) \cite{8409794}, Chatterjee and Dethlefs (2020) \cite{windenergy_journal}, Pan et al. (2020)  \cite{doi:10.1177/1475921720919073} & Transfer learning for domain knowledge transfer & Multiple applications spanning wind power prediction, short-term wind speed prediction, ice assessment on blades, fault prediction in turbine sub-components, monitoring vital O\&M parameters during anomalies etc. & Generally requires high quality historically labelled data for initial model training (in source domain); Can be complex to model requiring high computational power \\
Sainz et al. (2019) \cite{sainz_robust} & LMedS with random search for data filtering & Provides promising filtering ability from data such as wind speed and wind direction & Generally significantly outperformed by more sophisticated methods based on ML for data filtering; Cannot handle challenges posed by imbalanced datasets\\
Shen et al. (2019) \cite{8330024}  & Change point grouping and quartile algorithm for data filtering & Helps to improve data quality for modelling power curves; Demonstrates promising results in filtering outliers based on data distribution characteristics & Cannot handle challenges posed by imbalanced datasets \\
Yi et al. (2020) \cite{9302680}, Ge et al. (2017) \cite{8245530} and Chatterjee and Dethlefs (2020) \cite{windenergy_journal} & SMOTE for balancing datasets & Can generate synthetic data points to tackle imbalanced datasets in the wind industry; Helps overcome challenges posed by lack of abundantly labelled records (e.g. for anomalies) during AI model training & Techniques like RAMOBoost and ADASYN generally perform better than SMOTE in most cases in other domains e.g. healthcare (but have not been applied yet in the wind industry) \\ 
Chatterjee and Dethlefs (2020) \cite{renew_paper}  & SHAP for Explainable AI & Can provide transparent outputs for predictions made by black-box natured AI models; Demonstrates significant promise in O\&M planning for interpretable fault prediction in multiple turbine sub-components & Cannot be utilised out of the box for more complex AI architectures (especially deep learning models like RNNs and CNNs)  \\
Fu et al. (2019) \cite{8955569} and Chatterjee and Dethlefs (2020) \cite{wcci_paper} & Seq2Seq model with attention for Explainable AI & Applications spanning wind power forecasting and generation of alarm messages for faults; Can provide feature importances for predictions with added transparency & Generally outperformed significantly by more sophisticated neural architectures like transformers \\
Meng et al. (2019) \cite{en12244612} and Chatterjee and Dethlefs (2020) \cite{wcci_paper} & Transformers for Explainable AI &  Applications include short-term load forecasting  and prediction of alarm messages and maintenance actions during O\&M; Demonstrates that the model significantly outperforms conventional baselines such as Seq2Seq; Can help provide human-intelligible decisions during NLG & More complex to model and requires significantly greater training time compared to vanilla Seq2Seq architectures   \\
Zhang et al. (2018) \cite{8329419}, Chatterjee and Dethlefs (2020) \cite{windenergy_journal,renew_paper}, Wu et al. (2020) \cite{app10093258},  Yuan et al. (2019) \cite{RePEc:gam:jeners:v:12:y:2019:i:22:p:4224-:d:283963} and Browell et al. (2017) \cite{7981134} & XGBoost for Explainable AI & Multiple applications spanning detection of faults in multiple turbine sub-components, wind power forecasting etc.; Can provide feature importances for predictions made by AI models; Computationally efficient and scalable model in comparison to conventional ML techniques & Often outperformed in performance (e.g. accuracy) by more sophisticated architectures (especially deep learners)

\\ \hline
 \bottomrule
\end{longtable}
\end{small}



\printcredits

\bibliographystyle{elsarticle-num}

\bibliography{cas-refs}

\begin{thebibliography}{100}
\expandafter\ifx\csname url\endcsname\relax
  \def\url#1{\texttt{#1}}\fi
\expandafter\ifx\csname urlprefix\endcsname\relax\def\urlprefix{URL }\fi
\expandafter\ifx\csname href\endcsname\relax
  \def\href#1#2{#2} \def\path#1{#1}\fi

\bibitem{globalcap_review}
J.~Kaldellis, D.~Zafirakis, The wind energy (r)evolution: A short review of a
  long history, Renewable Energy 36 (2011) 1887--1901.
\newblock \href {http://dx.doi.org/10.1016/j.renene.2011.01.002}
  {\path{doi:10.1016/j.renene.2011.01.002}}.

\bibitem{STETCO2019620}
A.~Stetco, F.~Dinmohammadi, X.~Zhao, V.~Robu, D.~Flynn, M.~Barnes, J.~Keane,
  G.~Nenadic,
  \href{http://www.sciencedirect.com/science/article/pii/S096014811831231X}{Machine
  learning methods for wind turbine condition monitoring: A review}, Renewable
  Energy 133 (2019) 620--635.
\newblock \href
  {http://dx.doi.org/https://doi.org/10.1016/j.renene.2018.10.047}
  {\path{doi:https://doi.org/10.1016/j.renene.2018.10.047}}.
\newline\urlprefix\url{http://www.sciencedirect.com/science/article/pii/S096014811831231X}

\bibitem{windenergy_journal}
J.~Chatterjee, N.~Dethlefs,
  \href{https://onlinelibrary.wiley.com/doi/abs/10.1002/we.2510}{Deep learning
  with knowledge transfer for explainable anomaly prediction in wind turbines},
  Wind Energy 23~(8) (2020) 1693--1710.
\newblock \href
  {http://arxiv.org/abs/https://onlinelibrary.wiley.com/doi/pdf/10.1002/we.2510}
  {\path{arXiv:https://onlinelibrary.wiley.com/doi/pdf/10.1002/we.2510}}, \href
  {http://dx.doi.org/10.1002/we.2510} {\path{doi:10.1002/we.2510}}.
\newline\urlprefix\url{https://onlinelibrary.wiley.com/doi/abs/10.1002/we.2510}

\bibitem{Rockmann2017}
C.~R{\"o}ckmann, S.~Lagerveld, J.~Stavenuiter,
  \href{https://doi.org/10.1007/978-3-319-51159-7_4}{Operation and Maintenance
  Costs of Offshore Wind Farms and Potential Multi-use Platforms in the Dutch
  North Sea}, Springer International Publishing, Cham, 2017, Ch.~1, pp.
  97--113.
\newblock \href {http://dx.doi.org/10.1007/978-3-319-51159-7_4}
  {\path{doi:10.1007/978-3-319-51159-7_4}}.
\newline\urlprefix\url{https://doi.org/10.1007/978-3-319-51159-7_4}

\bibitem{en12020201}
K.~Leahy, C.~Gallagher, P.~O’Donovan, D.~T.~J. O’Sullivan,
  \href{https://www.mdpi.com/1996-1073/12/2/201}{Issues with data quality for
  wind turbine condition monitoring and reliability analyses}, Energies 12~(2).
\newblock \href {http://dx.doi.org/10.3390/en12020201}
  {\path{doi:10.3390/en12020201}}.
\newline\urlprefix\url{https://www.mdpi.com/1996-1073/12/2/201}

\bibitem{ibrahim_cbmchallenge}
I.~S. Kuseyri, Condition monitoring of wind turbines: Challenges and
  opportunities, 2015.

\bibitem{perf_energycost}
Y.~Charabi, S.~Abdul-wahab, Wind turbine performance analysis for energy cost
  minimization, Renewables: Wind, Water, and Solar 7.
\newblock \href {http://dx.doi.org/10.1186/s40807-020-00062-7}
  {\path{doi:10.1186/s40807-020-00062-7}}.

\bibitem{en12020225}
Y.~Merizalde, L.~Hernández-Callejo, O.~Duque-Perez, V.~Alonso-Gómez,
  \href{https://www.mdpi.com/1996-1073/12/2/225}{Maintenance models applied to
  wind turbines. a comprehensive overview}, Energies 12~(2).
\newblock \href {http://dx.doi.org/10.3390/en12020225}
  {\path{doi:10.3390/en12020225}}.
\newline\urlprefix\url{https://www.mdpi.com/1996-1073/12/2/225}

\bibitem{LIN2020117693}
Z.~Lin, X.~Liu,
  \href{http://www.sciencedirect.com/science/article/pii/S0360544220308008}{Wind
  power forecasting of an offshore wind turbine based on high-frequency scada
  data and deep learning neural network}, Energy 201 (2020) 117693.
\newblock \href
  {http://dx.doi.org/https://doi.org/10.1016/j.energy.2020.117693}
  {\path{doi:https://doi.org/10.1016/j.energy.2020.117693}}.
\newline\urlprefix\url{http://www.sciencedirect.com/science/article/pii/S0360544220308008}

\bibitem{recent_cbm}
J.~Maldonado, S.~Martin-Martinez, E.~Artigao, E.~Gomez-Lazaro, Using scada data
  for wind turbine condition monitoring: A systematic literature review,
  Energies 13 (2020) 3132.
\newblock \href {http://dx.doi.org/10.3390/en13123132}
  {\path{doi:10.3390/en13123132}}.

\bibitem{7947229}
Y.~{He}, A.~{Kusiak}, Performance assessment of wind turbines: Data-derived
  quantitative metrics, IEEE Transactions on Sustainable Energy 9~(1) (2018)
  65--73.
\newblock \href {http://dx.doi.org/10.1109/TSTE.2017.2715061}
  {\path{doi:10.1109/TSTE.2017.2715061}}.

\bibitem{yang_jiang_2011}
W.~Yang, J.~Jiang, Wind turbine condition monitoring and reliability analysis
  by scada information, 2011 Second International Conference on Mechanic
  Automation and Control Engineering\href
  {http://dx.doi.org/10.1109/mace.2011.5987329}
  {\path{doi:10.1109/mace.2011.5987329}}.

\bibitem{scada_windturbine_review}
J.~Tautz-Weinert, S.~J. Watson, Using scada data for wind turbine condition
  monitoring – a review, IET Renewable Power Generation 11 (2017) 382--394.

\bibitem{scada_planning}
D.~Peharda, I.~Ivankovic, N.~Jaman, Using data from scada for centralized
  transformer monitoring applications, Procedia Engineering 202 (2017) 65--75.
\newblock \href {http://dx.doi.org/10.1016/j.proeng.2017.09.695}
  {\path{doi:10.1016/j.proeng.2017.09.695}}.

\bibitem{exp_review}
C.~F. Tan, L.~Wahidin, S.~N. Khalil, N.~Tamaldin, J.~Hu, M.~Rauterberg, The
  application of expert system: A review of research and applications, ARPN
  Journal of Engineering and Applied Sciences 11 (2016) 2448--2453.

\bibitem{wang2018blessings}
Y.~Wang, D.~M. Blei, \href{https://doi.org/10.1080/01621459.2019.1686987}{{The
  Blessings of Multiple Causes}}, {Journal of the American Statistical
  Association} Volume 0 (2019) 1--71.
\newblock \href
  {http://arxiv.org/abs/https://doi.org/10.1080/01621459.2019.1686987}
  {\path{arXiv:https://doi.org/10.1080/01621459.2019.1686987}}, \href
  {http://dx.doi.org/10.1080/01621459.2019.1686987}
  {\path{doi:10.1080/01621459.2019.1686987}}.
\newline\urlprefix\url{https://doi.org/10.1080/01621459.2019.1686987}

\bibitem{windcontrol_ai}
Y.~Wang, Y.~Yu, S.~Cao, X.~Zhang, S.~Gao,
  \href{https://doi.org/10.1007/s10462-019-09768-7}{A review of applications of
  artificial intelligent algorithms in wind farms}, Artificial Intelligence
  Review 53~(5) (2020) 3447--3500.
\newblock \href {http://dx.doi.org/10.1007/s10462-019-09768-7}
  {\path{doi:10.1007/s10462-019-09768-7}}.
\newline\urlprefix\url{https://doi.org/10.1007/s10462-019-09768-7}

\bibitem{ann_surveywind}
A.~Pliego~Marugán, F.~P. García~Márquez, J.~M. Pinar~Pérez,
  D.~Ruiz-Hernández, A survey of artificial neural network in wind energy
  systems, Applied Energy 228 (2018) 1822--1836.
\newblock \href {http://dx.doi.org/10.1016/j.apenergy.2018.07.084}
  {\path{doi:10.1016/j.apenergy.2018.07.084}}.

\bibitem{doi:10.1177/0309524X19891672}
J.~Maldonado-Correa, J.~Solano, M.~Rojas-Moncayo,
  \href{https://doi.org/10.1177/0309524X19891672}{Wind power forecasting: A
  systematic literature review}, Wind Engineering 0~(0) (0) 0309524X19891672.
\newblock \href {http://arxiv.org/abs/https://doi.org/10.1177/0309524X19891672}
  {\path{arXiv:https://doi.org/10.1177/0309524X19891672}}, \href
  {http://dx.doi.org/10.1177/0309524X19891672}
  {\path{doi:10.1177/0309524X19891672}}.
\newline\urlprefix\url{https://doi.org/10.1177/0309524X19891672}

\bibitem{sun_sun_2018}
Z.~Sun, H.~Sun, Health status assessment for wind turbine with recurrent neural
  networks, Mathematical Problems in Engineering 2018 (2018) 1--16.
\newblock \href {http://dx.doi.org/10.1155/2018/6972481}
  {\path{doi:10.1155/2018/6972481}}.

\bibitem{torque_paper}
J.~Chatterjee, N.~Dethlefs, Temporal causal inference in wind turbine scada
  data using deep learning for explainable {AI}, Journal of Physics: Conf
  Series.

\bibitem{LEITE20181917}
G.~de~Novaes Pires~Leite, A.~M. Araújo, P.~A.~C. Rosas,
  \href{http://www.sciencedirect.com/science/article/pii/S1364032117309383}{Prognostic
  techniques applied to maintenance of wind turbines: a concise and specific
  review}, Renewable and Sustainable Energy Reviews 81 (2018) 1917 -- 1925.
\newblock \href {http://dx.doi.org/https://doi.org/10.1016/j.rser.2017.06.002}
  {\path{doi:https://doi.org/10.1016/j.rser.2017.06.002}}.
\newline\urlprefix\url{http://www.sciencedirect.com/science/article/pii/S1364032117309383}

\bibitem{wcci_paper}
J.~Chatterjee, N.~Dethlefs, A dual transformer model for intelligent decision
  support for maintenance of wind turbines, in: International Joint Conference
  on Neural Networks (IJCNN), Glasgow (UK), 2020, pp. 1--10.

\bibitem{bib_med}
P.~Kokol, H.~Blazun~Vosner, J.~Zavrsnik, Application of bibliometrics in
  medicine: a historical bibliometrics analysis, Health Information \&
  Libraries Journal\href {http://dx.doi.org/10.1111/hir.12295}
  {\path{doi:10.1111/hir.12295}}.

\bibitem{bib_finance}
J.~M. Merigo, J.-B. Yang, Bibliometric analysis in financial research, in: IEEE
  Conference on Computational Intelligence for Financial Engineering \&
  Economics (CIFEr), 2014, pp. 223--230.
\newblock \href {http://dx.doi.org/10.1109/CIFEr.2014.6924077}
  {\path{doi:10.1109/CIFEr.2014.6924077}}.

\bibitem{ramasamy2012}
P.~Kanagavel, G.~S, S.~S, R.~RU, A scientometric assessment of wind energy
  research productivity: A scientometric study, International Journal of
  Scientific Research 2~(5) (2012) 333–336.
\newblock \href {http://dx.doi.org/10.15373/22778179/may2013/113}
  {\path{doi:10.15373/22778179/may2013/113}}.

\bibitem{doi:10.1080/1331677X.2020.1734853}
P.~Ye, Y.~Li, H.~Zhang, H.~Shen,
  \href{https://doi.org/10.1080/1331677X.2020.1734853}{Bibliometric analysis on
  the research of offshore wind power based on web of science}, Economic
  Research-Ekonomska Istraživanja 33~(1) (2020) 887--903.
\newblock \href
  {http://arxiv.org/abs/https://doi.org/10.1080/1331677X.2020.1734853}
  {\path{arXiv:https://doi.org/10.1080/1331677X.2020.1734853}}, \href
  {http://dx.doi.org/10.1080/1331677X.2020.1734853}
  {\path{doi:10.1080/1331677X.2020.1734853}}.
\newline\urlprefix\url{https://doi.org/10.1080/1331677X.2020.1734853}

\bibitem{mohanathan}
P.~Mohanathan, N.~Rajendran, Mapping of wind energy research output: A
  scientometric analysis, Eco-Chronicle 13~(4) (2018) 239--245.

\bibitem{ARIA2017959}
M.~Aria, C.~Cuccurullo,
  \href{http://www.sciencedirect.com/science/article/pii/S1751157717300500}{bibliometrix:
  An r-tool for comprehensive science mapping analysis}, Journal of
  Informetrics 11~(4) (2017) 959--975.
\newblock \href {http://dx.doi.org/https://doi.org/10.1016/j.joi.2017.08.007}
  {\path{doi:https://doi.org/10.1016/j.joi.2017.08.007}}.
\newline\urlprefix\url{http://www.sciencedirect.com/science/article/pii/S1751157717300500}

\bibitem{doi:10.1002/asi.20317}
C.~Chen,
  \href{https://onlinelibrary.wiley.com/doi/abs/10.1002/asi.20317}{Citespace
  ii: Detecting and visualizing emerging trends and transient patterns in
  scientific literature}, Journal of the American Society for Information
  Science and Technology 57~(3) (2006) 359--377.
\newblock \href
  {http://arxiv.org/abs/https://onlinelibrary.wiley.com/doi/pdf/10.1002/asi.20317}
  {\path{arXiv:https://onlinelibrary.wiley.com/doi/pdf/10.1002/asi.20317}},
  \href {http://dx.doi.org/10.1002/asi.20317} {\path{doi:10.1002/asi.20317}}.
\newline\urlprefix\url{https://onlinelibrary.wiley.com/doi/abs/10.1002/asi.20317}

\bibitem{eck_waltman_2009}
N.~J.~V. Eck, L.~Waltman, Software survey: Vosviewer, a computer program for
  bibliometric mapping, Scientometrics 84~(2) (2009) 523–538.
\newblock \href {http://dx.doi.org/10.1007/s11192-009-0146-3}
  {\path{doi:10.1007/s11192-009-0146-3}}.

\bibitem{lorenz_aisch_kokkelink_2012}
M.~Lorenz, G.~Aisch, D.~Kokkelink,
  \href{https://github.com/datawrapper/datawrapper}{Create charts and maps
  [software]} (2012).
\newline\urlprefix\url{https://github.com/datawrapper/datawrapper}

\bibitem{bloomberg_volpe_2016}
L.~D. Bloomberg, M.~Volpe, Completing your qualitative dissertation a roadmap
  from beginning to end, SAGE, 2016.

\bibitem{keywordplus_paper}
J.~Zhang, Q.~Yu, F.~Zheng, C.~Long, Z.~Lu, Z.~Duan, Comparing keywords plus of
  wos and author keywords: A case study of patient adherence research, Journal
  of the Association for Information Science and Technology 67.
\newblock \href {http://dx.doi.org/10.1002/asi.23437}
  {\path{doi:10.1002/asi.23437}}.

\bibitem{RePEc:eee:renene:v:47:y:2012:i:c:p:112-126}
Z.~Feng, M.~Liang, Y.~Zhang, S.~Hou,
  \href{https://ideas.repec.org/a/eee/renene/v47y2012icp112-126.html}{{Fault
  diagnosis for wind turbine planetary gearboxes via demodulation analysis
  based on ensemble empirical mode decomposition and energy separation}},
  Renewable Energy 47~(C) (2012) 112--126.
\newblock \href {http://dx.doi.org/10.1016/j.renene.2012.04.}
  {\path{doi:10.1016/j.renene.2012.04.}}
\newline\urlprefix\url{https://ideas.repec.org/a/eee/renene/v47y2012icp112-126.html}

\bibitem{LIU2013954}
W.~Liu,
  \href{http://www.sciencedirect.com/science/article/pii/S0141029613002915}{The
  vibration analysis of wind turbine blade–cabin–tower coupling system},
  Engineering Structures 56 (2013) 954--957.
\newblock \href
  {http://dx.doi.org/https://doi.org/10.1016/j.engstruct.2013.06.008}
  {\path{doi:https://doi.org/10.1016/j.engstruct.2013.06.008}}.
\newline\urlprefix\url{http://www.sciencedirect.com/science/article/pii/S0141029613002915}

\bibitem{10.1007/978-3-642-28768-8_52}
R.~Zimroz, W.~Bartelmus, T.~Barszcz, J.~Urbanek, Statistical data processing
  for wind turbine generator bearing diagnostics, in: T.~Fakhfakh,
  W.~Bartelmus, F.~Chaari, R.~Zimroz, M.~Haddar (Eds.), Condition Monitoring of
  Machinery in Non-Stationary Operations, Springer Berlin Heidelberg, Berlin,
  Heidelberg, 2012, pp. 509--518.

\bibitem{5673197}
Y.~{Li}, A discussion on using empirical mode decomposition for incipient fault
  detection and diagnosis of the wind turbine gearbox, in: 2010 World
  Non-Grid-Connected Wind Power and Energy Conference, 2010, pp. 1--5.

\bibitem{2011JSV...330.3766Y}
W.~{Yang}, R.~{Court}, P.~J. {Tavner}, C.~J. {Crabtree}, {Bivariate empirical
  mode decomposition and its contribution to wind turbine condition
  monitoring}, Journal of Sound Vibration 330~(15) (2011) 3766--3782.
\newblock \href {http://dx.doi.org/10.1016/j.jsv.2011.02.027}
  {\path{doi:10.1016/j.jsv.2011.02.027}}.

\bibitem{kalman_pf}
Z.-W. Zheng, Y.-Y. Chen, X.-W. Zhou, M.-M. Huo, B.~Zhao, M.-Y. Guo, Short-term
  wind power forecasting using empirical mode decomposition and rbfnn,
  International Journal of Smart Grid and Clean Energy 2 (2013) 192--199.
\newblock \href {http://dx.doi.org/10.12720/sgce.2.2.192-199}
  {\path{doi:10.12720/sgce.2.2.192-199}}.

\bibitem{7505124}
L.~{Saidi}, E.~{Bechhoefer}, J.~{Ben Ali}, M.~{Benbouzid}, Wind turbine
  high-speed shaft bearing degradation analysis for run-to-failure testing
  using spectral kurtosis, in: 2015 16th International Conference on Sciences
  and Techniques of Automatic Control and Computer Engineering (STA), 2015, pp.
  267--272.

\bibitem{jia_lei_shan_lin_2015}
F.~Jia, Y.~Lei, H.~Shan, J.~Lin, Early fault diagnosis of bearings using an
  improved spectral kurtosis by maximum correlated kurtosis deconvolution,
  Sensors 15~(11) (2015) 29363–29377.
\newblock \href {http://dx.doi.org/10.3390/s151129363}
  {\path{doi:10.3390/s151129363}}.

\bibitem{5554606}
{Yanping Guo}, {Wenjun Yan}, {Zhejing Bao}, Gear fault diagnosis of wind
  turbine based on discrete wavelet transform, in: 2010 8th World Congress on
  Intelligent Control and Automation, 2010, pp. 5804--5808.

\bibitem{doi:10.1155/2013/212836}
Q.~Yang, D.~An, \href{https://doi.org/10.1155/2013/212836}{Emd and wavelet
  transform based fault diagnosis for wind turbine gear box}, Advances in
  Mechanical Engineering 5 (2013) 212836.
\newblock \href {http://arxiv.org/abs/https://doi.org/10.1155/2013/212836}
  {\path{arXiv:https://doi.org/10.1155/2013/212836}}, \href
  {http://dx.doi.org/10.1155/2013/212836} {\path{doi:10.1155/2013/212836}}.
\newline\urlprefix\url{https://doi.org/10.1155/2013/212836}

\bibitem{wavelet_dis}
A.~Sovic, D.~Sersic, Signal decomposition methods for reducing drawbacks of the
  dwt, Engineering Review 32~(2) (2012) 70--77.

\bibitem{Clifton_2013}
A.~Clifton, L.~Kilcher, J.~K. Lundquist, P.~Fleming,
  \href{https://doi.org/10.1088%2F1748-9326%2F8%2F2%2F024009}{Using machine
  learning to predict wind turbine power output}, Environmental Research
  Letters 8~(2) (2013) 024009.
\newblock \href {http://dx.doi.org/10.1088/1748-9326/8/2/024009}
  {\path{doi:10.1088/1748-9326/8/2/024009}}.
\newline\urlprefix\url{https://doi.org/10.1088%2F1748-9326%2F8%2F2%2F024009}

\bibitem{Pravilovic_2014}
S.~Pravilovic, A.~Appice, A.~Lanza, D.~Malerba,
  \href{https://doi.org/10.1007%2F978-3-319-11812-3_24}{Wind power forecasting
  using time series cluster analysis}, in: Discovery Science, Springer
  International Publishing, 2014, pp. 276--287.
\newblock \href {http://dx.doi.org/10.1007/978-3-319-11812-3_24}
  {\path{doi:10.1007/978-3-319-11812-3_24}}.
\newline\urlprefix\url{https://doi.org/10.1007%2F978-3-319-11812-3_24}

\bibitem{Yang_2015}
L.~Yang, M.~He, J.~Zhang, V.~Vittal,
  \href{https://doi.org/10.1109%2Ftste.2015.2406814}{Support-vector-machine-enhanced
  markov model for short-term wind power forecast}, {IEEE} Transactions on
  Sustainable Energy 6~(3) (2015) 791--799.
\newblock \href {http://dx.doi.org/10.1109/tste.2015.2406814}
  {\path{doi:10.1109/tste.2015.2406814}}.
\newline\urlprefix\url{https://doi.org/10.1109%2Ftste.2015.2406814}

\bibitem{Chen_2014}
N.~Chen, Z.~Qian, I.~T. Nabney, X.~Meng,
  \href{https://doi.org/10.1109%2Ftpwrs.2013.2282366}{Wind power forecasts
  using gaussian processes and numerical weather prediction}, {IEEE}
  Transactions on Power Systems 29~(2) (2014) 656--665.
\newblock \href {http://dx.doi.org/10.1109/tpwrs.2013.2282366}
  {\path{doi:10.1109/tpwrs.2013.2282366}}.
\newline\urlprefix\url{https://doi.org/10.1109%2Ftpwrs.2013.2282366}

\bibitem{Soleymani_Mohammadi_Rezayi_Moghimai_2015}
S.~Soleymani, S.~Mohammadi, H.-R. Rezayi, R.~Moghimai,
  \href{http://doi.org/10.3233/IFS-141433}{A new hybrid method to forecast wind
  turbine output power in power systems}, Journal of Intelligent \& Fuzzy
  Systems 28~(4) (2015) 1503–1508.
\newblock \href {http://dx.doi.org/10.3233/IFS-141433}
  {\path{doi:10.3233/IFS-141433}}.
\newline\urlprefix\url{http://doi.org/10.3233/IFS-141433}

\bibitem{windtur_cbm}
W.~Yang, P.~J. Tavner, C.~J. Crabtree, Y.~Feng, Y.~Qiu, Wind turbine condition
  monitoring: technical and commercial challenges, Wind Energy 17 (2014)
  673--693.

\bibitem{paper4}
A.~Clifton, L.~Kilcher, J.~Lundquist, P.~Fleming, Using machine learning to
  predict wind turbine power output, in: Environmental Research Letters, Volume
  8, Number 2, IOP Publishing Ltd., 2013, pp. 1--8.

\bibitem{wind_mountainpass}
A.~Clifton, M.~Daniels, M.~Lehning, Effect of winds in a mountain pass on
  turbine performance, Wind Energy 17 (2014) 1543--1562.

\bibitem{paper2}
P.~Zhao, J.~Xia, Y.~Dai, J.~He, Wind speed prediction using support vector
  regression, in: IEEE Conference on Industrial Electronics and Applications,
  IEEE, Taichung, Taiwan, 2010, pp. 882--886.

\bibitem{realtime_reconstruction}
C.~Yang, J.~Liu, Y.~Zeng, G.~Xie, Real-time condition monitoring and fault
  detection of components based on machine-learning reconstruction model,
  Renewable Energy 133 (2019) 433--441.

\bibitem{multivariate_windfield}
J.~Park, K.~Smarsly, K.~H. Law, , D.~Hartmann, Multivariate analysis and
  prediction of wind turbine response to varying wind field characteristics
  based on machine learning, in: ASCE International Workshop on Computing in
  Civil Engineering, Los Angeles, California, 2013, pp. 113--120.

\bibitem{paper5}
M.~Du, S.~Ma, Q.~He, A scada data based anomaly detection method for wind
  turbines, in: China International Conference on Electricity Distribution,
  IEEE, Xi'an, China, 2016, pp. 1--6.

\bibitem{outlook_windeurope}
A.~Nghiem, D.~Fraile, A.~Mbistrova, T.~Remy, Wind energy in europe: Outlook to
  2020, WindEurope.

\bibitem{paper3}
M.~Morshedizadeh, Condition monitoring of wind turbines using intelligent
  machine learning techniques, in: Electronic Theses and Dissertations. 6002,
  https://scholar.uwindsor.ca/etd/6002, University of Windsor, University of
  Windsor, Ontario, Canada, 2017, pp. 1--96.

\bibitem{QURESHI2017742}
A.~S. Qureshi, A.~Khan, A.~Zameer, A.~Usman,
  \href{http://www.sciencedirect.com/science/article/pii/S1568494617302946}{Wind
  power prediction using deep neural network based meta regression and transfer
  learning}, Applied Soft Computing 58 (2017) 742--755.
\newblock \href {http://dx.doi.org/https://doi.org/10.1016/j.asoc.2017.05.031}
  {\path{doi:https://doi.org/10.1016/j.asoc.2017.05.031}}.
\newline\urlprefix\url{http://www.sciencedirect.com/science/article/pii/S1568494617302946}

\bibitem{hochreiter_1998}
S.~Hochreiter, The vanishing gradient problem during learning recurrent neural
  nets and problem solutions, International Journal of Uncertainty, Fuzziness
  and Knowledge-Based Systems 06~(02) (1998) 107--116.
\newblock \href {http://dx.doi.org/10.1142/s0218488598000094}
  {\path{doi:10.1142/s0218488598000094}}.

\bibitem{hochreiter_schmidhuber_1997}
S.~Hochreiter, J.~Schmidhuber, Long short-term memory, Neural Computation 9~(8)
  (1997) 1735--1780.
\newblock \href {http://dx.doi.org/10.1162/neco.1997.9.8.1735}
  {\path{doi:10.1162/neco.1997.9.8.1735}}.

\bibitem{ann_paper}
P.~Zhang, E.~Patuwo, M.~Hu, Forecasting with artificial neural networks: The
  state of the art, International Journal of Forecasting 14 (1998) 35--62.
\newblock \href {http://dx.doi.org/10.1016/S0169-2070(97)00044-7}
  {\path{doi:10.1016/S0169-2070(97)00044-7}}.

\bibitem{doi:10.1177/0954406218797972}
P.~A. Kulkarni, A.~S. Dhoble, P.~M. Padole,
  \href{https://doi.org/10.1177/0954406218797972}{Deep neural network-based
  wind speed forecasting and fatigue analysis of a large composite wind turbine
  blade}, Proceedings of the Institution of Mechanical Engineers, Part C:
  Journal of Mechanical Engineering Science 233~(8) (2019) 2794--2812.
\newblock \href {http://arxiv.org/abs/https://doi.org/10.1177/0954406218797972}
  {\path{arXiv:https://doi.org/10.1177/0954406218797972}}, \href
  {http://dx.doi.org/10.1177/0954406218797972}
  {\path{doi:10.1177/0954406218797972}}.
\newline\urlprefix\url{https://doi.org/10.1177/0954406218797972}

\bibitem{st_windpower}
Q.~Zhu, H.~Li, Z.~Wang, J.~Chen, B.~Wang, Short-term wind power forecasting
  based on lstm, Dianwang Jishu/Power System Technology 41 (2017) 3797--3802.
\newblock \href {http://dx.doi.org/10.13335/j.1000-3673.pst.2017.1657}
  {\path{doi:10.13335/j.1000-3673.pst.2017.1657}}.

\bibitem{liu_guan_hou_han_liu_sun_zheng_2019}
Y.~Liu, L.~Guan, C.~Hou, H.~Han, Z.~Liu, Y.~Sun, M.~Zheng, Wind power
  short-term prediction based on lstm and discrete wavelet transform, Applied
  Sciences 9~(6) (2019) 1108.
\newblock \href {http://dx.doi.org/10.3390/app9061108}
  {\path{doi:10.3390/app9061108}}.

\bibitem{log_reg}
J.~Peng, K.~Lee, G.~Ingersoll, An introduction to logistic regression analysis
  and reporting, Journal of Educational Research - J EDUC RES 96 (2002) 3--14.
\newblock \href {http://dx.doi.org/10.1080/00220670209598786}
  {\path{doi:10.1080/00220670209598786}}.

\bibitem{paper1}
K.~Leahy, R.~L. Hu, I.~C. Konstantakopoulos, C.~J. Spanos, A.~M. Agogino,
  Diagnosing wind turbine faults using machine learning techniques applied to
  operational data, in: International Conference on Prognostics and Health
  Management (ICPHM), IEEE, Ottawa, ON, Canada, 2016, pp. 1--8.

\bibitem{data_randomforest}
Y.~Si, L.~Qian, B.~Mao, D.~Zhang, A data-driven approach for fault detection of
  offshore wind turbines using random forests, in: IECON 2017 - 43rd Annual
  Conference of the IEEE Industrial Electronics Society, Beijing, China, 2017,
  pp. 3149--3154.

\bibitem{7998308}
M.~{Canizo}, E.~{Onieva}, A.~{Conde}, S.~{Charramendieta}, S.~{Trujillo},
  Real-time predictive maintenance for wind turbines using big data frameworks,
  in: 2017 IEEE International Conference on Prognostics and Health Management
  (ICPHM), 2017, pp. 70--77.
\newblock \href {http://dx.doi.org/10.1109/ICPHM.2017.7998308}
  {\path{doi:10.1109/ICPHM.2017.7998308}}.

\bibitem{faultdiag_decisiontree}
I.~Abdallah, V.~Dertimanis, H.~Mylonas, K.~Tatsis, E.~Chatzi, N.~Dervilis,
  K.~Worden, E.~Maguire, Fault diagnosis of wind turbine structures using
  decision tree learning algorithms with big data, in: Safety and Reliability
  – Safe Societies in a Changing World, Proceedings of the European Safety
  and Reliability Conference, Trondheim, Norway, 2018, pp. 3053--3061.

\bibitem{abdallah_autodecision}
I.~Abdallah, V.~Dertimanis, E.~Chatzi, An autonomous real-time decision tree
  framework for monitoring \& diagnostics on wind turbines, in: 2nd
  International Conference on Wind Energy Harvesting (WINERCOST), Catanzaro,
  Italy, 2018, pp. 149--152.

\bibitem{neural_nonlinear}
J.~Suykens, J.~Vandewalle, B.~De~Moor, Artificial Neural Networks for Modelling
  and Control of Non-Linear Systems, Springer, 1996.
\newblock \href {http://dx.doi.org/10.1007/978-1-4757-2493-6}
  {\path{doi:10.1007/978-1-4757-2493-6}}.

\bibitem{goodfellow_bengio_courville_2017}
I.~Goodfellow, Y.~Bengio, A.~Courville, Deep learning, The MIT Press,
  Cambridge, Mass.

\bibitem{opp_cbm}
Y.~Lu, L.~Sun, X.~Zhang, F.~Fenga, J.~Kang, G.~Fu, Condition based maintenance
  optimization for offshore wind turbine considering opportunities based on
  neural network approach, Applied Ocean Research 74 (2018) 69--79.

\bibitem{cbm_elm}
P.~Qian, X.~Ma, Y.~Wang, Condition monitoring of wind turbines based on extreme
  learning machine, in: Proceedings of the 21st International Conference on
  Automation and Computing, Glasgow, UK, 2015, pp. 1--6.

\bibitem{image_turblade}
Y.~Yu, H.~Cao, S.~Liu, S.~Yang, R.~Bai, Image-based damage recognition of wind
  turbine blades, in: 2nd International Conference on Advanced Robotics and
  Mechatronics (ICARM), Hefei and Tai'an, China, 2017, pp. 161--166.

\bibitem{image_turblade1}
H.~Li, W.~Zhou, J.~Xu, Structural health monitoring of wind turbine blades,
  Wind Turbine Control and Monitoring-Part of the Advances in Industrial
  Control book series (AIC) (2014) 231--265.

\bibitem{image_turblade2}
D.~Li, S.~Ho, G.~Song, L.~Ren, H.~Li, A review of damage detection methods for
  wind turbine blades, Smart Materials and Structures 24 (2015) 1--24.

\bibitem{vision_turblade}
S.~Moreno, M.~Pena, A.~Toledo, R.~Trevino, H.~Ponce, A new vision-based method
  using deep learning for damage inspection in wind turbine blades, in: 15th
  International Conference on Electrical Engineering, Computing Science and
  Automatic Control (CCE), Mexico City, Mexico, 2018, pp. 1--5.

\bibitem{windturlayout_1}
Y.-K. Wu, C.-Y. Lee, C.-R. Chen, K.-W. Hsu, H.-T. Tseng, Optimization of the
  wind turbine layout and transmission system planning for a large-scale
  offshore wind farm by ai technology, IEEE Transactions on Industry
  Applications 50 (2014) 2071--2080.

\bibitem{windturlayout_2}
Z.~Menghua, C.~Zhe, B.~Freede, Generation ratio availability assessment of
  electrical systems for offshore wind farms, IEEE Transactions on Energy
  Conversion 22 (2007) 755--763.

\bibitem{windturlayout_3}
S.~Dutta, T.~J. Overbye, A clustering based wind farm collector system cable
  layout design, in: IEEE Power and Energy Conference at Illinois, Champaign,
  IL, USA, 2011, pp. 1--6.

\bibitem{online_faultdetection}
A.~Zaher, S.~McArthur, D.~Infield, Online wind turbine fault detection through
  automated scada data analysis, Wind Energy 12 (2009) 574--593.

\bibitem{scalable_faultdetection}
M.~Bach-Andersen, B.~Romer-Odgaard, O.~Winther, Scalable systems for early
  fault detection in wind turbines : A data driven approach, in: European Wind
  Energy Association Annual Conference and Exhibition, Paris, France, 2015, pp.
  382--390.

\bibitem{neuralnet_csa}
R.~K. Ibrahim, J.~Tautz-Weinert, S.~J. Watson, Neural networks for wind turbine
  fault detection via current signature analysis, in: WindEurope Summit,
  Hamburg, Germany, 2016, pp. 1--7.

\bibitem{pang_he_jiang_xie_2020}
Y.~Pang, Q.~He, G.~Jiang, P.~Xie, Spatio-temporal fusion neural network for
  multi-class fault diagnosis of wind turbines based on scada data, Renewable
  Energy 161 (2020) 510–524.
\newblock \href {http://dx.doi.org/10.1016/j.renene.2020.06.154}
  {\path{doi:10.1016/j.renene.2020.06.154}}.

\bibitem{kong_tang_deng_liu_han_2020}
Z.~Kong, B.~Tang, L.~Deng, W.~Liu, Y.~Han, Condition monitoring of wind
  turbines based on spatio-temporal fusion of scada data by convolutional
  neural networks and gated recurrent units, Renewable Energy 146 (2020)
  760–768.
\newblock \href {http://dx.doi.org/10.1016/j.renene.2019.07.033}
  {\path{doi:10.1016/j.renene.2019.07.033}}.

\bibitem{8059861}
G.~{Jiang}, P.~{Xie}, H.~{He}, J.~{Yan}, Wind turbine fault detection using a
  denoising autoencoder with temporal information, IEEE/ASME Transactions on
  Mechatronics 23~(1) (2018) 89--100.
\newblock \href {http://dx.doi.org/10.1109/TMECH.2017.2759301}
  {\path{doi:10.1109/TMECH.2017.2759301}}.

\bibitem{9041585}
X.~{Wang}, Z.~{Li}, J.~{Zhang}, H.~{Liu}, C.~{Qiu}, X.~{Cai}, An lstm-attention
  wind power prediction method considering multiple factors, in: 8th Renewable
  Power Generation Conference (RPG 2019), 2019, pp. 1--7.

\bibitem{luong_pham_manning_2015}
T.~Luong, H.~Pham, C.~D. Manning,
  \href{https://www.aclweb.org/anthology/D15-1166}{Effective approaches to
  attention-based neural machine translation}, Proceedings of the 2015
  Conference on Empirical Methods in Natural Language Processing (2015)
  1412--1421\href {http://dx.doi.org/10.18653/v1/d15-1166}
  {\path{doi:10.18653/v1/d15-1166}}.
\newline\urlprefix\url{https://www.aclweb.org/anthology/D15-1166}

\bibitem{en13071772}
K.~Shivam, J.-C. Tzou, S.-C. Wu,
  \href{https://www.mdpi.com/1996-1073/13/7/1772}{Multi-step short-term wind
  speed prediction using a residual dilated causal convolutional network with
  nonlinear attention}, Energies 13~(7).
\newblock \href {http://dx.doi.org/10.3390/en13071772}
  {\path{doi:10.3390/en13071772}}.
\newline\urlprefix\url{https://www.mdpi.com/1996-1073/13/7/1772}

\bibitem{en12142764}
J.~Chen, W.~Hu, D.~Cao, B.~Zhang, Q.~Huang, Z.~Chen, F.~Blaabjerg,
  \href{https://www.mdpi.com/1996-1073/12/14/2764}{An imbalance fault detection
  algorithm for variable-speed wind turbines: A deep learning approach},
  Energies 12~(14).
\newblock \href {http://dx.doi.org/10.3390/en12142764}
  {\path{doi:10.3390/en12142764}}.
\newline\urlprefix\url{https://www.mdpi.com/1996-1073/12/14/2764}

\bibitem{renew_paper}
J.~Chatterjee, N.~Dethlefs, Deep reinforcement learning for maintenance
  planning of offshore vessel transfer, in: Proceedings of the 4th
  International Conference on Renewable Energies Offshore (RENEW 2020), Lisbon,
  Portugal, 2020, pp. 435--443.

\bibitem{gong-etal-2019-enhanced}
L.~Gong, J.~Crego, J.~Senellart, Enhanced transformer model for data-to-text
  generation, in: Proceedings of the 3rd Workshop on Neural Generation and
  Translation, Association for Computational Linguistics, Hong Kong, 2019, pp.
  148--156.

\bibitem{webnlg-automatic2017}
C.~Gardent, A.~Shimorina, S.~Narayana, L.~Perez{-}Beltrachini, {The WebNLG
  Challenge: Generating Text from RDF Data}, in: Proceedings of the
  International Natural Language Generation Conference (INLG), Santiago de
  Compostella, Spain, 2017, p. 124–133.

\bibitem{Garoufi:2010}
K.~Garoufi, A.~Koller, Automated planning for situated natural language
  generation, in: Proceedings of the 48th Annual Meeting of the Association for
  Computational Linguistics, ACL '10, 2010, pp. 1573--1582.

\bibitem{JuraskaNAACL2018}
J.~Juraska, P.~Karagiannis, K.~Bowden, M.~Walker,
  \href{http://aclweb.org/anthology/N18-1014}{A deep ensemble model with slot
  alignment for sequence-to-sequence natural language generation}, in:
  Proceedings of the 2018 Conference of the North American Chapter of the
  Association for Computational Linguistics: Human Language Technologies,
  Volume 1 (Long Papers), Association for Computational Linguistics, 2018, pp.
  152--162.
\newline\urlprefix\url{http://aclweb.org/anthology/N18-1014}

\bibitem{chen-etal-2020-shot}
Z.~Chen, H.~Eavani, W.~Chen, Y.~Liu, W.~Y. Wang,
  \href{https://www.aclweb.org/anthology/2020.acl-main.18}{Few-shot {NLG} with
  pre-trained language model}, in: Proceedings of the 58th Annual Meeting of
  the Association for Computational Linguistics, Association for Computational
  Linguistics, Online, 2020, pp. 183--190.
\newblock \href {http://dx.doi.org/10.18653/v1/2020.acl-main.18}
  {\path{doi:10.18653/v1/2020.acl-main.18}}.
\newline\urlprefix\url{https://www.aclweb.org/anthology/2020.acl-main.18}

\bibitem{10.1007/978-3-642-54906-9_42}
P.~K.~V. Sowdaboina, S.~Chakraborti, S.~Sripada, Learning to summarize time
  series data, in: A.~Gelbukh (Ed.), Computational Linguistics and Intelligent
  Text Processing, Springer Berlin Heidelberg, Berlin, Heidelberg, 2014, pp.
  515--528.

\bibitem{DBLP:conf/flairs/DubeyCK18}
N.~Dubey, S.~Chakraborti, D.~Khemani,
  \href{https://aaai.org/ocs/index.php/FLAIRS/FLAIRS18/paper/view/17673}{Content
  selection for time series summarization using case-based reasoning}, in:
  K.~Brawner, V.~Rus (Eds.), Proceedings of the Thirty-First International
  Florida Artificial Intelligence Research Society Conference, {FLAIRS} 2018,
  Melbourne, Florida, {USA.} May 21-23 2018, {AAAI} Press, 2018, pp. 395--398.
\newline\urlprefix\url{https://aaai.org/ocs/index.php/FLAIRS/FLAIRS18/paper/view/17673}

\bibitem{NIPS2017_7181}
A.~Vaswani, N.~Shazeer, N.~Parmar, J.~Uszkoreit, L.~Jones, A.~N. Gomez, L.~u.
  Kaiser, I.~Polosukhin,
  \href{http://papers.nips.cc/paper/7181-attention-is-all-you-need.pdf}{Attention
  is all you need}, in: I.~Guyon, U.~V. Luxburg, S.~Bengio, H.~Wallach,
  R.~Fergus, S.~Vishwanathan, R.~Garnett (Eds.), Advances in Neural Information
  Processing Systems 30, Curran Associates, Inc., 2017, pp. 5998--6008.
\newline\urlprefix\url{http://papers.nips.cc/paper/7181-attention-is-all-you-need.pdf}

\bibitem{vig2019transformervis}
J.~Vig, \href{https://arxiv.org/abs/1906.05714}{A multiscale visualization of
  attention in the transformer model}, arXiv preprint arXiv:1906.05714.
\newline\urlprefix\url{https://arxiv.org/abs/1906.05714}

\bibitem{devlin-etal-2019-bert}
J.~Devlin, M.-W. Chang, K.~Lee, K.~Toutanova,
  \href{https://www.aclweb.org/anthology/N19-1423}{{BERT}: Pre-training of deep
  bidirectional transformers for language understanding}, in: Proceedings of
  the 2019 Conference of the North {A}merican Chapter of the Association for
  Computational Linguistics: Human Language Technologies, Volume 1 (Long and
  Short Papers), Association for Computational Linguistics, Minneapolis,
  Minnesota, 2019, pp. 4171--4186.
\newline\urlprefix\url{https://www.aclweb.org/anthology/N19-1423}

\bibitem{robbins1951}
H.~Robbins, S.~Monro, \href{https://doi.org/10.1214/aoms/1177729586}{A
  stochastic approximation method}, Ann. Math. Statist. 22~(3) (1951) 400--407.
\newblock \href {http://dx.doi.org/10.1214/aoms/1177729586}
  {\path{doi:10.1214/aoms/1177729586}}.
\newline\urlprefix\url{https://doi.org/10.1214/aoms/1177729586}

\bibitem{drl_paper}
S.~Mousavi, M.~Schukat, E.~Howley, Deep reinforcement learning: An overview,
  in: Proceedings of SAI Intelligent Systems Conference, 2018, pp. 426--440.
\newblock \href {http://dx.doi.org/10.1007/978-3-319-56991-8_32}
  {\path{doi:10.1007/978-3-319-56991-8_32}}.

\bibitem{8771645}
N.~{TOMIN}, V.~{KURBATSKY}, H.~{GULIYEV}, Intelligent control of a wind turbine
  based on reinforcement learning, in: 2019 16th Conference on Electrical
  Machines, Drives and Power Systems (ELMA), 2019, pp. 1--6.

\bibitem{gauna_paper}
B.~Fernandez-Gauna, U.~Fernandez-Gamiz, M.~Graña, Variable speed wind turbine
  controller adaptation by reinforcement learning, Integrated Computer-Aided
  Engineering 24 (2016) 1--13.
\newblock \href {http://dx.doi.org/10.3233/ICA-160531}
  {\path{doi:10.3233/ICA-160531}}.

\bibitem{doi:10.1002/we.2451}
A.~Saenz-Aguirre, E.~Zulueta, U.~Fernandez-Gamiz, A.~Ulazia, D.~Teso-Fz-Betono,
  \href{https://onlinelibrary.wiley.com/doi/abs/10.1002/we.2451}{Performance
  enhancement of the artificial neural network–based reinforcement learning
  for wind turbine yaw control}, Wind Energy 23~(3) (2020) 676--690.
\newblock \href
  {http://arxiv.org/abs/https://onlinelibrary.wiley.com/doi/pdf/10.1002/we.2451}
  {\path{arXiv:https://onlinelibrary.wiley.com/doi/pdf/10.1002/we.2451}}, \href
  {http://dx.doi.org/10.1002/we.2451} {\path{doi:10.1002/we.2451}}.
\newline\urlprefix\url{https://onlinelibrary.wiley.com/doi/abs/10.1002/we.2451}

\bibitem{doi:10.1002/we.2497}
J.~W. Messner, P.~Pinson, J.~Browell, M.~B. Bjerregård, I.~Schicker,
  \href{https://onlinelibrary.wiley.com/doi/abs/10.1002/we.2497}{Evaluation of
  wind power forecasts—an up-to-date view}, Wind Energy 23~(6) (2020)
  1461--1481.
\newblock \href
  {http://arxiv.org/abs/https://onlinelibrary.wiley.com/doi/pdf/10.1002/we.2497}
  {\path{arXiv:https://onlinelibrary.wiley.com/doi/pdf/10.1002/we.2497}}, \href
  {http://dx.doi.org/10.1002/we.2497} {\path{doi:10.1002/we.2497}}.
\newline\urlprefix\url{https://onlinelibrary.wiley.com/doi/abs/10.1002/we.2497}

\bibitem{anomaly_dl}
G.~Helbing, M.~Ritter, Deep learning for fault detection in wind turbines,
  Renewable and Sustainable Energy Reviews 98 (2018) 189--198.
\newblock \href {http://dx.doi.org/10.1016/j.rser.2018.09.012}
  {\path{doi:10.1016/j.rser.2018.09.012}}.

\bibitem{phmsoc_datachallenge}
{PHM Society}, {2011 PHM society conference data challenge},
  \url{https://www.phmsociety.org/competition/phm/11}, accessed: 2021-01-09
  (2011).

\bibitem{hsb_prognosismatlab}
{E. {Bechhoefer}}, {Wind Turbine High-Speed Bearing Prognosis Dataset via
  MathWorks},
  \url{https://uk.mathworks.com/help/predmaint/ug/wind-turbine-high-speed-bearing-prognosis.html},
  accessed: 2021-01-09 (Oct 2013).

\bibitem{hsb_prognosisgithub}
{E. {Bechhoefer}}, {Wind Turbine High-Speed Bearing Prognosis Dataset via
  GitHub},
  \url{https://github.com/mathworks/WindTurbineHighSpeedBearingPrognosis-Data},
  accessed: 2021-01-09 (2018).

\bibitem{conf_hsbprognosis}
E.~Bechhoefer, B.~Hecke, D.~He, Processing for improved spectral analysis, in:
  Annual Conference of Prognostics and Health Management Society, Vol.~4, 2013,
  pp. 1--6.

\bibitem{engie_data}
{Engie Renewables}, {La Haute Borne Data},
  \url{https://opendata-renewables.engie.com/}, accessed: 2020-07-26.

\bibitem{nrel_wgbbench}
{National Renewable Energy Laboratory- National Wind Technology Center}, {Wind
  Turbine Gearbox Condition Monitoring Vibration Analysis Benchmarking
  Datasets},
  \url{https://openei.org/datasets/dataset/wind-turbine-gearbox-condition-monitoring-vibration-analysis-benchmarking-datasets},
  accessed: 2021-01-09 (Mar 2014).

\bibitem{ldt_data}
{ORE Catapult}, {Platform for Operational Data: Levenmouth Demonstration
  Turbine},
  \url{https://pod.ore.catapult.org.uk/source/levenmouth-demonstration-turbine},
  accessed: 2020-07-26.

\bibitem{orsted_data}
{Ørsted}, {Ørsted Offshore Operational Data },
  \url{https://orsted.com/en/our-business/offshore-wind/offshore-operational-data},
  accessed: 2021-01-02.

\bibitem{edpr_data}
{EDP Renewables}, {EDPR Wind Farm Data},
  \url{https://opendata.edp.com/pages/windfarms/ }, accessed: 2021-01-10
  (2018).

\bibitem{edpr_data_comp}
{EDP Renewables}, {Hack The Wind - Wind Turbine Failures Detection
  Competition},
  \url{https://opendata.edp.com/pages/hackthewind/description#description },
  accessed: 2021-01-10 (2018).

\bibitem{west_data}
{National Renewable Energy Laboratory}, {Western Wind Dataset},
  \url{https://www.nrel.gov/grid/western-wind-data.html}, accessed: 2020-07-26.

\bibitem{east_data}
{National Renewable Energy Laboratory}, {Eastern Wind Dataset},
  \url{https://www.nrel.gov/grid/eastern-wind-data.html}, accessed: 2020-07-26.

\bibitem{floating_data}
{ORE Catapult}, {Platform for Operational Data: Equinor Hywind Scotland
  Windfarm },
  \url{https://pod.ore.catapult.org.uk/source/equinor-hywind-scotland-windfarm},
  accessed: 2020-07-26.

\bibitem{FORESTI2020}
R.~Foresti, S.~Rossi, M.~Magnani, C.~{Guarino Lo Bianco}, N.~Delmonte,
  \href{http://www.sciencedirect.com/science/article/pii/S2095809920300266}{Smart
  society and artificial intelligence: Big data scheduling and the global
  standard method applied to smart maintenance}, Engineering\href
  {http://dx.doi.org/https://doi.org/10.1016/j.eng.2019.11.014}
  {\path{doi:https://doi.org/10.1016/j.eng.2019.11.014}}.
\newline\urlprefix\url{http://www.sciencedirect.com/science/article/pii/S2095809920300266}

\bibitem{en13184702}
D.~Menezes, M.~Mendes, J.~A. Almeida, T.~Farinha,
  \href{https://www.mdpi.com/1996-1073/13/18/4702}{Wind farm and resource
  datasets: A comprehensive survey and overview}, Energies 13~(18).
\newblock \href {http://dx.doi.org/10.3390/en13184702}
  {\path{doi:10.3390/en13184702}}.
\newline\urlprefix\url{https://www.mdpi.com/1996-1073/13/18/4702}

\bibitem{7011548}
A.~{Goudarzi}, I.~E. {Davidson}, A.~{Ahmadi}, G.~K. {Venayagamoorthy},
  Intelligent analysis of wind turbine power curve models, in: 2014 IEEE
  Symposium on Computational Intelligence Applications in Smart Grid (CIASG),
  2014, pp. 1--7.

\bibitem{Mittelmeier_2016}
N.~Mittelmeier, T.~Blodau, G.~Steinfeld, A.~Rott, M.~Kühn,
  \href{https://doi.org/10.1088%2F1742-6596%2F753%2F3%2F032052}{An analysis of
  offshore wind farm {SCADA} measurements to identify key parameters
  influencing the magnitude of wake effects}, Journal of Physics: Conference
  Series 753 (2016) 032052.
\newblock \href {http://dx.doi.org/10.1088/1742-6596/753/3/032052}
  {\path{doi:10.1088/1742-6596/753/3/032052}}.
\newline\urlprefix\url{https://doi.org/10.1088%2F1742-6596%2F753%2F3%2F032052}

\bibitem{goldstein_uchida_2016}
M.~Goldstein, S.~Uchida, A comparative evaluation of unsupervised anomaly
  detection algorithms for multivariate data, Plos One 11~(4).
\newblock \href {http://dx.doi.org/10.1371/journal.pone.0152173}
  {\path{doi:10.1371/journal.pone.0152173}}.

\bibitem{RePEc:eee:rensus:v:98:y:2018:i:c:p:189-198}
G.~Helbing, M.~Ritter,
  \href{https://ideas.repec.org/a/eee/rensus/v98y2018icp189-198.html}{{Deep
  Learning for fault detection in wind turbines}}, Renewable and Sustainable
  Energy Reviews 98~(C) (2018) 189--198.
\newblock \href {http://dx.doi.org/10.1016/j.rser.2018.09.01}
  {\path{doi:10.1016/j.rser.2018.09.01}}.
\newline\urlprefix\url{https://ideas.repec.org/a/eee/rensus/v98y2018icp189-198.html}

\bibitem{data_longtermchallenge}
G.~A.~M. van Kuik, et~al., Long-term research challenges in wind energy – a
  research agenda by the european academy of wind energy, WindEnergy Science 1
  (2016) 1--39.
\newblock \href {http://dx.doi.org/10.5194/wes-1-1-2016}
  {\path{doi:10.5194/wes-1-1-2016}}.

\bibitem{8862913}
Y.~{Roh}, G.~{Heo}, S.~E. {Whang}, A survey on data collection for machine
  learning: A big data - ai integration perspective, IEEE Transactions on
  Knowledge and Data Engineering (2019) 1--1\href
  {http://dx.doi.org/10.1109/TKDE.2019.2946162}
  {\path{doi:10.1109/TKDE.2019.2946162}}.

\bibitem{reliabilityanalysis_turbine}
C.~Zhu, Y.~Li, Reliability analysis of wind turbines, in: Stability Control and
  Reliable Performance of Wind Turbines, 2018.
\newblock \href {http://dx.doi.org/10.5772/intechopen.74859}
  {\path{doi:10.5772/intechopen.74859}}.

\bibitem{Leahy_2017}
K.~Leahy, C.~Gallagher, K.~Bruton, P.~O'Donovan, D.~T. O'Sullivan,
  \href{https://doi.org/10.1088/1742-6596/926/1/012011}{Automatically
  identifying and predicting unplanned wind turbine stoppages using {SCADA} and
  alarms system data: Case study and results}, Journal of Physics: Conference
  Series 926 (2017) 012011.
\newblock \href {http://dx.doi.org/10.1088/1742-6596/926/1/012011}
  {\path{doi:10.1088/1742-6596/926/1/012011}}.
\newline\urlprefix\url{https://doi.org/10.1088/1742-6596/926/1/012011}

\bibitem{Gonzalez_2016}
E.~Gonzalez, M.~Reder, J.~J. Melero,
  \href{https://doi.org/10.1088/1742-6596/753/7/072019}{{SCADA} alarms
  processing for wind turbine component failure detection}, Journal of Physics:
  Conference Series 753 (2016) 072019.
\newblock \href {http://dx.doi.org/10.1088/1742-6596/753/7/072019}
  {\path{doi:10.1088/1742-6596/753/7/072019}}.
\newline\urlprefix\url{https://doi.org/10.1088/1742-6596/753/7/072019}

\bibitem{nlgquality_paper1}
M.~Rikters, Impact of corpora quality on neural machine translation, Frontiers
  in Artificial Intelligence and Applications: Human Language Technologies –
  The Baltic Perspective 307 (2018) 126--133.
\newblock \href {http://dx.doi.org/10.3233/978-1-61499-912-6-126}
  {\path{doi:10.3233/978-1-61499-912-6-126}}.

\bibitem{doi:10.2514/6.2013-1695}
N.~J. Myrent, J.~F. Kusnick, D.~Adams, D.~T. Griffith,
  \href{https://arc.aiaa.org/doi/abs/10.2514/6.2013-1695}{Pitch Error and Shear
  Web Disbond Detection on Wind Turbine Blades for Offshore Structural Health
  and Prognostics Management}.
\newblock \href
  {http://arxiv.org/abs/https://arc.aiaa.org/doi/pdf/10.2514/6.2013-1695}
  {\path{arXiv:https://arc.aiaa.org/doi/pdf/10.2514/6.2013-1695}}, \href
  {http://dx.doi.org/10.2514/6.2013-1695} {\path{doi:10.2514/6.2013-1695}}.
\newline\urlprefix\url{https://arc.aiaa.org/doi/abs/10.2514/6.2013-1695}

\bibitem{joyjit_nips}
J.~Chatterjee, N.~Dethlefs, Natural language generation for operations and
  maintenance in wind turbines, in: NeurIPS 2019 Workshop on Tackling Climate
  Change with Machine Learning, Vancouver, Canada, 2019.

\bibitem{ji-etal-2020-amazing}
Y.~Ji, A.~Bosselut, T.~Wolf, A.~Celikyilmaz,
  \href{https://www.aclweb.org/anthology/2020.emnlp-tutorials.7}{The amazing
  world of neural language generation}, in: Proceedings of the 2020 Conference
  on Empirical Methods in Natural Language Processing: Tutorial Abstracts,
  Association for Computational Linguistics, Online, 2020, pp. 37--42.
\newblock \href {http://dx.doi.org/10.18653/v1/2020.emnlp-tutorials.7}
  {\path{doi:10.18653/v1/2020.emnlp-tutorials.7}}.
\newline\urlprefix\url{https://www.aclweb.org/anthology/2020.emnlp-tutorials.7}

\bibitem{khayrallah-koehn-2018-impact}
H.~Khayrallah, P.~Koehn, \href{https://www.aclweb.org/anthology/W18-2709}{On
  the impact of various types of noise on neural machine translation}, in:
  Proceedings of the 2nd Workshop on Neural Machine Translation and Generation,
  Association for Computational Linguistics, Melbourne, Australia, 2018, pp.
  74--83.
\newblock \href {http://dx.doi.org/10.18653/v1/W18-2709}
  {\path{doi:10.18653/v1/W18-2709}}.
\newline\urlprefix\url{https://www.aclweb.org/anthology/W18-2709}

\bibitem{dl_hugedata}
M.~Najafabadi, F.~Villanustre, T.~Khoshgoftaar, N.~Seliya, R.~Wald,
  E.~Muharemagic, Deep learning applications and challenges in big data
  analytics, Journal of Big Data 2.
\newblock \href {http://dx.doi.org/10.1186/s40537-014-0007-7}
  {\path{doi:10.1186/s40537-014-0007-7}}.

\bibitem{kusiak_2016}
A.~Kusiak, Renewables: Share data on wind energy, Nature 529~(7584) (2016)
  19–21.
\newblock \href {http://dx.doi.org/10.1038/529019a}
  {\path{doi:10.1038/529019a}}.

\bibitem{TOPHAM2017470}
E.~Topham, D.~McMillan,
  \href{http://www.sciencedirect.com/science/article/pii/S0960148116309430}{Sustainable
  decommissioning of an offshore wind farm}, Renewable Energy 102 (2017)
  470--480.
\newblock \href
  {http://dx.doi.org/https://doi.org/10.1016/j.renene.2016.10.066}
  {\path{doi:https://doi.org/10.1016/j.renene.2016.10.066}}.
\newline\urlprefix\url{http://www.sciencedirect.com/science/article/pii/S0960148116309430}

\bibitem{tl_defination}
K.~Weiss, T.~M. Khoshgoftaar, D.~Wang,
  \href{https://doi.org/10.1186/s40537-016-0043-6}{A survey of transfer
  learning}, Journal of Big Data 3~(1) (2016) 9.
\newblock \href {http://dx.doi.org/10.1186/s40537-016-0043-6}
  {\path{doi:10.1186/s40537-016-0043-6}}.
\newline\urlprefix\url{https://doi.org/10.1186/s40537-016-0043-6}

\bibitem{10.1007/978-3-319-97982-3_16}
M.~Hussain, J.~J. Bird, D.~R. Faria, A study on cnn transfer learning for image
  classification, in: A.~Lotfi, H.~Bouchachia, A.~Gegov, C.~Langensiepen,
  M.~McGinnity (Eds.), Advances in Computational Intelligence Systems, Springer
  International Publishing, Cham, 2019, pp. 191--202.

\bibitem{ruder-etal-2019-transfer}
S.~Ruder, M.~E. Peters, S.~Swayamdipta, T.~Wolf,
  \href{https://www.aclweb.org/anthology/N19-5004}{Transfer learning in natural
  language processing}, in: Proceedings of the 2019 Conference of the North
  {A}merican Chapter of the Association for Computational Linguistics:
  Tutorials, Association for Computational Linguistics, Minneapolis, Minnesota,
  2019, pp. 15--18.
\newblock \href {http://dx.doi.org/10.18653/v1/N19-5004}
  {\path{doi:10.18653/v1/N19-5004}}.
\newline\urlprefix\url{https://www.aclweb.org/anthology/N19-5004}

\bibitem{RePEc:eee:renene:v:85:y:2016:i:c:p:83-95}
Q.~Hu, R.~Zhang, Y.~Zhou,
  \href{https://ideas.repec.org/a/eee/renene/v85y2016icp83-95.html}{{Transfer
  learning for short-term wind speed prediction with deep neural networks}},
  Renewable Energy 85~(C) (2016) 83--95.
\newblock \href {http://dx.doi.org/10.1016/j.renene.2015.06.}
  {\path{doi:10.1016/j.renene.2015.06.}}
\newline\urlprefix\url{https://ideas.repec.org/a/eee/renene/v85y2016icp83-95.html}

\bibitem{8409794}
C.~{Zhang}, J.~{Bin}, Z.~{Liu}, Wind turbine ice assessment through inductive
  transfer learning, in: 2018 IEEE International Instrumentation and
  Measurement Technology Conference (I2MTC), 2018, pp. 1--6.

\bibitem{doi:10.1177/1475921720919073}
Y.~Pan, R.~Hong, J.~Chen, J.~Feng, W.~Wu,
  \href{https://doi.org/10.1177/1475921720919073}{Performance degradation
  assessment of wind turbine gearbox based on maximum mean discrepancy and
  multi-sensor transfer learning}, Structural Health Monitoring 0~(0) (0)
  1475921720919073.
\newblock \href {http://arxiv.org/abs/https://doi.org/10.1177/1475921720919073}
  {\path{arXiv:https://doi.org/10.1177/1475921720919073}}, \href
  {http://dx.doi.org/10.1177/1475921720919073}
  {\path{doi:10.1177/1475921720919073}}.
\newline\urlprefix\url{https://doi.org/10.1177/1475921720919073}

\bibitem{Wu_2020}
J.~Wu, Z.~Shao, S.~Yang,
  \href{https://doi.org/10.1088/1742-6596/1639/1/012044}{A combined algorithm
  for data cleaning of wind power scatter diagram considering actual
  engineering characteristics}, Journal of Physics: Conference Series 1639
  (2020) 012044.
\newblock \href {http://dx.doi.org/10.1088/1742-6596/1639/1/012044}
  {\path{doi:10.1088/1742-6596/1639/1/012044}}.
\newline\urlprefix\url{https://doi.org/10.1088/1742-6596/1639/1/012044}

\bibitem{llombart_robust}
A.~Llombart, C.~Pueyo, J.~Fandos, J.~Guerrero, Robust data filtering in wind
  power systems, European wind energy conference EWEC, Athens 2 (2006) 149--54.

\bibitem{sainz_robust}
E.~Sainz, A.~Llombart, J.~Guerrero, Robust filtering for the characterization
  of wind turbines: Improving its operation and maintenance, Energy Conversion
  and Management (2009) 2136--2147\href
  {http://dx.doi.org/10.1016/j.enconman.2009.04.036}
  {\path{doi:10.1016/j.enconman.2009.04.036}}.

\bibitem{8330024}
X.~{Shen}, X.~{Fu}, C.~{Zhou}, A combined algorithm for cleaning abnormal data
  of wind turbine power curve based on change point grouping algorithm and
  quartile algorithm, IEEE Transactions on Sustainable Energy 10~(1) (2019)
  46--54.
\newblock \href {http://dx.doi.org/10.1109/TSTE.2018.2822682}
  {\path{doi:10.1109/TSTE.2018.2822682}}.

\bibitem{melero_kalmanfilter}
J.~Melero, J.~Guerrero, J.~Beltran, C.~Pueyo, Efficient data filtering for wind
  energy assessment, Renewable Power Generation, IET 6 (2012) 446--454.
\newblock \href {http://dx.doi.org/10.1049/iet-rpg.2011.0288}
  {\path{doi:10.1049/iet-rpg.2011.0288}}.

\bibitem{chawla_bowyer_hall_kegelmeyer_2002}
N.~V. Chawla, K.~W. Bowyer, L.~O. Hall, W.~P. Kegelmeyer, Smote: Synthetic
  minority over-sampling technique, Journal of Artificial Intelligence Research
  16 (2002) 321--357.
\newblock \href {http://dx.doi.org/10.1613/jair.953}
  {\path{doi:10.1613/jair.953}}.

\bibitem{9302680}
H.~{Yi}, Q.~{Jiang}, X.~{Yan}, B.~{Wang}, Imbalanced classification based on
  minority clustering smote with wind turbine fault detection application, IEEE
  Transactions on Industrial Informatics (2020) 1--1\href
  {http://dx.doi.org/10.1109/TII.2020.3046566}
  {\path{doi:10.1109/TII.2020.3046566}}.

\bibitem{8245530}
Y.~{Ge}, D.~{Yue}, L.~{Chen}, Prediction of wind turbine blades icing based on
  mbk-smote and random forest in imbalanced data set, in: 2017 IEEE Conference
  on Energy Internet and Energy System Integration (EI2), 2017, pp. 1--6.
\newblock \href {http://dx.doi.org/10.1109/EI2.2017.8245530}
  {\path{doi:10.1109/EI2.2017.8245530}}.

\bibitem{SAEZ2016164}
J.~A. Sáez, B.~Krawczyk, M.~Woźniak,
  \href{http://www.sciencedirect.com/science/article/pii/S0031320316001072}{Analyzing
  the oversampling of different classes and types of examples in multi-class
  imbalanced datasets}, Pattern Recognition 57 (2016) 164 -- 178.
\newblock \href
  {http://dx.doi.org/https://doi.org/10.1016/j.patcog.2016.03.012}
  {\path{doi:https://doi.org/10.1016/j.patcog.2016.03.012}}.
\newline\urlprefix\url{http://www.sciencedirect.com/science/article/pii/S0031320316001072}

\bibitem{shoeybi2020megatronlm}
M.~Shoeybi, M.~Patwary, R.~Puri, P.~LeGresley, J.~Casper, B.~Catanzaro,
  Megatron-lm: Training multi-billion parameter language models using model
  parallelism (2020).
\newblock \href {http://arxiv.org/abs/1909.08053} {\path{arXiv:1909.08053}}.

\bibitem{DEMNERFUSHMAN2009760}
D.~Demner-Fushman, W.~W. Chapman, C.~J. McDonald,
  \href{http://www.sciencedirect.com/science/article/pii/S1532046409001087}{What
  can natural language processing do for clinical decision support?}, Journal
  of Biomedical Informatics 42~(5) (2009) 760 -- 772, biomedical Natural
  Language Processing.
\newblock \href {http://dx.doi.org/https://doi.org/10.1016/j.jbi.2009.08.007}
  {\path{doi:https://doi.org/10.1016/j.jbi.2009.08.007}}.
\newline\urlprefix\url{http://www.sciencedirect.com/science/article/pii/S1532046409001087}

\bibitem{4633969}
{Haibo He}, {Yang Bai}, E.~A. {Garcia}, {Shutao Li}, Adasyn: Adaptive synthetic
  sampling approach for imbalanced learning, in: 2008 IEEE International Joint
  Conference on Neural Networks (IEEE World Congress on Computational
  Intelligence), 2008, pp. 1322--1328.
\newblock \href {http://dx.doi.org/10.1109/IJCNN.2008.4633969}
  {\path{doi:10.1109/IJCNN.2008.4633969}}.

\bibitem{5559472}
S.~{Chen}, H.~{He}, E.~A. {Garcia}, Ramoboost: Ranked minority oversampling in
  boosting, IEEE Transactions on Neural Networks 21~(10) (2010) 1624--1642.
\newblock \href {http://dx.doi.org/10.1109/TNN.2010.2066988}
  {\path{doi:10.1109/TNN.2010.2066988}}.

\bibitem{8483334}
Y.~{Xie}, T.~{Zhang}, Imbalanced learning for fault diagnosis problem of
  rotating machinery based on generative adversarial networks, in: 2018 37th
  Chinese Control Conference (CCC), 2018, pp. 6017--6022.
\newblock \href {http://dx.doi.org/10.23919/ChiCC.2018.8483334}
  {\path{doi:10.23919/ChiCC.2018.8483334}}.

\bibitem{radford2019language}
A.~Radford, J.~Wu, R.~Child, D.~Luan, D.~Amodei, I.~Sutskever, Language models
  are unsupervised multitask learners.

\bibitem{DBLP:conf/nips/BrownMRSKDNSSAA20}
T.~B. Brown, B.~Mann, N.~Ryder, M.~Subbiah, J.~Kaplan, P.~Dhariwal,
  A.~Neelakantan, P.~Shyam, G.~Sastry, A.~Askell, S.~Agarwal,
  A.~Herbert{-}Voss, G.~Krueger, T.~Henighan, R.~Child, A.~Ramesh, D.~M.
  Ziegler, J.~Wu, C.~Winter, C.~Hesse, M.~Chen, E.~Sigler, M.~Litwin, S.~Gray,
  B.~Chess, J.~Clark, C.~Berner, S.~McCandlish, A.~Radford, I.~Sutskever,
  D.~Amodei,
  \href{https://proceedings.neurips.cc/paper/2020/hash/1457c0d6bfcb4967418bfb8ac142f64a-Abstract.html}{Language
  models are few-shot learners}, in: H.~Larochelle, M.~Ranzato, R.~Hadsell,
  M.~Balcan, H.~Lin (Eds.), Advances in Neural Information Processing Systems
  33: Annual Conference on Neural Information Processing Systems 2020, NeurIPS
  2020, December 6-12, 2020, virtual, 2020.
\newline\urlprefix\url{https://proceedings.neurips.cc/paper/2020/hash/1457c0d6bfcb4967418bfb8ac142f64a-Abstract.html}

\bibitem{KARIMI2020101759}
D.~Karimi, H.~Dou, S.~K. Warfield, A.~Gholipour,
  \href{http://www.sciencedirect.com/science/article/pii/S1361841520301237}{Deep
  learning with noisy labels: Exploring techniques and remedies in medical
  image analysis}, Medical Image Analysis 65 (2020) 101759.
\newblock \href {http://dx.doi.org/https://doi.org/10.1016/j.media.2020.101759}
  {\path{doi:https://doi.org/10.1016/j.media.2020.101759}}.
\newline\urlprefix\url{http://www.sciencedirect.com/science/article/pii/S1361841520301237}

\bibitem{GUPTA2019466}
S.~Gupta, A.~Gupta,
  \href{http://www.sciencedirect.com/science/article/pii/S1877050919318575}{Dealing
  with noise problem in machine learning data-sets: A systematic review},
  Procedia Computer Science 161 (2019) 466 -- 474, the Fifth Information
  Systems International Conference, 23-24 July 2019, Surabaya, Indonesia.
\newblock \href {http://dx.doi.org/https://doi.org/10.1016/j.procs.2019.11.146}
  {\path{doi:https://doi.org/10.1016/j.procs.2019.11.146}}.
\newline\urlprefix\url{http://www.sciencedirect.com/science/article/pii/S1877050919318575}

\bibitem{Minutti_2018}
C.~Minutti, S.~Gomez, G.~Ramos,
  \href{https://doi.org/10.1088/1742-6596/1047/1/012010}{A machine-learning
  approach for noise reduction in parameter estimation inverse problems,
  applied to characterization of oil reservoirs}, Journal of Physics:
  Conference Series 1047 (2018) 012010.
\newblock \href {http://dx.doi.org/10.1088/1742-6596/1047/1/012010}
  {\path{doi:10.1088/1742-6596/1047/1/012010}}.
\newline\urlprefix\url{https://doi.org/10.1088/1742-6596/1047/1/012010}

\bibitem{JEREZ2010105}
J.~M. Jerez, I.~Molina, P.~J. García-Laencina, E.~Alba, N.~Ribelles,
  M.~Martín, L.~Franco,
  \href{http://www.sciencedirect.com/science/article/pii/S0933365710000679}{Missing
  data imputation using statistical and machine learning methods in a real
  breast cancer problem}, Artificial Intelligence in Medicine 50~(2) (2010) 105
  -- 115.
\newblock \href
  {http://dx.doi.org/https://doi.org/10.1016/j.artmed.2010.05.002}
  {\path{doi:https://doi.org/10.1016/j.artmed.2010.05.002}}.
\newline\urlprefix\url{http://www.sciencedirect.com/science/article/pii/S0933365710000679}

\bibitem{10.1007/s10489-015-0666-x}
R.~Pan, T.~Yang, J.~Cao, K.~Lu, Z.~Zhang,
  \href{https://doi.org/10.1007/s10489-015-0666-x}{Missing data imputation by k
  nearest neighbours based on grey relational structure and mutual
  information}, Applied Intelligence 43~(3) (2015) 614–632.
\newblock \href {http://dx.doi.org/10.1007/s10489-015-0666-x}
  {\path{doi:10.1007/s10489-015-0666-x}}.
\newline\urlprefix\url{https://doi.org/10.1007/s10489-015-0666-x}

\bibitem{FOLGUERA2015146}
L.~Folguera, J.~Zupan, D.~Cicerone, J.~F. Magallanes,
  \href{http://www.sciencedirect.com/science/article/pii/S016974391500060X}{Self-organizing
  maps for imputation of missing data in incomplete data matrices},
  Chemometrics and Intelligent Laboratory Systems 143 (2015) 146 -- 151.
\newblock \href
  {http://dx.doi.org/https://doi.org/10.1016/j.chemolab.2015.03.002}
  {\path{doi:https://doi.org/10.1016/j.chemolab.2015.03.002}}.
\newline\urlprefix\url{http://www.sciencedirect.com/science/article/pii/S016974391500060X}

\bibitem{8052714}
D.~{Kalyanraj}, S.~L. {Prakash}, S.~{Sabareswar}, Wind turbine monitoring and
  control systems using internet of things, in: 2016 21st Century Energy Needs
  - Materials, Systems and Applications (ICTFCEN), 2016, pp. 1--4.
\newblock \href {http://dx.doi.org/10.1109/ICTFCEN.2016.8052714}
  {\path{doi:10.1109/ICTFCEN.2016.8052714}}.

\bibitem{IoTApplicationsinWindEnergyConversionSystems}
L.~Alhmoud, H.~Al-Zoubi,
  \href{https://www.degruyter.com/view/journals/eng/9/1/article-p490.xml}{Iot
  applications in wind energy conversion systems}, Open Engineering 9~(1) (01
  Jan. 2019) 490 -- 499.
\newblock \href {http://dx.doi.org/https://doi.org/10.1515/eng-2019-0061}
  {\path{doi:https://doi.org/10.1515/eng-2019-0061}}.
\newline\urlprefix\url{https://www.degruyter.com/view/journals/eng/9/1/article-p490.xml}

\bibitem{10.1145/3243904}
H.~Jin, B.~Liu, W.~Jiang, Y.~Ma, X.~Shi, B.~He, S.~Zhao,
  \href{https://doi.org/10.1145/3243904}{Layer-centric memory reuse and data
  migration for extreme-scale deep learning on many-core architectures}, ACM
  Trans. Archit. Code Optim. 15~(3).
\newblock \href {http://dx.doi.org/10.1145/3243904}
  {\path{doi:10.1145/3243904}}.
\newline\urlprefix\url{https://doi.org/10.1145/3243904}

\bibitem{MLSYS2020_084b6fbb}
P.~Jain, A.~Jain, A.~Nrusimha, A.~Gholami, P.~Abbeel, J.~Gonzalez, K.~Keutzer,
  I.~Stoica,
  \href{https://proceedings.mlsys.org/paper/2020/file/084b6fbb10729ed4da8c3d3f5a3ae7c9-Paper.pdf}{Checkmate:
  Breaking the memory wall with optimal tensor rematerialization}, in:
  I.~Dhillon, D.~Papailiopoulos, V.~Sze (Eds.), Proceedings of Machine Learning
  and Systems, Vol.~2, 2020, pp. 497--511.
\newline\urlprefix\url{https://proceedings.mlsys.org/paper/2020/file/084b6fbb10729ed4da8c3d3f5a3ae7c9-Paper.pdf}

\bibitem{inmemory_computing}
E.~Eleftheriou, M.~Gallo, N.~S.R., C.~Piveteau, I.~Boybat, V.~Joshi,
  R.~Khaddam-Aljameh, M.~Dazzi, I.~Giannopoulos, G.~Karunaratne, B.~Kersting,
  M.~Stanisavijevic, V.~Jonnalagadda, N.~Ioannou, K.~Kourtis, P.~Francese,
  A.~Sebastian, Deep learning acceleration based on in-memory computing, IBM
  Journal of Research and Development PP (2019) 1--1.
\newblock \href {http://dx.doi.org/10.1147/JRD.2019.2947008}
  {\path{doi:10.1147/JRD.2019.2947008}}.

\bibitem{largescalemem_frame}
G.~Nguyen, S.~Dlugolinsky, M.~Bobak, V.~Tran, A.~Lopez~Garcia, I.~Heredia,
  P.~Malík, L.~Hluchý, Machine learning and deep learning frameworks and
  libraries for large-scale data mining: a survey, Artificial Intelligence
  Review 52 (2019) 77--124.
\newblock \href {http://dx.doi.org/10.1007/s10462-018-09679-z}
  {\path{doi:10.1007/s10462-018-09679-z}}.

\bibitem{chen2015mxnet}
T.~Chen, M.~Li, Y.~Li, M.~Lin, N.~Wang, M.~Wang, T.~Xiao, B.~Xu, C.~Zhang,
  Z.~Zhang, Mxnet: A flexible and efficient machine learning library for
  heterogeneous distributed systems (2015).
\newblock \href {http://arxiv.org/abs/1512.01274} {\path{arXiv:1512.01274}}.

\bibitem{NIPS2019_9168}
R.~Anil, V.~Gupta, T.~Koren, Y.~Singer,
  \href{http://papers.nips.cc/paper/9168-memory-efficient-adaptive-optimization.pdf}{Memory
  efficient adaptive optimization}, in: H.~Wallach, H.~Larochelle,
  A.~Beygelzimer, F.~d\textquotesingle Alch\'{e}-Buc, E.~Fox, R.~Garnett
  (Eds.), Advances in Neural Information Processing Systems 32, Curran
  Associates, Inc., 2019, pp. 9749--9758.
\newline\urlprefix\url{http://papers.nips.cc/paper/9168-memory-efficient-adaptive-optimization.pdf}

\bibitem{vandeLeemput2019MemCNN}
S.~C.~v. Leemput, J.~Teuwen, B.~v. Ginneken, R.~Manniesing,
  \href{http://dx.doi.org/10.21105/joss.01576}{Memcnn: A python/pytorch package
  for creating memory-efficient invertible neural networks}, Journal of Open
  Source Software 4~(39) (2019) 1576.
\newblock \href {http://dx.doi.org/10.21105/joss.01576}
  {\path{doi:10.21105/joss.01576}}.
\newline\urlprefix\url{http://dx.doi.org/10.21105/joss.01576}

\bibitem{10.1145/3315573.3329984}
T.~D. Le, H.~Imai, Y.~Negishi, K.~Kawachiya,
  \href{https://doi.org/10.1145/3315573.3329984}{Automatic gpu memory
  management for large neural models in tensorflow}, in: Proceedings of the
  2019 ACM SIGPLAN International Symposium on Memory Management, ISMM 2019,
  Association for Computing Machinery, New York, NY, USA, 2019, p. 1–13.
\newblock \href {http://dx.doi.org/10.1145/3315573.3329984}
  {\path{doi:10.1145/3315573.3329984}}.
\newline\urlprefix\url{https://doi.org/10.1145/3315573.3329984}

\bibitem{37631}
V.~Vanhoucke, A.~Senior, M.~Z. Mao, Improving the speed of neural networks on
  cpus, in: Deep Learning and Unsupervised Feature Learning Workshop, NIPS
  2011, 2011.

\bibitem{8763885}
J.~{Chen}, X.~{Ran}, Deep learning with edge computing: A review, Proceedings
  of the IEEE 107~(8) (2019) 1655--1674.
\newblock \href {http://dx.doi.org/10.1109/JPROC.2019.2921977}
  {\path{doi:10.1109/JPROC.2019.2921977}}.

\bibitem{1575717}
D.~{Steinkraus}, I.~{Buck}, P.~Y. {Simard}, Using gpus for machine learning
  algorithms, in: Eighth International Conference on Document Analysis and
  Recognition (ICDAR'05), 2005, pp. 1115--1120 Vol. 2.
\newblock \href {http://dx.doi.org/10.1109/ICDAR.2005.251}
  {\path{doi:10.1109/ICDAR.2005.251}}.

\bibitem{wang2019benchmarking}
Y.~E. Wang, G.-Y. Wei, D.~Brooks, Benchmarking tpu, gpu, and cpu platforms for
  deep learning (2019).
\newblock \href {http://arxiv.org/abs/1907.10701} {\path{arXiv:1907.10701}}.

\bibitem{CHEN2020264}
Y.~Chen, Y.~Xie, L.~Song, F.~Chen, T.~Tang,
  \href{http://www.sciencedirect.com/science/article/pii/S2095809919306356}{A
  survey of accelerator architectures for deep neural networks}, Engineering
  6~(3) (2020) 264 -- 274.
\newblock \href {http://dx.doi.org/https://doi.org/10.1016/j.eng.2020.01.007}
  {\path{doi:https://doi.org/10.1016/j.eng.2020.01.007}}.
\newline\urlprefix\url{http://www.sciencedirect.com/science/article/pii/S2095809919306356}

\bibitem{10.1145/3295500.3356156}
S.~Lym, E.~Choukse, S.~Zangeneh, W.~Wen, S.~Sanghavi, M.~Erez,
  \href{https://doi.org/10.1145/3295500.3356156}{Prunetrain: Fast neural
  network training by dynamic sparse model reconfiguration}, in: Proceedings of
  the International Conference for High Performance Computing, Networking,
  Storage and Analysis, SC '19, Association for Computing Machinery, New York,
  NY, USA, 2019.
\newblock \href {http://dx.doi.org/10.1145/3295500.3356156}
  {\path{doi:10.1145/3295500.3356156}}.
\newline\urlprefix\url{https://doi.org/10.1145/3295500.3356156}

\bibitem{8578269}
A.~{Gordon}, E.~{Eban}, O.~{Nachum}, B.~{Chen}, H.~{Wu}, T.~{Yang}, E.~{Choi},
  Morphnet: Fast simple resource-constrained structure learning of deep
  networks, in: 2018 IEEE/CVF Conference on Computer Vision and Pattern
  Recognition, 2018, pp. 1586--1595.
\newblock \href {http://dx.doi.org/10.1109/CVPR.2018.00171}
  {\path{doi:10.1109/CVPR.2018.00171}}.

\bibitem{8279419}
D.~{He}, Z.~{Wang}, J.~{Liu}, A survey to predict the trend of ai-able server
  evolution in the cloud, IEEE Access 6 (2018) 10591--10602.
\newblock \href {http://dx.doi.org/10.1109/ACCESS.2018.2801293}
  {\path{doi:10.1109/ACCESS.2018.2801293}}.

\bibitem{8455947}
G.~{Clark}, M.~{Doran}, W.~{Glisson}, A malicious attack on the machine
  learning policy of a robotic system, in: 2018 17th IEEE International
  Conference On Trust, Security And Privacy In Computing And Communications/
  12th IEEE International Conference On Big Data Science And Engineering
  (TrustCom/BigDataSE), 2018, pp. 516--521.
\newblock \href {http://dx.doi.org/10.1109/TrustCom/BigDataSE.2018.00079}
  {\path{doi:10.1109/TrustCom/BigDataSE.2018.00079}}.

\bibitem{adv_attack}
S.~Qiu, Q.~Liu, S.~Zhou, C.~Wu, Review of artificial intelligence adversarial
  attack and defense technologies, Applied Sciences 9 (2019) 909.
\newblock \href {http://dx.doi.org/10.3390/app9050909}
  {\path{doi:10.3390/app9050909}}.

\bibitem{5772593}
J.~{Yan}, C.~{Liu}, M.~{Govindarasu}, Cyber intrusion of wind farm scada system
  and its impact analysis, in: 2011 IEEE/PES Power Systems Conference and
  Exposition, 2011, pp. 1--6.
\newblock \href {http://dx.doi.org/10.1109/PSCE.2011.5772593}
  {\path{doi:10.1109/PSCE.2011.5772593}}.

\bibitem{8591200}
A.~{Zabetian-Hosseini}, A.~{Mehrizi-Sani}, C.~{Liu}, Cyberattack to
  cyber-physical model of wind farm scada, in: IECON 2018 - 44th Annual
  Conference of the IEEE Industrial Electronics Society, 2018, pp. 4929--4934.
\newblock \href {http://dx.doi.org/10.1109/IECON.2018.8591200}
  {\path{doi:10.1109/IECON.2018.8591200}}.

\bibitem{8406613}
N.~{Papernot}, P.~{McDaniel}, A.~{Sinha}, M.~P. {Wellman}, Sok: Security and
  privacy in machine learning, in: 2018 IEEE European Symposium on Security and
  Privacy (EuroS P), 2018, pp. 399--414.
\newblock \href {http://dx.doi.org/10.1109/EuroSP.2018.00035}
  {\path{doi:10.1109/EuroSP.2018.00035}}.

\bibitem{ldpc_handling}
J.~Jang, I.~Jung, J.~Park, An effective handling of secure data stream in iot,
  Applied Soft Computing 68.
\newblock \href {http://dx.doi.org/10.1016/j.asoc.2017.05.020}
  {\path{doi:10.1016/j.asoc.2017.05.020}}.

\bibitem{sezer2019financial}
O.~B. Sezer, M.~U. Gudelek, A.~M. Ozbayoglu, Financial time series forecasting
  with deep learning : A systematic literature review: 2005-2019 (2019).
\newblock \href {http://arxiv.org/abs/1911.13288} {\path{arXiv:1911.13288}}.

\bibitem{ANGELOV2020185}
P.~Angelov, E.~Soares,
  \href{http://www.sciencedirect.com/science/article/pii/S0893608020302513}{Towards
  explainable deep neural networks (xdnn)}, Neural Networks 130 (2020)
  185--194.
\newblock \href
  {http://dx.doi.org/https://doi.org/10.1016/j.neunet.2020.07.010}
  {\path{doi:https://doi.org/10.1016/j.neunet.2020.07.010}}.
\newline\urlprefix\url{http://www.sciencedirect.com/science/article/pii/S0893608020302513}

\bibitem{zheng_fu_mei_luo_2017}
H.~Zheng, J.~Fu, T.~Mei, J.~Luo, Learning multi-attention convolutional neural
  network for fine-grained image recognition, 2017 IEEE International
  Conference on Computer Vision (ICCV)\href
  {http://dx.doi.org/10.1109/iccv.2017.557} {\path{doi:10.1109/iccv.2017.557}}.

\bibitem{NIPS2017_7062}
S.~M. Lundberg, S.-I. Lee,
  \href{http://papers.nips.cc/paper/7062-a-unified-approach-to-interpreting-model-predictions.pdf}{A
  unified approach to interpreting model predictions}, in: I.~Guyon, U.~V.
  Luxburg, S.~Bengio, H.~Wallach, R.~Fergus, S.~Vishwanathan, R.~Garnett
  (Eds.), Advances in Neural Information Processing Systems 30, Curran
  Associates, Inc., 2017, pp. 4765--4774.
\newline\urlprefix\url{http://papers.nips.cc/paper/7062-a-unified-approach-to-interpreting-model-predictions.pdf}

\bibitem{10.1145/2939672.2939778}
M.~T. Ribeiro, S.~Singh, C.~Guestrin,
  \href{https://doi.org/10.1145/2939672.2939778}{"why should i trust you?":
  Explaining the predictions of any classifier}, in: Proceedings of the 22nd
  ACM SIGKDD International Conference on Knowledge Discovery and Data Mining,
  KDD '16, Association for Computing Machinery, New York, NY, USA, 2016, p.
  1135–1144.
\newblock \href {http://dx.doi.org/10.1145/2939672.2939778}
  {\path{doi:10.1145/2939672.2939778}}.
\newline\urlprefix\url{https://doi.org/10.1145/2939672.2939778}

\bibitem{luong-etal-2015-effective}
T.~Luong, H.~Pham, C.~D. Manning,
  \href{https://www.aclweb.org/anthology/D15-1166}{Effective approaches to
  attention-based neural machine translation}, in: Proceedings of the 2015
  Conference on Empirical Methods in Natural Language Processing, Association
  for Computational Linguistics, Lisbon, Portugal, 2015, pp. 1412--1421.
\newblock \href {http://dx.doi.org/10.18653/v1/D15-1166}
  {\path{doi:10.18653/v1/D15-1166}}.
\newline\urlprefix\url{https://www.aclweb.org/anthology/D15-1166}

\bibitem{8955569}
X.~{Fu}, F.~{Gao}, J.~{Wu}, X.~{Wei}, F.~{Duan}, Spatiotemporal attention
  networks for wind power forecasting, in: 2019 International Conference on
  Data Mining Workshops (ICDMW), 2019, pp. 149--154.

\bibitem{en12244612}
Z.~Meng, X.~Xu, \href{https://www.mdpi.com/1996-1073/12/24/4612}{A hybrid
  short-term load forecasting framework with an attention-based
  encoder–decoder network based on seasonal and trend adjustment}, Energies
  12~(24).
\newblock \href {http://dx.doi.org/10.3390/en12244612}
  {\path{doi:10.3390/en12244612}}.
\newline\urlprefix\url{https://www.mdpi.com/1996-1073/12/24/4612}

\bibitem{10.1145/2939672.2939785}
T.~Chen, C.~Guestrin, \href{https://doi.org/10.1145/2939672.2939785}{Xgboost: A
  scalable tree boosting system}, in: Proceedings of the 22nd ACM SIGKDD
  International Conference on Knowledge Discovery and Data Mining, KDD '16,
  Association for Computing Machinery, New York, NY, USA, 2016, p. 785–794.
\newblock \href {http://dx.doi.org/10.1145/2939672.2939785}
  {\path{doi:10.1145/2939672.2939785}}.
\newline\urlprefix\url{https://doi.org/10.1145/2939672.2939785}

\bibitem{8329419}
D.~{Zhang}, L.~{Qian}, B.~{Mao}, C.~{Huang}, B.~{Huang}, Y.~{Si}, A data-driven
  design for fault detection of wind turbines using random forests and xgboost,
  IEEE Access 6 (2018) 21020--21031.

\bibitem{app10093258}
Z.~Wu, X.~Wang, B.~Jiang, \href{https://www.mdpi.com/2076-3417/10/9/3258}{Fault
  diagnosis for wind turbines based on relieff and extreme gradient boosting},
  Applied Sciences 10~(9).
\newblock \href {http://dx.doi.org/10.3390/app10093258}
  {\path{doi:10.3390/app10093258}}.
\newline\urlprefix\url{https://www.mdpi.com/2076-3417/10/9/3258}

\bibitem{7981134}
J.~{Browell}, C.~{Gilbert}, D.~{McMillan}, Use of turbine-level data for
  improved wind power forecasting, in: 2017 IEEE Manchester PowerTech, 2017,
  pp. 1--6.

\bibitem{RePEc:gam:jeners:v:12:y:2019:i:22:p:4224-:d:283963}
T.~Yuan, Z.~Sun, S.~Ma,
  \href{https://ideas.repec.org/a/gam/jeners/v12y2019i22p4224-d283963.html}{{Gearbox
  Fault Prediction of Wind Turbines Based on a Stacking Model and Change-Point
  Detection}}, Energies 12~(22) (2019) 1--20.
\newline\urlprefix\url{https://ideas.repec.org/a/gam/jeners/v12y2019i22p4224-d283963.html}

\bibitem{BARREDOARRIETA202082}
A.~{Barredo Arrieta}, N.~Díaz-Rodríguez, J.~{Del Ser}, A.~Bennetot, S.~Tabik,
  A.~Barbado, S.~Garcia, S.~Gil-Lopez, D.~Molina, R.~Benjamins, R.~Chatila,
  F.~Herrera,
  \href{http://www.sciencedirect.com/science/article/pii/S1566253519308103}{Explainable
  artificial intelligence (xai): Concepts, taxonomies, opportunities and
  challenges toward responsible ai}, Information Fusion 58 (2020) 82--115.
\newblock \href
  {http://dx.doi.org/https://doi.org/10.1016/j.inffus.2019.12.012}
  {\path{doi:https://doi.org/10.1016/j.inffus.2019.12.012}}.
\newline\urlprefix\url{http://www.sciencedirect.com/science/article/pii/S1566253519308103}

\end{thebibliography}


\end{document}